\documentclass{article}

\PassOptionsToPackage{numbers, compress}{natbib}

\usepackage[preprint]{neurips_2024}

\usepackage[utf8]{inputenc} 
\usepackage[T1]{fontenc}    
\usepackage{url}            
\usepackage{booktabs}       
\usepackage{amsfonts}       
\usepackage{nicefrac}       
\usepackage{microtype}      
\usepackage{xcolor}         
\usepackage{times}
\usepackage{epsfig}
\usepackage{amsmath}
\usepackage{makecell} 
\usepackage{tabularx}
\usepackage{diagbox}
\usepackage{enumitem} 

\usepackage{xspace}
\usepackage{graphicx}
\usepackage{subcaption}
\usepackage{multirow} 
\usepackage{colortbl}
\usepackage{pifont}
\usepackage{wrapfig}
\usepackage{tcolorbox}
\usepackage{float}
\usepackage{marvosym}
\definecolor{citecolor}{HTML}{0071BC}
\usepackage[
    pagebackref,           
    breaklinks=true,       
    colorlinks=true,       
    linkcolor=red,         
    citecolor=citecolor,   
    urlcolor=black,        
    bookmarks=false        
]{hyperref}

\newcommand{\ie}{{\emph{i.e.}}}
\newcommand{\eg}{{\emph{e.g.}}}
\newcommand{\etc}{\emph{etc}}
\newcommand{\etal}{\emph{et al.}}

\newcommand{\yes}{\ding{51}}
\newcommand{\no}{\ding{55}}

\title{
Expanding Performance Boundaries of Open-Source Multimodal Models with Model, Data, and Test-Time Scaling
}

\author{
\scalebox{0.79}{
Zhe Chen$^{4,1*\dagger}$, 
Weiyun Wang$^{5,1*\dagger}$, 
Yue Cao$^{4,1*\dagger}$,
Yangzhou Liu$^{4,1*\dagger}$,
Zhangwei Gao$^{7,1*\dagger}$,
Erfei Cui$^{7,1*\dagger}$,
Jinguo Zhu$^{1*}$,
}\\
\scalebox{0.79}{
\textbf{
Shenglong Ye$^{1*}$,
Hao Tian$^{2*}$,
Zhaoyang Liu$^{1*\dagger}$,
Lixin Gu$^{1}$,
Xuehui Wang$^{1\dagger}$,
Qingyun Li$^{1\dagger}$,
Yimin Reng$^{3,1\dagger}$,
Zixuan Chen$^{2}$,
}}\\ 
\scalebox{0.79}{
\textbf{
Jiapeng Luo$^{2}$,
Jiahao Wang$^{2}$,
Tan Jiang$^{2}$,
Bo Wang$^{2}$,
Conghui He$^{1}$,
Botian Shi$^{1}$,
Xingcheng Zhang$^{1}$,
Han Lv$^{1}$,
Yi Wang$^{1}$,
}}\\ 
\scalebox{0.79}{
\textbf{
Wenqi Shao$^{1}$,
Pei Chu$^{1}$,
Zhongying Tu$^{1}$,
Tong He$^{1}$,
Zhiyong Wu$^{1}$,
Huipeng Deng$^{1}$,
Jiaye Ge$^{1}$,
Kai Chen$^{1}$,
Kaipeng Zhang$^{1}$,
}}\\
\scalebox{0.79}{
\textbf{
Limin Wang$^{4,1}$,
Min Dou$^{1}$,
Lewei Lu$^{2}$,
Xizhou Zhu$^{3,1}$,
Tong Lu$^{4}$,
Dahua Lin$^{6,1}$,
Yu Qiao$^{1}$,
Jifeng Dai$^{3,1}$\textsuperscript{\Letter}, 
Wenhai Wang$^{6,1}$\textsuperscript{\Letter} 
}}
\vspace{4px}\\
\scalebox{0.91}{
$^1$Shanghai AI Laboratory~~~
$^2$SenseTime Research~~~
$^3$Tsinghua University~~~
$^4$Nanjing University~~~}\\
\scalebox{0.91}{
$^5$Fudan University~~~
$^6$The Chinese University of Hong Kong~~~
$^7$Shanghai Jiao Tong University
}\\
\\
\small Code: \url{https://github.com/OpenGVLab/InternVL} \\
\small Model: \url{https://huggingface.co/OpenGVLab/InternVL2_5-78B} \\
\small HF Demo: \url{https://huggingface.co/spaces/OpenGVLab/InternVL} \\
}

\newcommand\blfootnote[1]{%
\begingroup
\renewcommand\thefootnote{}\footnote{#1}%
\addtocounter{footnote}{-1}%
\endgroup
}

\definecolor{baselinecolor}{gray}{.9}

\begin{document}

\maketitle

\blfootnote{
* equal contribution; $\dagger$ interns at OpenGVLab, Shanghai AI Laboratory; \\
\Letter\  corresponding authors (daijifeng@tsinghua.edu.cn, wangwenhai@pjlab.org.cn).
}

\begin{abstract}
We introduce InternVL 2.5, an advanced multimodal large language model (MLLM) series that builds upon InternVL 2.0, maintaining its core model architecture while introducing significant enhancements in training and testing strategies as well as data quality.
In this work, we delve into the relationship between model scaling and performance, systematically exploring the performance trends in vision encoders, language models, dataset sizes, and test-time configurations.
Through extensive evaluations on a wide range of benchmarks, including multi-discipline reasoning, document understanding, multi-image / video understanding, real-world comprehension, multimodal hallucination detection, visual grounding, multilingual capabilities, and pure language processing, InternVL 2.5 exhibits competitive performance, rivaling leading commercial models such as GPT-4o and Claude-3.5-Sonnet. Notably, our model is the first open-source MLLMs to surpass 70\% on the MMMU benchmark, achieving a 3.7-point improvement through Chain-of-Thought (CoT) reasoning and showcasing strong potential for test-time scaling. HuggingFace demo see \url{https://huggingface.co/spaces/OpenGVLab/InternVL}

\end{abstract}

\section{Introduction}

In recent years, multimodal large language models (MLLMs)~\cite{dong2024xc24khd, li2021improved, wang2024qwen2vl, chen2023internvl, chen2024far, wang2023cogvlm, li2023monkey, team2023gemini, gpt4v, ye2023mplug2, lin2024vila, deitke2024molmo, lu2024bluelm} have emerged as a pivotal technology in artificial intelligence, capable of processing and understanding information from multiple modalities such as text, images, and videos. These models promise breakthroughs across fields like natural language processing, computer vision, and human-computer interaction. However, developing large-scale MLLMs remains a challenging task, requiring significant computational resources, sophisticated architectures, and the ability to effectively integrate diverse data types in a scalable manner. 

Various attempts have been made to address these challenges, including enhancing model architectures~\cite{sun2023emu, tian2024mminterleaved, alayrac2022flamingo, luo2024mono_internvl, liu2024points, shi2024eagle}, scaling vision encoders~\cite{wang2023internimage, fang2022eva, chen2023internvl, zhai2023siglip, mckinzie2024mm1} and language models~\cite{teknium2024hermes3, touvron2023llama2, dubey2024llama3, cai2024internlm2, qwen2.5, sun2023moss, du2022glm}, incorporating more diverse and high-quality datasets~\cite{li2024llavaov, tong2024cambrian, chen2024allava, liu2024mminstruct}, and refining the test-time scaling process~\cite{snell2024scaling, wang2024mpo, qwq-32b-preview} to boost performance. Notable commercial models, like GPT-4o~\cite{gpt4v} and Claude-3.5-Sonnet~\cite{claude3series2024}, have demonstrated exceptional performance, their closed nature limits transparency and accessibility, leaving gaps in the open-source community.
While open-source multimodal models such as the InternVL series~\cite{chen2023internvl, chen2024far, gao2024mini_internvl} and Qwen-VL series~\cite{bai2023qwenvl, wang2024qwen2vl} have provided high-performance, transparent alternatives, they still fall short in terms of achieving the desired levels of performance and efficiency.

\begin{figure*}[t!]
    \centering
    \includegraphics[width=1\linewidth]{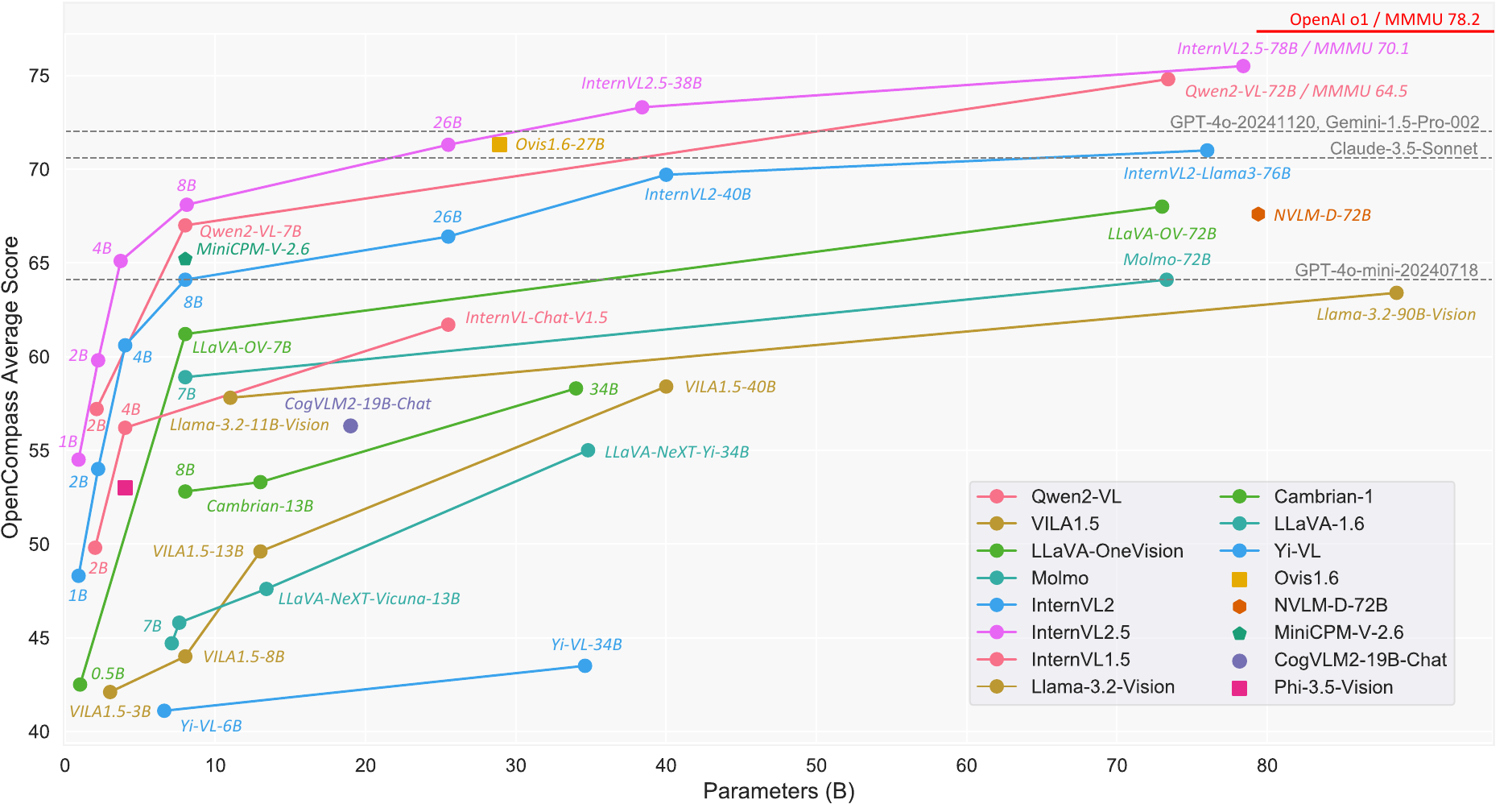}
    \caption{
    \textbf{Performance of various MLLMs on the OpenCompass leaderboard.} 
    InternVL 2.5 showcases strong multimodal capabilities, rivaling closed-source models like GPT-4o~\cite{gpt4v} and Claude-3.5-Sonnet~\cite{claude3series2024}. 
    However, since the OpenCompass score is derived from 8 academic VQA benchmarks and covers only a subset of overall capabilities, we still need further effort to match the performance with closed-source models. 
    } 
    \label{fig:overall_performance}
\end{figure*}

In this work, we introduce InternVL 2.5, an advanced large-scale MLLM series that builds upon the foundational architecture of InternVL 2.0. Continuing the objectives of the entire InternVL series, we aim to bridge the performance gap between commercial closed-source models and open-source multimodal models.
In InternVL 2.5, we systematically explore various factors in MLLM, including how changes in vision encoders, language models, dataset sizes, and inference times affect the overall performance of the model, demonstrating the relationship between scaling and performance in multimodal models. Specifically, we have some interesting findings: 
(1) \emph{Large vision encoders significantly reduce the dependency on training data when scaling up MLLMs.} 
As shown in Table~\ref{tab:train_hyperparameter}, compared to Qwen2-VL-72B~\cite{wang2024qwen2vl} equipped with a 600M vision encoder, our InternVL2.5-78B with a 6B vision encoder can achieve better performance using only 1/10 of the training tokens. This greatly reduces the exploration cost when scaling up MLLMs; 
(2) \emph{Data quality matters.} 
Upgrading InternVL from 2.0 to 2.5 doubled the dataset size, but strict filtering greatly improved quality. For example, we carefully excluded the anomalous samples (\eg, repetitive patterns), achieving substantial improvements in Chain-of-Thought (CoT) reasoning tasks such as MMMU~\cite{yue2023mmmu} and complex challenges like the OlympiadBench~\cite{he2024olympiadbench}. Note that, most existing open-source MLLMs tend to underperform when using CoT~\cite{wang2024mpo}.
(3) \emph{Test-time scaling is beneficial for difficult multimodal QA.} For challenging tasks such as MMMU, the InternVL2.5-78B with CoT reaches 70.1\%, which is 3.7 points higher than the direct response. Subsequently, we have successfully verified that CoT can be further combined with majority voting and bring additional improvements.

Our contributions can be summarized as follows:

(1) We release InternVL 2.5 to the open-source community, providing a powerful tool for the development and application of multimodal AI systems and encouraging further research in this domain.

(2) We investigate how scaling different components of the MLLMs such as vision encoders, language models, dataset sizes, and inference time affect performance.

(3) Through extensive evaluations on diverse benchmarks---including multi-discipline reasoning, document understanding, multi-image / video understanding, real-world comprehension, multimodal hallucination detection, visual grounding, multilingual capabilities, and pure language processing—InternVL 2.5 exhibits competitive performance, rivaling leading commercial models like GPT-4o~\cite{gpt4v} and Claude-3.5-Sonnet~\cite{claude3series2024}. It is the first open-source MLLM to surpass 70\% on the MMMU validation set~\cite{yue2023mmmu}, setting a new benchmark and highlighting the potential of open-source solutions in advancing multimodal AI.

\section{Model Architecture}
\label{sec:model_architecture}

\begin{figure*}[t!]
    \centering
    \includegraphics[width=1\linewidth]{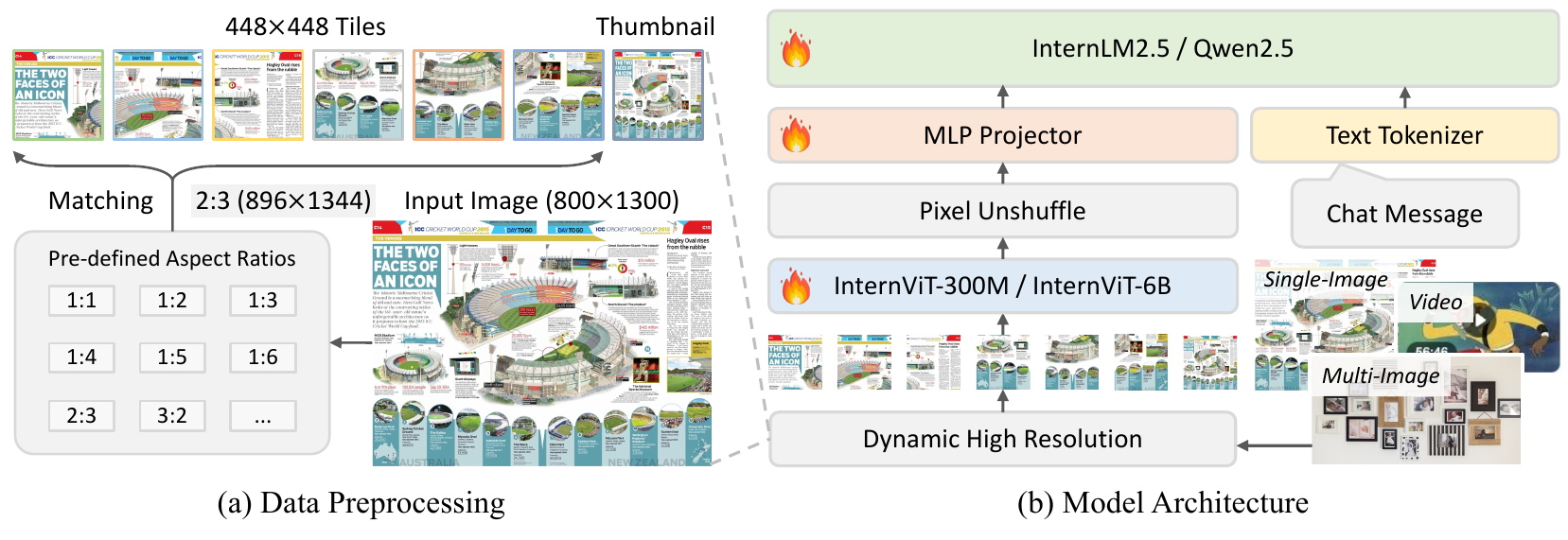}
    \caption{\textbf{Overall architecture.}
    InternVL 2.5 retains the same model architecture as InternVL 1.5~\cite{chen2024far} and InternVL 2.0, \ie~the widely-used ``ViT-MLP-LLM'' paradigm, which combines a pre-trained InternViT-300M or InternViT-6B with LLMs~\cite{cai2024internlm2,qwen2.5} of various sizes via an MLP projector. Consistent with previous versions, we apply a pixel unshuffle operation to reduce the 1024 visual tokens produced by each 448$\times$448 image tile to 256 tokens. 
    Moreover, compared to InternVL 1.5, InternVL 2.0 and 2.5 introduced additional data types, incorporating multi-image and video data alongside the existing single-image and text-only data.
    } 
    \label{fig:overall}
\end{figure*}

\subsection{Overall Architecture}

As shown in Figure~\ref{fig:overall} and Table~\ref{tab:pretrain_model_detail}, InternVL 2.5 retains the same model architecture as its predecessors, InternVL 1.5~\cite{chen2024far} and InternVL 2.0, following the ``ViT-MLP-LLM'' paradigm widely adopted in various MLLM studies~\cite{liu2023improved, liu2024llavanext, chen2023internvl, zhu2023minigpt4, lu2024deepseekvl, wang2024qwen2vl, li2024llavaov, wang2024internvideo2}. 

In this new version, our implementation of this architecture integrates a newly incrementally pre-trained InternViT-6B or InternViT-300M with various pre-trained LLMs of different sizes and types, including InternLM 2.5~\cite{cai2024internlm2} and Qwen 2.5~\cite{qwen2.5}, using a randomly initialized 2-layer MLP projector. As in the previous version, to enhance scalability for high-resolution processing, we simply applied a pixel unshuffle operation, reducing the number of visual tokens to one-quarter of the original. Consequently, in our model, a 448$\times$448 image tile is represented by 256 visual tokens.

In terms of input data preprocessing, we adopted a similar dynamic resolution strategy as InternVL 1.5, dividing images into tiles of 448$\times$448 pixels based on the aspect ratio and resolution of the input images. The key difference, starting from InternVL 2.0, is that we additionally introduced support for multi-image and video data, as shown in Figure~\ref{fig:overall}(b). Different data types correspond to different preprocessing configurations, which we will detail in Section \ref{sec:dynamic_high_resolution}.

\subsection{Vision Encoder}

InternVL employs InternViT~\cite{chen2023internvl} as the vision encoder. To better document the training progression of InternViT, we have provided detailed information in Table \ref{tab:internvit_detail}. InternViT currently has two different model sizes, including InternViT-6B and InternViT-300M.

\textbf{InternViT-6B.}~~\href{https://huggingface.co/OpenGVLab/InternViT-6B-224px}{InternViT-6B-224px} was first introduced in our CVPR paper~\cite{chen2023internvl}, and its structure follows the vanilla ViT~\cite{dosovitskiy2020image}, with minor adjustments incorporating QK-Norm~\cite{dehghani2023vit22b} and RMSNorm~\cite{zhang2019rmsnorm}. 
It had 5.9B parameters, 48 layers, a hidden size of 3200, and 25 heads, and it was trained using a contrastive loss~\cite{radford2021clip}.
Due to the limited gains at that time, we adopted an incremental pre-training strategy to continuously refine its weights. Specifically, we connected InternViT-6B to an LLM via an MLP projector and, following a brief MLP warmup, jointly trained the InternViT-6B using a next token prediction loss (as shown in Figure~\ref{fig:training_strategy}(a)) to enhance its visual feature extraction capabilities. 
In the V1.0 and V1.2 versions, we used a fixed resolution of 448$\times$448 for training, but in later versions, we switched to dynamic resolution training to improve high-resolution processing. 
As detailed in the InternVL 1.5 report~\cite{chen2024far}, we removed the last three layers of \href{https://huggingface.co/OpenGVLab/InternViT-6B-448px-V1-2}{InternViT-6B-448px-V1.2}, reducing its depth from 48 to 45 layers, as these layers were more tuned to the CLIP loss objective, prioritizing global alignment over local information.
As a result, all subsequent versions, including the latest \href{https://huggingface.co/OpenGVLab/InternViT-6B-448px-V2_5}{InternViT-6B-448px-V2.5}, have 45 layers and 5.5B parameters.

\textbf{InternViT-300M.}~~InternViT-300M-448px-Distill is a distilled variant of the teacher model, \href{https://huggingface.co/OpenGVLab/InternViT-6B-448px-V1-5}{InternViT-6B-448px-V1.5}, utilizing a cosine distillation loss.
This model comprises 0.3B parameters, 24 layers, a hidden size of 1024, and 16 attention heads. Unlike the 6B version, the 0.3B variant employs standard LayerNorm~\cite{ba2016layer} without QK-Norm~\cite{dehghani2023vit22b}.
To reduce distillation costs, we initialized this model using CLIP-ViT-Large-336px~\cite{radford2021clip} where applicable, despite some architectural differences. 
After distillation, we integrated this model with an LLM and, following a similar procedure as described above, trained the vision encoder with dynamic high-resolution and the NTP loss. 
Then, we extracted the vision encoder and released it as \href{https://huggingface.co/OpenGVLab/InternViT-300M-448px}{InternViT-300M-448px}.
In this report, we further refined the InternViT-300M by incrementally pre-training the previous weights on a more diverse data mixture using the NTP loss, leading to the enhanced \href{https://huggingface.co/OpenGVLab/InternViT-300M-448px-V2_5}{InternViT-300M-448px-V2.5}.

\begin{table}[t]
    \centering
    {\fontsize{8}{10}\selectfont 
    \renewcommand\arraystretch{1.05} 
    \setlength\tabcolsep{4.3pt}
    \begin{tabular}{l|c|cccccc|c|c}
    Model Name                                                                                   &  Train Res. & Width & Depth & MLP   & \#Heads & QK-Norm & Norm Type & Loss Type & \#Param    \\
    \hline
    \href{https://huggingface.co/OpenGVLab/InternViT-6B-224px}{InternViT-6B-224px}               & fixed 224   & 3200  & 48    & 12800 & 25      & \yes    & RMS       & CLIP      & 5.9B       \\
    \href{https://huggingface.co/OpenGVLab/InternViT-6B-448px-V1-0}{InternViT-6B-448px-V1.0}     & fixed 448   & 3200  & 48    & 12800 & 25      & \yes    & RMS       & NTP       & 5.9B       \\
    \href{https://huggingface.co/OpenGVLab/InternViT-6B-448px-V1-2}{InternViT-6B-448px-V1.2}     & fixed 448   & 3200  & 45    & 12800 & 25      & \yes    & RMS       & NTP       & 5.5B       \\
    \href{https://huggingface.co/OpenGVLab/InternViT-6B-448px-V1-5}{InternViT-6B-448px-V1.5}     & dynamic 448 & 3200  & 45    & 12800 & 25      & \yes    & RMS       & NTP       & 5.5B       \\
    \rowcolor{gray!15}
    \href{https://huggingface.co/OpenGVLab/InternViT-6B-448px-V2_5}{InternViT-6B-448px-V2.5}     & dynamic 448 & 3200  & 45    & 12800 & 25      & \yes    & RMS       & NTP       & 5.5B       \\
    \hline
    InternViT-300M-448px-Distill                                                                 & fixed 448   & 1024  & 24    & 4096  & 16      & \no     & LN        & Cosine    & 0.3B       \\
    \href{https://huggingface.co/OpenGVLab/InternViT-300M-448px}{InternViT-300M-448px}           & dynamic 448 & 1024  & 24    & 4096  & 16      & \no     & LN        & NTP       & 0.3B       \\
    \rowcolor{gray!15}
    \href{https://huggingface.co/OpenGVLab/InternViT-300M-448px-V2_5}{InternViT-300M-448px-V2.5} & dynamic 448 & 1024  & 24    & 4096  & 16      & \no     & LN        & NTP       & 0.3B       \\
    \end{tabular}
    \vspace{1em}
    }
    \caption{\textbf{Details of InternViT-6B and InternViT-300M models.}
    ``fixed 224'' refers to training images resized to 224$\times$224, while ``dynamic 448'' means the model is trained with dynamic high resolution, with each image tile being 448$\times$448.
    ``CLIP'' refers to the contrastive loss, ``Cosine'' represents the cosine distillation loss, while ``NTP'' indicates  the next token prediction loss.
}
\label{tab:internvit_detail}
\end{table}

\begin{table}[t]
    {\fontsize{8}{10}\selectfont 
    \renewcommand\arraystretch{1.05} 
    \centering
    \setlength\tabcolsep{9.5pt}
    \newcommand{\TODO}{\textcolor{red}{TODO}}
    \centering
    \begin{tabular}{l|c|c|c|c}
        Model Name & \#Param & Vision Encoder & Language Model & OpenCompass \\
        \hline
        \href{https://huggingface.co/OpenGVLab/InternVL-Chat-V1-5}{InternVL-Chat-V1.5} & 25.5B &  \href{https://huggingface.co/OpenGVLab/InternViT-6B-448px-V1-5}{InternViT-6B-448px-V1.5} & \href{https://huggingface.co/internlm/internlm2-chat-20b}{internlm2-chat-20b} & 61.7 \\
        \href{https://huggingface.co/OpenGVLab/InternVL2-1B}{InternVL2-1B} & 0.9B & \href{https://huggingface.co/OpenGVLab/InternViT-300M-448px}{InternViT-300M-448px} & \href{https://huggingface.co/Qwen/Qwen2-0.5B-Instruct}{Qwen2-0.5B-Instruct} & 48.3 \\
        \href{https://huggingface.co/OpenGVLab/InternVL2-2B}{InternVL2-2B} & 2.2B & \href{https://huggingface.co/OpenGVLab/InternViT-300M-448px}{InternViT-300M-448px} & \href{https://huggingface.co/internlm/internlm2-chat-1_8b}{internlm2-chat-1.8b} & 54.0 \\
        \href{https://huggingface.co/OpenGVLab/InternVL2-4B}{InternVL2-4B} & 4.2B & \href{https://huggingface.co/OpenGVLab/InternViT-300M-448px}{InternViT-300M-448px} & \href{https://huggingface.co/microsoft/Phi-3-mini-128k-instruct}{Phi-3-mini-128k-instruct} & 60.6 \\
        \href{https://huggingface.co/OpenGVLab/InternVL2-8B}{InternVL2-8B} & 8.1B & \href{https://huggingface.co/OpenGVLab/InternViT-300M-448px}{InternViT-300M-448px} & \href{https://huggingface.co/internlm/internlm2_5-7b-chat}{internlm2\_5-7b-chat} & 64.1 \\
        \href{https://huggingface.co/OpenGVLab/InternVL2-26B}{InternVL2-26B} & 25.5B & \href{https://huggingface.co/OpenGVLab/InternViT-6B-448px-V1-5}{InternViT-6B-448px-V1.5} & \href{https://huggingface.co/internlm/internlm2-chat-20b}{internlm2-chat-20b} & 66.4 \\
        \href{https://huggingface.co/OpenGVLab/InternVL2-40B}{InternVL2-40B} & 40.1B & \href{https://huggingface.co/OpenGVLab/InternViT-6B-448px-V1-5}{InternViT-6B-448px-V1.5} & \href{https://huggingface.co/NousResearch/Nous-Hermes-2-Yi-34B}{Nous-Hermes-2-Yi-34B} & 69.7 \\
        \href{https://huggingface.co/OpenGVLab/InternVL2-Llama3-76B}{InternVL2-Llama3-76B} & 76.3B & \href{https://huggingface.co/OpenGVLab/InternViT-6B-448px-V1-5}{InternViT-6B-448px-V1.5} & \href{https://huggingface.co/NousResearch/Hermes-2-Theta-Llama-3-70B}{Hermes-2-Theta-Llama-3-70B} & 71.0 \\
        \hline
        \rowcolor{gray!15}
        \href{https://huggingface.co/OpenGVLab/InternVL2_5-1B}{InternVL2.5-1B} & 0.9B & \href{https://huggingface.co/OpenGVLab/InternViT-300M-448px-V2_5}{InternViT-300M-448px-V2.5} & \href{https://huggingface.co/Qwen/Qwen2.5-0.5B-Instruct}{Qwen2.5-0.5B-Instruct}  & 54.5 \\
        \rowcolor{gray!15}
        \href{https://huggingface.co/OpenGVLab/InternVL2_5-2B}{InternVL2.5-2B} & 2.2B & \href{https://huggingface.co/OpenGVLab/InternViT-300M-448px-V2_5}{InternViT-300M-448px-V2.5} & \href{https://huggingface.co/internlm/internlm2_5-1_8b-chat}{internlm2\_5-1\_8b-chat} & 59.8 \\
        \rowcolor{gray!15}
        \href{https://huggingface.co/OpenGVLab/InternVL2_5-4B}{InternVL2.5-4B} & 3.7B & \href{https://huggingface.co/OpenGVLab/InternViT-300M-448px-V2_5}{InternViT-300M-448px-V2.5} & \href{https://huggingface.co/Qwen/Qwen2.5-3B-Instruct}{Qwen2.5-3B-Instruct} & 65.1 \\
        \rowcolor{gray!15}
        \href{https://huggingface.co/OpenGVLab/InternVL2_5-8B}{InternVL2.5-8B} & 8.1B & \href{https://huggingface.co/OpenGVLab/InternViT-300M-448px-V2_5}{InternViT-300M-448px-V2.5} & \href{https://huggingface.co/internlm/internlm2_5-7b-chat}{internlm2\_5-7b-chat} & 68.1 \\
        \rowcolor{gray!15}
        \href{https://huggingface.co/OpenGVLab/InternVL2_5-26B}{InternVL2.5-26B} & 25.5B & \href{https://huggingface.co/OpenGVLab/InternViT-6B-448px-V2_5}{InternViT-6B-448px-V2.5} & \href{https://huggingface.co/internlm/internlm2_5-20b-chat}{internlm2\_5-20b-chat} & 71.3 \\
        \rowcolor{gray!15}
        \href{https://huggingface.co/OpenGVLab/InternVL2_5-38B}{InternVL2.5-38B} & 38.4B & \href{https://huggingface.co/OpenGVLab/InternViT-6B-448px-V2_5}{InternViT-6B-448px-V2.5} & \href{https://huggingface.co/Qwen/Qwen2.5-32B-Instruct}{Qwen2.5-32B-Instruct} & 73.3 \\
        \rowcolor{gray!15}
        \href{https://huggingface.co/OpenGVLab/InternVL2_5-78B}{InternVL2.5-78B} & 78.4B & \href{https://huggingface.co/OpenGVLab/InternViT-6B-448px-V2_5}{InternViT-6B-448px-V2.5} & \href{https://huggingface.co/Qwen/Qwen2.5-72B-Instruct}{Qwen2.5-72B-Instruct} & 75.5 \\
        \rowcolor{gray!15}
        InternVL2.5-Pro & --    & \href{https://huggingface.co/OpenGVLab/InternViT-6B-448px-V2_5}{InternViT-6B-448px-V2.5} & -- & -- \\
    \end{tabular}
    }
    \vspace{1em}
    \caption{
    \textbf{Pre-trained models used in the InternVL series.}
    In the InternVL 2.5 series, we upgraded both the vision encoder and the language model, resulting in improved performance.
    The OpenCompass scores for InternVL 1.5 and InternVL 2.0 were collected from the OpenCompass leaderboard, while the scores for InternVL 2.5 series were obtained through our local testing. 
    }
\label{tab:pretrain_model_detail}
\end{table}

\subsection{Large Language Model}

In Table \ref{tab:pretrain_model_detail}, we provide an overview of the language models used across different versions of InternVL, including InternVL 1.5, InternVL 2.0, and the latest InternVL 2.5. As shown, earlier versions primarily built on language models such as InternLM 2~\cite{cai2024internlm2}, Qwen 2~\cite{yang2024qwen2}, Phi 3~\cite{abdin2024phi3}, Yi~\cite{young2024yi}, and Llama 3~\cite{dubey2024llama3}. To achieve better performance, in the InternVL 2.5 series, we have comprehensively upgraded the language backbones to the latest state-of-the-art models, including InternLM 2.5~\cite{cai2024internlm2} and Qwen 2.5~\cite{qwen2.5}.

\begin{figure*}[t!]
    \centering
    \includegraphics[width=1\linewidth]{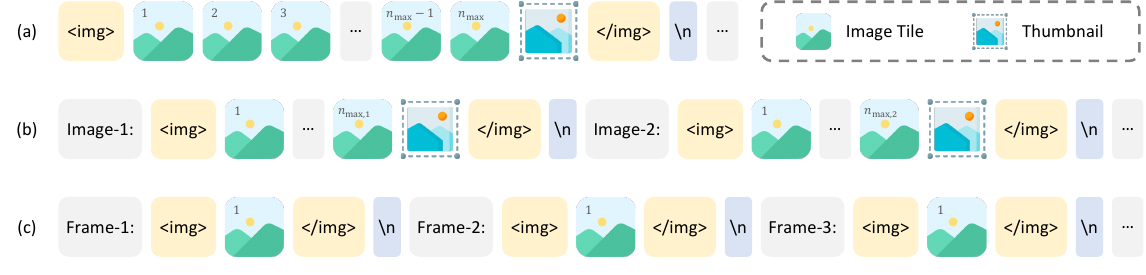}
    \caption{\textbf{Illustration of the data formats for various data types.}
    (a) For single-image datasets, the maximum number of tiles \( n_{\text{max}} \) is allocated to a single image, ensuring maximum resolution for the input.
    (b) For multi-image datasets, the total number of tiles \( n_{\text{max}} \) is distributed proportionally across all images within the sample.
    (c) For video datasets, the method simplifies the approach by setting \( n_{\text{max}} = 1 \), resizing individual frames to a fixed resolution of 448$\times$448.
    } 
    \label{fig:data_format}
\end{figure*}

\section{Training Strategy}

\subsection{Dynamic High-Resolution for Multimodal Data}
\label{sec:dynamic_high_resolution}

In InternVL 2.0 and 2.5, we extend the dynamic high-resolution training approach introduced in InternVL 1.5~\cite{chen2024far}, enhancing its capabilities to handle multi-image and video datasets.
The process mainly consists of the following steps:

\textbf{Closest Aspect Ratio Matching.} 
Given an input image \( I \) with dimensions \( W \times H \), the aspect ratio is computed as \( r = \frac{W}{H} \). 
The objective is to resize the image into tiles of size \( S \times S \) (where \( S = 448 \)) while selecting the closest aspect ratio that minimizes distortion.
The number of tiles, \( n_{\text{tiles}} \), is constrained within a predefined range \( [n_{\min}, n_{\max}] \).

To find the optimal aspect ratio for resizing, we define the set of target aspect ratios \( \mathcal{R} \) as:

\vspace{-0.5em}
\begin{equation}
\mathcal{R} = \left\{i / j \mid 1 \leq i, j \leq n, \, i \times j \in [n_{\min}, n_{\max}] \right\}.
\vspace{-0.3em}
\end{equation}

The closest aspect ratio \( r_{\text{best}} \) is selected by minimizing the difference between the original aspect ratio \( r \) and each target aspect ratio \( r_{\text{target}} \):

\vspace{-0.5em}
\begin{equation}
r_{\text{best}} = \arg\min_{r_{\text{target}} \in \mathcal{R}} \left| r - r_{\text{target}} \right|.
\vspace{-0.3em}
\end{equation}

In cases where multiple aspect ratios produce the same difference (\eg, 1:2 and 2:4), we prioritize the aspect ratio that results in an area less than or equal to twice the original image size.
This helps to some extent in preventing the excessive enlargement of low-resolution images.

\textbf{Image Resizing and Splitting.} Once the best aspect ratio is determined, the image is resized to new dimensions \( W_{\text{new}} \times H_{\text{new}} \), where \( i_{\text{best}} \) and \( j_{\text{best}} \) are the factors corresponding to \( r_{\text{best}} \):

\vspace{-0.5em}
\begin{equation}
W_{\text{new}} = S \times i_{\text{best}}, \quad H_{\text{new}} = S \times j_{\text{best}}.
\vspace{-0.3em}
\end{equation}

The image is then split into tiles of size \( S \times S \), with the number of tiles calculated as $n_{\text{tiles}} = i_{\text{best}} \times j_{\text{best}}$. Each tile is cropped from the resized image to ensure consistent size.

\textbf{Thumbnail Generation. }
Optionally, if the number of tiles \( n_{\text{tiles}} > 1 \), the original image \( I \) is resized to a square of dimensions \( S \times S \) to generate an additional thumbnail \( I_{\text{thumb}} \).
This thumbnail is appended to the list of tiles, providing a global view alongside the localized tiles.
In cases where \( n_{\text{tiles}} = 1 \), there is no thumbnail to append, and the mechanism naturally skips this step.

\textbf{Data Formats for Different Data Types.}  
As shown in Figure~\ref{fig:data_format}, the dynamic high-resolution method in InternVL 2.0 and 2.5 extends beyond single-image datasets to also support multi-image and video datasets.

For single-image datasets, the maximum number of tiles \( n_{\text{max}} \) is allocated to a single image, ensuring that it is processed at the highest possible resolution. In this scenario, visual tokens are enclosed within \texttt{<img>} and \texttt{</img>} tags, with no additional auxiliary tags used.

In the case of multi-image datasets, the total number of tiles \( n_{\text{max}} \) is distributed across all images within one sample. Each image is identified by an auxiliary tag like \texttt{Image-1} to clearly label individual images. The images themselves are enclosed within \texttt{<img>} and \texttt{</img>} tags, denoting the start and end of the image data. The number of tiles assigned to each image \( I_i \) is proportional to the total number of images \( N_{\text{image}} \), following the equation:

\vspace{-0.5em}
\begin{equation}
n_{\text{max, i}} = \max \left(1, \left\lfloor \frac{n_{\text{max}}}{N_{\text{image}}} \right\rfloor \right).
\vspace{-0.3em}
\end{equation}

For video data, this approach is simplified by setting \( n_{\text{max}} = 1 \). Each video frame is resized to a fixed resolution of 448$\times$448, eliminating the need for tiling. This is because, during training, a large number of frames (\eg, 32 or 64) are typically extracted from a single video. For our model, even without high-resolution input, this still results in 8,192 or 16,384 visual tokens.
Each video frame, labeled with tags like \texttt{Frame-1}, is enclosed within the \texttt{<img>} and \texttt{</img>} tags, similar to image data.

\subsection{Single Model Training Pipeline}

\begin{figure*}[t!]
    \centering
    \includegraphics[width=0.98\linewidth]{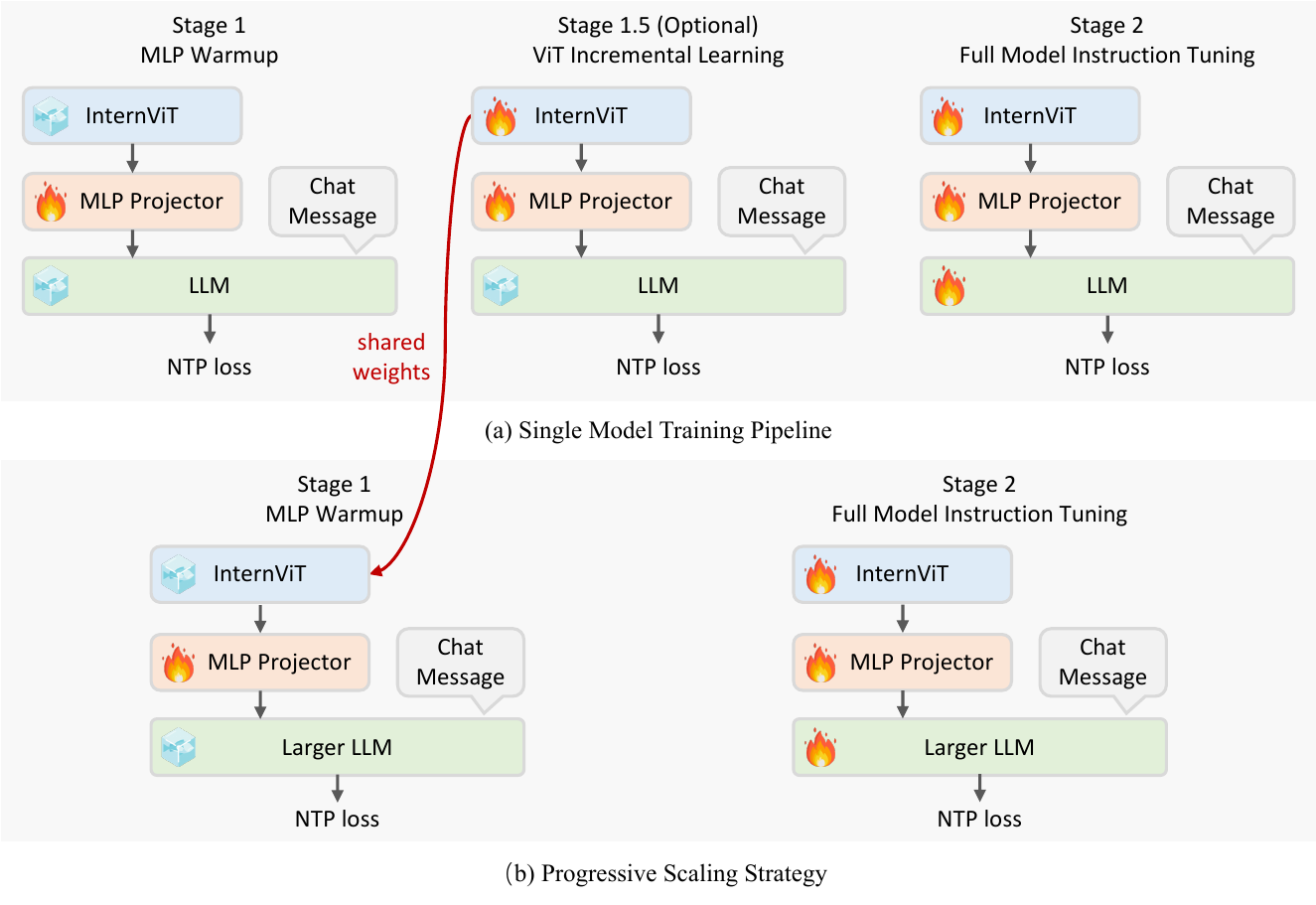}
    \caption{
    \textbf{Illustration of the training pipeline and progressive scaling strategy.}
    (a) Single model training pipeline. The training process is divided into three stages—Stage 1 (MLP warmup), optional Stage 1.5 (ViT incremental learning), and Stage 2 (full model instruction tuning). 
    The multi-stage design progressively enhances vision-language alignment, stabilizes training, and prepares modules for integration with larger LLMs. 
    (b) Progressive scaling strategy. The ViT module trained with a smaller LLM in earlier stages can be easily integrated with larger LLMs, enabling scalable model alignment with affordable resource overhead.
    } 
    \label{fig:training_strategy}
\end{figure*}

\begin{table*}[t!]
\centering
{\fontsize{8}{10}\selectfont 
\newcommand{\Pretrain}{\makecell{Pre-train\\Mixture}}
\newcommand{\Finetune}{\makecell{Fine-tune\\Mixture}}
\renewcommand{\arraystretch}{1.0}
{\setlength\tabcolsep{2.8pt}
\begin{tabular}{l|cc|cc|cc|ccc}
\multirow{2}{*}{Settings}         & \multicolumn{2}{c|}{InternVL2.5-1B} & \multicolumn{2}{c|}{InternVL2.5-2B} & \multicolumn{2}{c|}{InternVL2.5-4B} & \multicolumn{3}{c}{InternVL2.5-8B}       \\
                                  & Stage 1    & Stage 2                & Stage 1          & Stage 2          & Stage 1          & Stage 2          & Stage 1     & Stage 1.5    & Stage 2     \\
\hline
Dataset                           & \Pretrain  & \Finetune              & \Pretrain        & \Finetune        & \Pretrain        & \Finetune        & \Pretrain   & \Pretrain    & \Finetune   \\
Trainable                         & MLP        & Full Model             & MLP              & Full Model       & MLP              & Full Model       & MLP         & ViT+MLP      & Full Model  \\
Packed Batch Size                 & 512        & 512                    & 512              & 512              & 512              & 512              & 512         & 1024         & 512         \\
Learning Rate                     & 2e-4       & 4e-5                   & 2e-4             & 4e-5             & 2e-4             & 4e-5             & 2e-4        & 1e-5         & 4e-5        \\
Context Length                    & 16384      & 16384                  & 16384            & 16384            & 16384            & 16384            & 16384       & 16384        & 16384       \\
Image Tile Threshold              & 48         & 48                     & 48               & 48               & 48               & 48               & 48          & 48           & 48          \\
ViT Drop Path                     & 0.0        & 0.1                    & 0.0              & 0.1              & 0.0              & 0.1              & 0.0         & 0.1          & 0.1         \\
Weight Decay                      & 0.01       & 0.01                   & 0.01             & 0.01             & 0.01             & 0.01             & 0.05        & 0.05         & 0.05        \\
Training Epochs                   & --         & 4                      & --               & 4                & --               & 2                & --          & --           & 1           \\
\hline
Training Tokens                   & $\sim$191B & $\sim$176B             & $\sim$277B       & $\sim$176B       & $\sim$164B        & $\sim$88B        & $\sim$22B   & $\sim$76B    & $\sim$44B  \\
\end{tabular}
}

\bigskip

{
\setlength\tabcolsep{8pt}
\begin{tabular}{l|ccc|cc|cc}
\multirow{2}{*}{Settings}         & \multicolumn{3}{c|}{InternVL2.5-26B}   & \multicolumn{2}{c|}{InternVL2.5-38B} & \multicolumn{2}{c}{InternVL2.5-78B} \\
                                  & Stage 1    & Stage 1.5    & Stage 2    & Stage 1          & Stage 2           & Stage 1          & Stage 2          \\
\hline
Dataset                           & \Pretrain  & \Pretrain    & \Finetune  & \Pretrain        & \Finetune         & \Pretrain        & \Finetune        \\
Trainable                         & MLP        & ViT+MLP      & Full Model & MLP              & Full Model        & MLP              & Full Model       \\
Packed Batch Size                 & 512        & 1024         & 512        & 512              & 512               & 512              & 512              \\
Learning Rate                     & 2e-4       & 1e-5         & 2e-5       & 2e-4             & 2e-5              & 2e-4             & 2e-5             \\
Context Length                    & 16384      & 16384        & 16384      & 16384            & 16384             & 16384            & 16384            \\
Image Tile Threshold              & 48         & 48           & 48         & 48               & 48                & 48               & 48               \\
ViT Drop Path                     & 0.0        & 0.4          & 0.4        & 0.0              & 0.4               & 0.0              & 0.4              \\
Weight Decay                      & 0.05       & 0.05         & 0.05       & 0.05             & 0.05              & 0.05             & 0.05             \\
Training Epochs                   & --         & --           & 1          & --               & 1                 & --               & 1                \\
\hline
Training Tokens                   & $\sim$31B  & $\sim$146B   & $\sim$44B  & $\sim$107B       & $\sim$44B         & $\sim$76B        & $\sim$44B        \\
\end{tabular}
}
}
\caption{\textbf{Training configurations and hyperparameters for InternVL 2.5.} 
This table presents the training setups for various scales of InternVL 2.5 models. 
The configurations are carefully optimized to ensure efficient scaling and performance across different parameter sizes and training stages.
Notably, Qwen2-VL~\cite{wang2024qwen2vl} processed a cumulative total of 1.4T tokens, while our InternVL2.5-78B is trained on just $\sim$120B tokens.
}
\label{tab:train_hyperparameter}
\end{table*}

The training pipeline for a single model in InternVL 2.5 is structured across three stages, designed to enhance the model's visual perception and multimodal capabilities. Each stage progressively integrates vision and language modalities, balancing performance optimization with training efficiency.

\textbf{Stage 1: MLP Warmup.}
As shown in Figure~\ref{fig:training_strategy}(a), the training begins with warming up the MLP projector, which is the initial bridge between visual and language representations. In this stage, only the MLP projector is trained while both the vision encoder (\ie, InternViT~\cite{chen2023internvl}) and language model are frozen. 
To achieve optimal performance, we begin using the dynamic high-resolution training strategy from this stage, even though it increases the training cost.

In this phase, we utilize the pre-training data mixture as outlined in Table~\ref{tab:pretraining_datasets}. The data is formatted in a structured ChatML style and optimized with the NTP loss.
Additionally, a higher learning rate is applied to accelerate convergence, allowing the MLP to quickly adapt to the LLM’s input space and establish robust cross-modal alignment. 
The MLP warmup phase ensures the model is well-prepared to handle multimodal tasks before unlocking additional trainable components in later stages, thereby improving training stability.

\textbf{Stage 1.5: ViT Incremental Learning (Optional).}
As shown in Figure~\ref{fig:training_strategy}(a), Stage 1.5 introduces incremental learning for the vision encoder. During this stage, both the vision encoder and MLP projector are trainable, and training is conducted using the same pre-training data mixture and NTP loss as in Stage 1. 
The aim of this stage is to enhance the vision encoder's ability to extract visual features, allowing it to capture more comprehensive information, especially for domains that are relatively rare in web-scale datasets (\eg, LAION-5B~\cite{schuhmann2022laion5b}), such as multilingual OCR data and mathematical charts, among others.

As shown in Table~\ref{tab:train_hyperparameter}, a lower learning rate is used in this stage to prevent catastrophic forgetting, ensuring the encoder doesn’t lose previously learned capabilities. Additionally, the vision encoder only needs to be trained once unless new domain requirements or data are introduced. Once trained, it can be reused with different LLMs without retraining (see Figure~\ref{fig:training_strategy}(b) and Section \ref{sec:progressive_scaling}), making Stage 1.5 optional. This is particularly beneficial when the encoder has already been optimized for some specific tasks, allowing it to integrate with LLMs of various sizes without significant additional costs.

\textbf{Stage 2: Full Model Instruction Tuning.}
In the final stage, as illustrated in Figure~\ref{fig:training_strategy}(a), the entire model—comprising the ViT, MLP, and LLM—is trained on high-quality multimodal instruction datasets. Data quality is especially important here, as the LLM, responsible for generating the final user-facing output, is now trainable. Even a small amount of noisy data (\eg, a few thousand samples) can lead to abnormal model behavior, like repetitive output or specific erroneous results. To mitigate the degradation of the LLM, we implement strict data quality controls during this stage.

Additionally, the training hyperparameters in this stage are kept simple, with a unified learning rate applied to the entire model rather than different learning rates for various components. 
After completing this stage, InternVL 2.5’s full training process is finished. Although further improvements could be made through Stage 3—post-training with higher-quality data or other training methods (\eg, preference optimization)—we plan to leave this for the future.

\subsection{Progressive Scaling Strategy}
\label{sec:progressive_scaling}

As shown in Figure~\ref{fig:training_strategy}, we propose a progressive scaling strategy to efficiently align the vision encoder (\eg, InternViT) with LLMs. 
We previously adopted similar strategies in the training of InternVL 1.5 and 2.0, but this time we formalized the approach into a clearly defined methodology.
This strategy adopts a staged training approach, \emph{starting with smaller, resource-efficient LLMs and progressively scaling up to larger LLMs}. This approach stems from our observation that \emph{even when the ViT and LLM are jointly trained using NTP loss, the resulting visual features are generalizable representations that can be easily understood by other LLMs}. 

Specifically, in Stage 1.5, the InternViT is trained alongside a smaller LLM (\eg, 20B), focusing on optimizing fundamental visual capabilities and cross-modal alignment. This phase avoids the high computational costs associated with training directly with a large LLM. Using a shared-weight mechanism, the trained InternViT can be easily transferred to a larger LLM (\eg, 72B) without requiring retraining. Consequently, when training a larger model, Stage 1.5 can be skipped (see Table~\ref{tab:train_hyperparameter}), as the optimized InternViT module from earlier stages is reused. This not only accelerates training but also ensures that the vision encoder's learned representations are preserved and effectively integrated into the larger model.

By employing this progressive scaling strategy, we achieve scalable model updates at a fraction of the cost typically associated with large-scale MLLM training. For example, Qwen2-VL~\cite{wang2024qwen2vl} processes a cumulative total of 1.4 trillion tokens, whereas our InternVL2.5-78B is trained on only about 120 billion tokens—\emph{less than one-tenth of Qwen2-VL}. This approach proves particularly advantageous in resource-constrained settings by maximizing the reuse of pre-trained components, minimizing redundant computations, and enabling the efficient training of models capable of addressing complex vision-language tasks.

\subsection{Training Enhancements}
\label{sec:training_enhancements}

To enhance the model’s adaptability to real-world scenarios and overall performance, two key techniques are introduced. These optimizations are essential in improving the user experience and the model’s benchmark performance.

\textbf{Random JPEG Compression.} 
To avoid overfitting during training and enhance the model's real-world performance, we apply a data augmentation technique that preserves spatial information: JPEG compression. Specifically, random JPEG compression with quality levels between 75 and 100 is applied to simulate the degradation commonly found in internet-sourced images. This augmentation improves the model's robustness to noisy, compressed images and enhances the user experience by ensuring more consistent performance across varied image qualities.

\textbf{Loss Reweighting.}
Token averaging and sample averaging are two widely applied strategies for weighting the NTP loss.
Token averaging computes the average NTP loss across all tokens, whereas sample averaging first averages the NTP loss within each sample (across tokens) and then averages across the number of samples. These strategies can be expressed in a unified format:

\vspace{-0.5em}
\begin{equation}
\mathcal{L} = \frac{w_i}{\sum_j w_j} \cdot \mathcal{L}_i, \quad w_i =
\begin{cases}
    \frac{1}{x^0}, & \text{for token averaging} \\
    \frac{1}{x^1}, & \text{for sample averaging},
\end{cases}
\vspace{-0.3em}
\end{equation}

where $\mathcal{L}_i$ and $w_i$ denote the loss and weight of token $i$, respectively, and $x$ denotes the number of tokens in the response to which token $i$ belongs.

When using token averaging, each token contributes equally to the final loss, which can result in gradients biased toward responses with more tokens, leading to a drop in benchmark performance. In contrast, sample averaging ensures that each sample contributes equally, but it can cause the model to favor shorter responses, negatively impacting the user experience. To mitigate bias toward either longer or shorter responses during training, we apply a reweighting strategy where \( w_i = \frac{1}{x^{0.5}} \). This approach, named \textbf{square averaging}, balances the contribution of responses with different lengths.

\begin{figure*}[t!]
    \centering
    \includegraphics[width=1\linewidth]{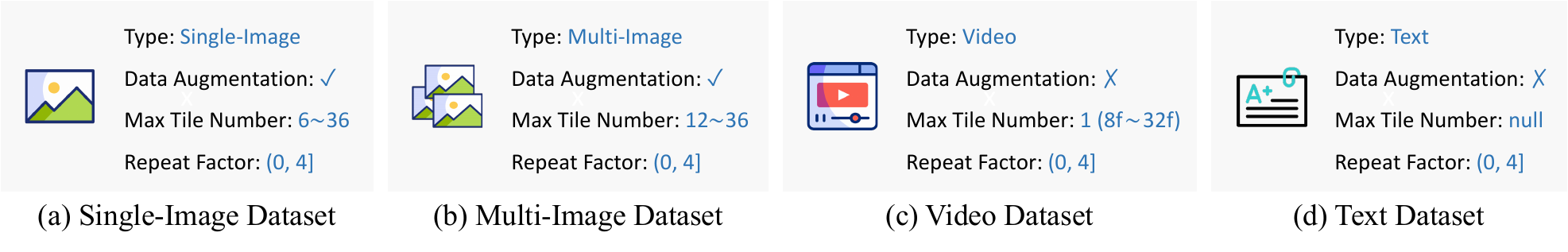}
    \caption{\textbf{Dataset configuration.}
    In InternVL 2.0 and 2.5, data augmentation is applied selectively, enabled for image datasets and disabled for videos and text. The maximum tile number (\(n_{\max}\)) controls the resolution of inputs, with higher values for multi-image datasets and lower values for videos. The repeat factor (\(r\)) balances dataset sampling by adjusting the frequency of each dataset, ensuring robust and balanced training.
    }
    \label{fig:dataset_configuration}
\end{figure*}

\section{Data Organization}

\subsection{Dataset Configuration}

In InternVL 2.0 and 2.5, the organization of the training data is controlled by several key parameters to optimize the balance and distribution of datasets during training, as shown in Figure~\ref{fig:dataset_configuration}. 

\textbf{Data Augmentation.}
Firstly, data augmentation (\ie, JPEG compression introduced in Section~\ref{sec:training_enhancements}) is applied conditionally, allowing for enhanced robustness by enabling or disabling augmentation techniques based on dataset characteristics. 
Specifically, we enable this augmentation for all image datasets, while disabling it for all video datasets, to ensure that different video frames have the same image quality.

\textbf{Maximum Tile Number.}  
The parameter \( n_{\max} \) defines the maximum number of tiles allowed per dataset, effectively controlling the resolution of the image or video frame fed into the model. This ensures flexibility in handling datasets of varying complexity and type. For example, we can set \( n_{\max} = 24 \) or \( 36 \) for multi-image datasets, high-resolution documents, or infographics, use \( n_{\max} = 6 \) or \( 12 \) for most other low-resolution image datasets, and set \( n_{\max} = 1 \) for video datasets. This adjustment was first introduced in InternVL 2.0, whereas in InternVL 1.5, a uniform value of \( n_{\max} = 12 \) was applied across all datasets.

\textbf{Repeat Factor.}
Finally, the repeat factor \( r \) determines the sampling frequency of each dataset. With \( r \in (0, 4] \), this parameter enables down-sampling when \( r < 1 \), reducing the dataset's weight during training, or up-sampling when \( r > 1 \), effectively increasing the number of epochs for that dataset. This mechanism finely adjusts the relative proportions of datasets, ensuring a balanced distribution across training data. By adjusting \( r \), especially in multi-task learning, the data from each domain or task receives appropriate training, preventing overfitting or underfitting of any single dataset, leading to more balanced model performance. 

\subsection{Multimodal Data Packing}
In InternVL 2.0 and 2.5, we implement a data-packing strategy to enhance GPU utilization and improve training efficiency. This approach reduces padding by concatenating multiple samples into longer sequences, thereby maximizing the utilization of the model's input sequence capacity.
Specifically, for multimodal models like InternVL, data packing should account for two dimensions:  
(a) \emph{{Sequence length for the LLM}}, which corresponds to the standard input sequence length used in language models. This remains essential in multimodal tasks;
(b) \emph{{Image tile number for the ViT}}, which denotes the number of image tiles processed by the vision encoder. Efficient management of this dimension is crucial for optimizing training efficiency.

To handle these dimensions efficiently, our data-packing strategy comprises the following steps:

(1) \textbf{{Select}}: During the selection phase, the algorithm operates similarly to a standard dataset without data-packing, directly sampling independent data. Each sampled item is truncated into multiple smaller items and treated as separate samples. This ensures that the sequence length and image tile count of each sample are within the predefined thresholds $l_{\rm max}$ (context length) and $t_{\rm max}$ (image tile limit), respectively.

(2) \textbf{{Search}}: For a given independent sample, the algorithm searches for another sample from the buffer list to pack them together. The resulting packed sample must have a sequence length shorter than $l_{max}$ and include fewer than $t_{max}$ image tiles. If multiple buffers satisfy these requirements, the one with the longest sequence length and the maximum number of image tiles is selected. In practice, the buffer list is maintained in descending order and a binary search is performed to accelerate the search process.

(3) \textbf{{Pack}}: The sampled data and the selected buffer are packed into a single sequence. If no buffer is selected in the previous step, the sample remains unchanged and proceeds directly to the next phase. Notably, tokens in the packed data can only attend to the context within their respective samples and cannot attend to tokens from other packed samples. Furthermore, the positional index of each sample is maintained independently.

(4) \textbf{{Maintain}}: In the maintenance phase, if a packed sample exceeds \( l_{\text{max}} \) or contains more than \( t_{\text{max}} \) image tiles, it is immediately yielded for training. Otherwise, the packed sample is inserted into the buffer list. If the buffer list exceeds its capacity, the sample with the longest sequence length and the highest number of image tiles is yielded to maintain buffer efficiency.

\begin{figure*}[t!]
    \centering
    \includegraphics[width=1\linewidth]{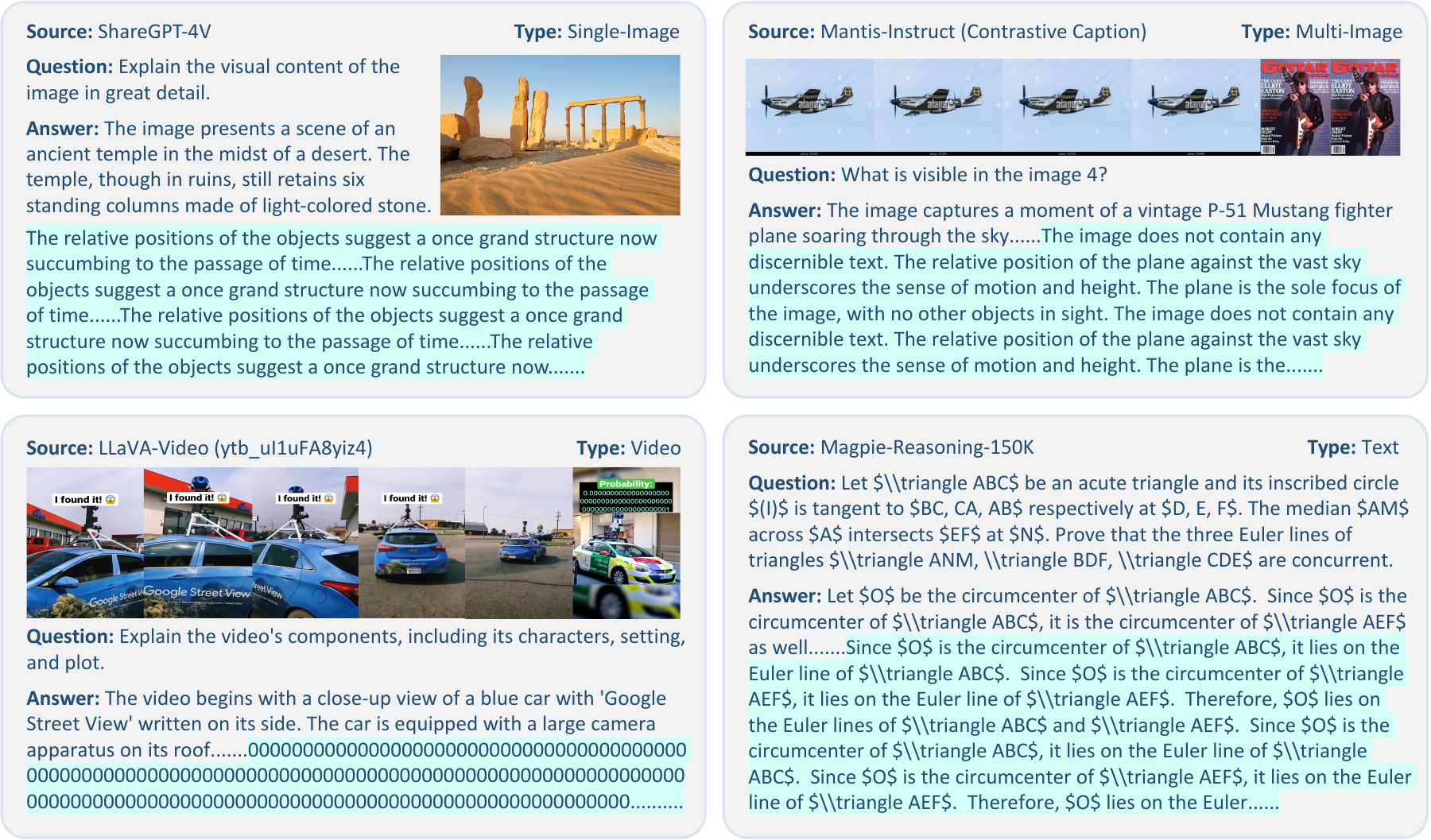}
    \caption{\textbf{Visualization of abnormal samples in open-source datasets.} 
    Abnormal samples are prevalent across various data types, including single-image, multi-image, video, and pure text datasets, with ``repetitive patterns'' being a prominent issue. We identify this as one of the most detrimental problems for test-time scaling, often leading models into loops in long-form outputs and CoT reasoning tasks. Thoroughly filtering the fine-tuning data mixture can mitigate this issue to some extent.}
    \label{fig:open_dataset_abnormal}
\end{figure*}

\subsection{Data Filtering Pipeline}

During model development, we observed that LLMs are significantly more sensitive to data noise than vision encoders.
As shown in Figure~\ref{fig:training_strategy}, during Stage 2, when all model weights are fully trainable, even a small fraction of anomalous samples—such as outliers or repetitive data, numbering only a few thousand—can lead to aberrant model behavior during inference. While conventional wisdom assumes that minor noise in large-scale datasets can be ignored, our findings indicate otherwise: even a tiny fraction of noisy samples can degrade MLLM performance and user experience.

Among these anomalies, we identify \emph{repetitive generation as one of the most detrimental issues}. In many open-source or synthetic datasets, a small number of samples with repetitive patterns—comprising merely thousands of examples in our fine-tuning data mixture—can cause the model to spiral into repetitive loops, particularly in long-form outputs or CoT reasoning tasks. This phenomenon undermines the effectiveness of test-time scaling strategies. To address this challenge and support future research, we designed an efficient data filtering pipeline to remove low-quality samples, thereby minimizing the risk of repetitive generation.

\begin{figure*}[t!]
    \centering
    \includegraphics[width=1\linewidth]{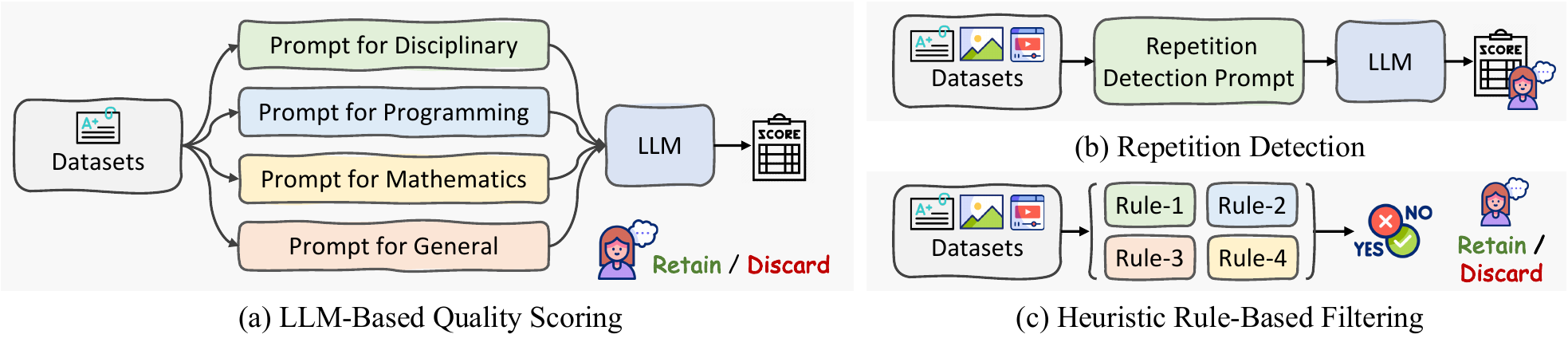}
    \caption{\textbf{Dataset filtering pipeline.}
    For text data, we use three methods: (a) LLM-based quality scoring to assign domain-specific quality scores and filter low-quality samples; (b) Repetition detection to identify and remove data with repetitive patterns; and (c) heuristic rule-based filtering to detect anomalies using predefined rules. 
    For multimodal data, only (b) repetition detection and (c) heuristic rule-based filtering are applied to mitigate repetitive patterns and ensure dataset integrity.
    }
    \label{fig:dataset_filtering_pipeline}
\end{figure*}

As shown in Figure~\ref{fig:dataset_filtering_pipeline}, our data filtering pipeline consists of two modules. For pure-text data, we implemented three key strategies:
(1) \emph{LLM-Based Quality Scoring}: We begin by categorizing datasets into distinct domains (\eg, disciplinary, programming, mathematics, general). Next, we assign a quality score, ranging from 0 to 10, to each sample using a pre-trained LLM~\cite{qwen2.5} with a domain-specific prompt. Samples with scores below a specified threshold (\eg, 7) are then removed to ensure data quality.
(2) \emph{Repetition Detection}: We use an LLM combined with a specialized prompt to identify repetitive patterns. These samples are then subjected to manual review, and those scoring below a threshold (\eg, 3) are removed to maintain data quality.
(3) \emph{Heuristic Rule-Based Filtering}: We apply specific rules, such as filtering out sentences with abnormal lengths, excessively long sequences of zeros, text with an excessive number of duplicate lines, \etc, to identify anomalies in the data. Although this approach may occasionally produce false positives, it improves the detection of anomalous samples. All flagged samples are manually reviewed before final removal.

For multimodal data, given the limitations of open-source MLLMs in scoring such data, we focused on mitigating repetitive patterns through two strategies:
(1) \emph{Repetition Detection}: We exempted high-quality academic datasets and used a specific prompt to identify repetitive patterns in the remaining data. These samples were removed following the same manual review process we applied to textual data.
(2) \emph{Heuristic Rule-Based Filtering}: Similar heuristic rules are applied, followed by manual verification to ensure dataset integrity.

\emph{This rigorous data-filtering pipeline significantly reduced the occurrence of anomalous behaviors, particularly repetitive generation, with notable improvements in CoT reasoning tasks}. However, we recognize that data filtering alone cannot completely eliminate such issues. 
This may be due to the inherent noise introduced during the LLM's pre-training process, which our multimodal post-training efforts can only partially mitigate without fundamentally resolving the issue of repetitive outputs. Future work will explore preference optimization and other strategies to further suppress anomalies and enhance both model performance and user experience.

\begin{figure*}[t!]
    \centering
    \includegraphics[width=0.98\linewidth]{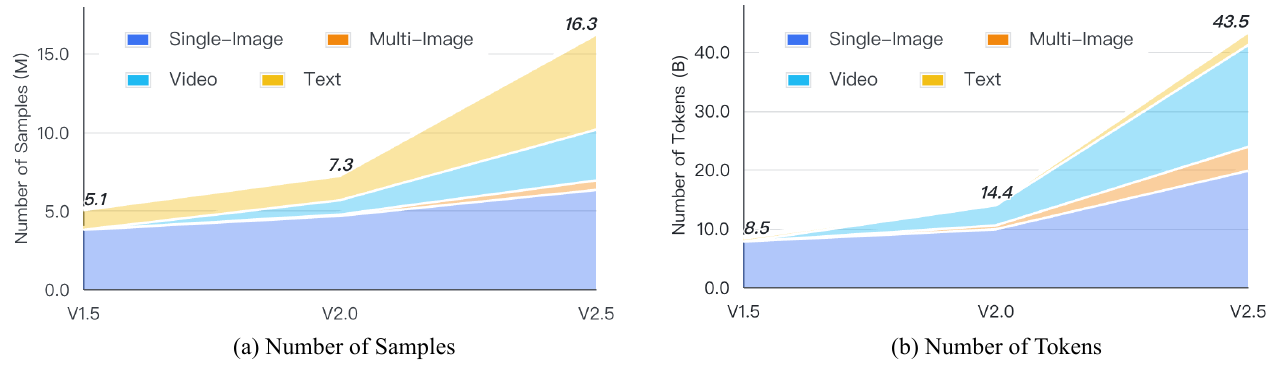}
    \caption{
    \textbf{Statistics of the fine-tuning data mixture.}
     The dataset shows consistent growth from InternVL 1.5 to 2.5 in terms of (a) the number of samples and (b) the number of tokens across multiple dataset types, including single-image, multi-image, video, and text. 
     Note that the token count here refers to the total number of tokens in a specific modality dataset. For example, in the case of single-image datasets, the token count is the sum of the visual tokens and text tokens in these datasets.
     These statistics reflect iterative improvements in data scale and diversity, which enhance the model's multimodal understanding capabilities.
    }
    \label{fig:dataset_analysis}
\end{figure*}

\subsection{Pre-training Data Mixture}

To comprehensively enhance the model's performance and strengthen its ability to handle complex tasks in real-world scenarios, we collect a broader range of domain-specific data compared to the training corpus of InternVL 1.5 and 2.0. As shown in Table~\ref{tab:pretraining_datasets}, our training corpus is sourced from captioning, general QA, mathematics, charts, OCR, knowledge, grounding, documents, conversation, medical, and GUI tasks.

Notably, during the development of our models, we utilized conversation-format instruction data. 
For non-conversational datasets, such as image captioning, OCR, and object detection datasets, we construct questions to transform the data into a conversational format.
At this stage, since only the parameters of MLP (\ie, Stage 1) or MLP and ViT (\ie, Stage 1.5) are trainable, both low-quality and high-quality data are incorporated. 
The goal is to enrich the model's world knowledge as much as possible by exposing it to diverse domain data, thereby improving its generalization capabilities.

In our view, the ideal scenario is for the fine-tuning data mixture to be a subset of the pre-training data mixture. This ensures that the data in this subset can be adequately trained within the vision encoder. However, in practice, due to the high training costs of Stage 1.5, achieving this is often difficult.
Therefore, in the training of InternVL 2.5, only a subset of the datasets from the fine-tuning data mixture was included in the pre-training data mixture.

\begin{table*}[t!]
{\fontsize{8}{10}\selectfont 
    \centering
    \setlength{\tabcolsep}{3mm} 
    \renewcommand{\arraystretch}{0.95}
    \begin{tabular}{l|l}
         Task & Dataset \\
         \hline
\multicolumn{2}{l}{\emph{Type: Single/Multi-Image Datasets}} \\
                            & FaceCaption~\cite{dai202415m},
                            COCO-Caption~\cite{singh2025benchmarking},
                            OpenImages-Caption~\cite{kuznetsova2020openimage},
                            Objects365-Caption~\cite{shao2019objects365},
                            TextCap~\cite{sidorov2020textcaps},
                            \\
                            &Laion-ZH~\cite{schuhmann2022laion5b},
                            Laion-EN~\cite{schuhmann2022laion5b}, 
                            Laion-COCO~\cite{schuhmann2022laioncoco},
                            LLaVAR~\cite{zhang2023llavar}, 
                            InternVL-SA-1B-Caption~\cite{kirillov2023segment}, 
                            \\
\multirow{-2}{*}{Captioning}& 
                            MMInstruct~\cite{liu2024mminstruct},
                            GRIT-Caption~\cite{peng2023kosmos2},
                            ShareGPT4V~\cite{chen2023sharegpt4v},
                            LVIS-Instruct-4V~\cite{wang2023lvisinstruct4v},
                            ShareCaptioner~\cite{chen2023sharegpt4v},
                            \\
                            &
                            OmniCorpus~\cite{li2024omnicorpus},
                            ShareGPT4o~\cite{chen2024far}
                            \\

\rowcolor{gray!15}          
                            & GQA~\cite{hudson2019gqa},
                            OKVQA~\cite{marino2019okvqa},
                            A-OKVQA~\cite{schwenk2022aokvqa},
                            Visual7W~\cite{zhu2016visual7w},
                            VisText~\cite{tang2023vistext},
                            VSR~\cite{liu2023vsr},
                            TallyQA~\cite{acharya2019tallyqa},
                            \\
\rowcolor{gray!15}
\multirow{-2}{*}{General QA}   
                            & 
                            Objects365-YorN~\cite{shao2019objects365},
                            IconQA~\cite{lu2021iconqa},
                            Stanford40~\cite{yao2011human},
                            VisDial~\cite{das2017visdial},
                            VQAv2~\cite{goyal2017vqav2}, 
                            Hateful-Memes~\cite{kiela2020hateful} \\

                            & MAVIS~\cite{zhang2024mavis},
                            GeomVerse~\cite{kazemi2023geomverse},
                            MetaMath-Rendered~\cite{yu2023metamath}, 
                            MapQA~\cite{chang2022mapqa}, 
                            GeoQA+~\cite{cao2022geoqa_plus},
                            Geometry3K~\cite{lu2021geometry3k}, \\
\multirow{-2}{*}{Mathematics}
                            & UniGeo~\cite{chen2022unigeo},
                            GEOS~\cite{seo2015solving},
                            CLEVR-Math~\cite{lindstrom2022clevrmath}  \\

\rowcolor{gray!15}
                            & 
                            ChartQA~\cite{masry2022chartqa}, 
                            PlotQA~\cite{methani2020plotqa},
                            FigureQA~\cite{kahou2017figureqa},
                            LRV-Instruction~\cite{liu2023lrv-instruction}, 
                            ArxivQA~\cite{li2024multimodal},
                            MMC-Inst~\cite{liu2023mmc}, \\
\rowcolor{gray!15}
                            & TabMWP~\cite{lu2022tablemwp},
                            DVQA~\cite{kafle2018dvqa},
                            UniChart~\cite{masry2023unichart}, 
                            SimChart9K~\cite{xia2023structchart},
                            Chart2Text~\cite{obeid2020chart},
                            FinTabNet~\cite{zheng2021global}, \\
\rowcolor{gray!15}
\multirow{-3}{*}{Chart}                            
                            & SciTSR~\cite{chi2019complicated},
                            \textcolor{gray}{Synthetic Chart2Markdown}
                            \\
                            & 
                            LaionCOCO-OCR~\cite{schuhmann2022laioncoco}, 
                            Wukong-OCR~\cite{gu2022wukong},  
                            ParsynthOCR~\cite{hezarai_parsynth_ocr_200k},
                            SynthDoG-EN~\cite{kim2022synthdog},
                            SynthDoG-ZH~\cite{kim2022synthdog},
                            \\
                            & 
                            SynthDoG-RU~\cite{kim2022synthdog},
                            SynthDoG-JP~\cite{kim2022synthdog},
                            SynthDoG-KO~\cite{kim2022synthdog},
                            IAM~\cite{marti2002iam},
                            EST-VQA~\cite{wang2020general},
                            ST-VQA~\cite{biten2019stvqa}, \\
                            &
                            NAF~\cite{davis2019deep}, 
                            InfoVQA~\cite{mathew2022infographicvqa}, 
                            HME100K~\cite{yuan2022syntax},
                            OCRVQA~\cite{mishra2019ocrvqa}, 
                            SROIE~\cite{huang2019icdar2019},
                            POIE~\cite{kuang2023visual},
                            CTW~\cite{yuan2019ctw}, \\
                            & SynthText~\cite{gupta2016synthtext},
                            ArT~\cite{chng2019art}, 
                            LSVT~\cite{sun2019lsvt}, 
                            RCTW-17~\cite{shi2017rctw17}, 
                            ReCTs~\cite{zhang2019rects},
                            MTWI~\cite{he2018icpr2018}, 
                            TextVQA~\cite{singh2019textvqa}, \\
                            & CASIA~\cite{liu2020casia},
                            TextOCR~\cite{singh2021textocr}, 
                            Chinese-OCR~\cite{chinese-ocr}, 
                            EATEN~\cite{guo2019eaten},
                            COCO-Text~\cite{veit2016coco},
                            \textcolor{gray}{Synthetic Arxiv OCR},
                             \\
\multirow{-5}{*}{OCR} 
                            & 
                            \textcolor{gray}{Synthetic Image2Latex}, 
                            \textcolor{gray}{Synthetic Handwritten OCR}, 
                            \textcolor{gray}{Synthetic Infographic2Markdown} 
                            \\

\rowcolor{gray!15}
                            & KVQA~\cite{shah2019kvqa},
                            A-OKVQA~\cite{schwenk2022aokvqa},
                            ViQuAE~\cite{lerner2022viquae}, 
                            iNaturalist2018~\cite{van2018inaturalist}, 
                            MovieNet~\cite{huang2020movienet},
                            ART500K~\cite{mao2017deepart}, 
                            \\      
\rowcolor{gray!15}
                            & 
                            KonIQ-10K~\cite{hosu2020koniq},
                            IconQA~\cite{lu2021iconqa}, 
                            VisualMRC~\cite{tanaka2021visualmrc},
                            ChemVLM Data~\cite{li2024chemvlm},
                            ScienceQA~\cite{lu2022scienceqa},
                            AI2D~\cite{kembhavi2016ai2d}, \\
\rowcolor{gray!15}
\multirow{-3}{*}{Knowledge}
                            &
                            TQA~\cite{kembhavi2017tqa},
                            Wikipedia-QA~\cite{he2023wanjuan}, 
                            \textcolor{gray}{Synthetic Multidisciplinary Knowledge / QA} \\
                            & Objects365~\cite{shao2019objects365},
                            GRIT~\cite{you2023ferret}, 
                            RefCOCO~\cite{yu2016refcoco},
                            GPT4Gen-RD-BoxCoT~\cite{chen2023shikra},
                            All-Seeing-V1~\cite{wang2023allseeing}, \\
\multirow{-2}{*}{Grounding}
                            & 
                            All-Seeing-V2~\cite{wang2024allseeingv2},
                            V3Det~\cite{wang2023v3det},
                            TolokaVQA~\cite{ustalov2023toloka} \\

\rowcolor{gray!15}
Document
                            & DocReason25K~\cite{hu2024mplug_docowl_1_5},
                            DocVQA~\cite{mathew2021docvqa},
                            Docmatix~\cite{2024docmatrix}, 
                            \textcolor{gray}{Synthetic Arxiv QA} \\

                            & ALLaVA~\cite{chen2024allava},
                            SVIT~\cite{zhao2023svit},
                            Cambrain-GPT4o~\cite{tong2024cambrian},
                            TextOCR-GPT4V~\cite{textocr_gpt4v_dataset}, 
                            MMDU~\cite{liu2024mmdu}, \\
\multirow{-2}{*}{Conversation}
                            &
                            \textcolor{gray}{Synthetic Real-World Conversations}
                            \\

\rowcolor{gray!15}
                            & PMC-VQA~\cite{zhang2023pmc},
                            VQA-RAD~\cite{lau2018dataset},
                            ImageCLEF~\cite{garcia2015overview},
                            SLAKE~\cite{liu2021slake}, 
                            Medical-Diff-VQA~\cite{hu2023medical},\\
\rowcolor{gray!15}
\multirow{-2}{*}{Medical}
                            & PMC-CaseReport~\cite{pmccase},
                            GMAI-VL (subset)~\cite{li2024gmai}\\

GUI
                            & Screen2Words~\cite{wang2021screen2words},
                            WebSight~\cite{laurenccon2024unlocking} \\

\hline
\multicolumn{2}{l}{\emph{Type: Video Datasets}} \\

\rowcolor{gray!15}
Captioning                  & Mementos~\cite{wang2024mementos},
                            ShareGPT4Video~\cite{chen2024sharegpt4video},
                            VideoGPT+~\cite{Maaz2024VideoGPT+},
                            ShareGPT4o-Video~\cite{chen2024far} \\
                            
General QA                  & VideoChat2-IT~\cite{li2024mvbench},
                            EgoTaskQA~\cite{jia2022egotaskqa},
                            NTU RGB+D~\cite{liu2020ntu},
                            CLEVRER~\cite{yi2019clevrer},
                            STAR~\cite{wu2024star}, 
                            LSMDC~\cite{rohrbach2015dataset} \\
    \end{tabular}
    \caption{\textbf{Summary of the pre-training data mixture of InternVL 2.5.}
    Notably, we exclusively use conversaiton-format instruction data, and at this stage, only the MLP or both MLP and ViT parameters are trainable, allowing the incorporation of both low-quality and high-quality data.
    } 
\label{tab:pretraining_datasets}
}
\end{table*}

\begin{table*}[t!]
\centering
{
\renewcommand{\arraystretch}{0.97}
\fontsize{8}{10}\selectfont 
\setlength\tabcolsep{3.5pt}
\begin{tabular}{l|l}
Task &  Dataset \\
\hline
\multicolumn{2}{l}{\emph{Type: Single-Image Datasets}} \\
                              & TextCaps (en)~\cite{sidorov2020textcaps}, 
                              ShareGPT4o (en \& zh)~\cite{chen2024far}, 
                              InternVL-SA-1B-Caption (en \& zh)~\cite{chen2023internvl}, \\
\multirow{-2}{*}{Captioning}
                              & NewYorkerCaptionContest (en)~\cite{hessel2023androids},
                              MMInstruct (en \& zh)~\cite{liu2024mminstruct}  \\
\rowcolor{gray!15}
                              & VQAv2 (en)~\cite{goyal2017vqav2}, 
                              GQA (en)~\cite{hudson2019gqa}, 
                              OKVQA (en)~\cite{marino2019okvqa}, 
                              Visual7W (en)~\cite{zhu2016visual7w}, 
                              MMInstruct (en \& zh)~\cite{liu2024mminstruct}, \\
\rowcolor{gray!15}
\multirow{-2}{*}{General QA}
                              &  VSR (en)~\cite{liu2023vsr},
                              FSC147 (en)~\cite{ranjan2021learning},
                              Objects365-YorN (en)~\cite{shao2019objects365}, 
                              Hateful-Memes (en)~\cite{kiela2020hateful} \\
                              & GeoQA+ (en)~\cite{cao2022geoqa_plus}, 
                              CLEVR-Math (en)~\cite{lindstrom2022clevrmath}, 
                              Super-CLEVR (en)~\cite{li2023superclevr},
                              MapQA (en)~\cite{chang2022mapqa}, 
                              MAVIS (en)~\cite{zhang2024mavis}, \\ 
                              & Geometry3K (en)~\cite{lu2021geometry3k}, 
                              TallyQA (en)~\cite{acharya2019tallyqa}, 
                              MetaMath (en)~\cite{yu2023metamath}, 
                              GEOS (en)~\cite{seo2015solving}, 
                              UniGeo (en)~\cite{chen2022unigeo}, \\
\multirow{-3}{*}{Mathematics} 
                              & 
                              GeomVerse (en)~\cite{kazemi2023geomverse}, 
                              CMM-Math (zh)~\cite{liu2024cmmmath}  \\
                              
\rowcolor{gray!15}
                              & ChartQA (en)~\cite{masry2022chartqa}, 
                              MMTab (en)~\cite{zheng2024multimodal}, 
                              PlotQA (en)~\cite{methani2020plotqa}, 
                              FigureQA (en)~\cite{kahou2017figureqa}, 
                              VisText (en)~\cite{tang2023vistext},  \\
\rowcolor{gray!15}
                              & LRV-Instruction (en)~\cite{liu2023lrv-instruction}, 
                              ArxivQA (en)~\cite{li2024multimodal}, 
                              TabMWP (en)~\cite{lu2022tablemwp}, 
                              MMC-Inst (en)~\cite{liu2023mmc},
                              DVQA (en)~\cite{kafle2018dvqa},\\
\rowcolor{gray!15}
                              & UniChart (en)~\cite{masry2023unichart}, 
                              SimChart9K (en)~\cite{xia2023structchart}, 
                              Chart2Text (en)~\cite{obeid2020chart}, 
                              FinTabNet (zh)~\cite{zheng2021global},
                              SciTSR (zh)~\cite{chi2019complicated}, \\
\rowcolor{gray!15}
\multirow{-4}{*}{Chart}
                              & \textcolor{gray}{Synthetic Chart2Markdown (en)}  \\
                              & 
                              OCRVQA (en)~\cite{mishra2019ocrvqa}, 
                              InfoVQA (en)~\cite{mathew2022infographicvqa}, 
                              TextVQA (en)~\cite{singh2019textvqa}, 
                              ArT (en \& zh)~\cite{chng2019art}, 
                              HME100K (en)~\cite{yuan2022syntax},  \\ 
                              & COCO-Text (en)~\cite{veit2016coco}, 
                              CTW (zh)~\cite{yuan2019ctw}, 
                              LSVT (zh)~\cite{sun2019lsvt}, 
                              RCTW-17 (zh)~\cite{shi2017rctw17}, 
                              VCR (en \& zh)~\cite{zhang2024vcr}, \\ 
                              & EST-VQA (en \& zh)~\cite{wang2020general}, 
                              ST-VQA (en)~\cite{biten2019stvqa}, 
                              EATEN (zh)~\cite{guo2019eaten}, 
                              LLaVAR (en)~\cite{zhang2023llavar}, 
                              CASIA (zh)~\cite{liu2020casia},  \\
                              & Chinese-OCR (zh)~\cite{chinese-ocr}, 
                              CyrillicHandwriting (ru)~\cite{cyrillic}, 
                              IAM (en)~\cite{marti2002iam}, 
                              NAF (en)~\cite{davis2019deep}, 
                              POIE (en)~\cite{kuang2023visual}, \\
                              & ReCTs (zh)~\cite{zhang2019rects}, 
                              MTWI (zh)~\cite{he2018icpr2018}, 
                              TextOCR (en)~\cite{singh2021textocr}, 
                              SROIE (en)~\cite{huang2019icdar2019}, 
                              \textcolor{gray}{Synthetic Arxiv OCR (en)}, \\
                              & MTVQA (ko \& ja \& it \& ru \& de \& fr \& th \& ar \& vi)~\cite{tang2024mtvqa}, 
                              \textcolor{gray}{Synthetic Image2Latex (en)}, 
                              \\ 
\multirow{-7}{*}{OCR}         & \textcolor{gray}{Synthetic Handwritten OCR (zh)}, 
                              \textcolor{gray}{Synthetic Infographic2Markdown (en \& zh)} \\
\rowcolor{gray!15}
                              & KVQA (en)~\cite{shah2019kvqa}, 
                              A-OKVQA (en)~\cite{schwenk2022aokvqa}, 
                              ViQuAE (en)~\cite{lerner2022viquae}, 
                              iNaturalist2018 (en)~\cite{van2018inaturalist},
                              MovieNet (en)~\cite{huang2020movienet}, \\
\rowcolor{gray!15}
\multirow{-2}{*}{Knowledge} 
                              & ART500K (en)~\cite{mao2017deepart}, 
                              KonIQ-10K (en)~\cite{hosu2020koniq},
                              \textcolor{gray}{Synthetic Multidisciplinary Knowledge / QA (en \& zh)} \\
Document
                              & DocVQA (en)~\cite{clark2017docqa}, 
                              Docmatix (en)~\cite{2024docmatrix}, 
                              DocReason25K (en)~\cite{hu2024mplug_docowl_1_5}, 
                              Sujet-Finance-QA-Vision (en)~\cite{sujet-finance} \\
\rowcolor{gray!15}
                              & RefCOCO/+/g (en)~\cite{yu2016refcoco,mao2016generation}, 
                              GPT4Gen-RD-BoxCoT (en)~\cite{chen2023shikra}, 
                              All-Seeing-V2 (en)~\cite{wang2024allseeingv2}, \\
\rowcolor{gray!15}
\multirow{-2}{*}{Grounding}   
                              & V3Det (en \& zh)~\cite{wang2023v3det}, 
                              DsLMF (en)~\cite{yang2023open}, 
                              COCO-ReM (en \& zh)~\cite{singh2025benchmarking}, 
                              TolokaVQA (en)~\cite{ustalov2023toloka} \\
\multirow{-1}{*}{Science}     & AI2D (en)~\cite{kembhavi2016ai2d}, 
                              ScienceQA (en)~\cite{lu2022scienceqa}, 
                              TQA (en)~\cite{kembhavi2017tqa}, 
                              ChemVLM Data (en \& zh)~\cite{li2024chemvlm} \\
\rowcolor{gray!15}
                              & ALLaVA (en \& zh)~\cite{chen2024allava}, 
                              Viet-ShareGPT4o (vi)~\cite{doan2024vintern}, 
                              Cambrain-GPT4o (en)~\cite{tong2024cambrian} , 
                              RLAIF-V (en)~\cite{yu2024rlaifv}, \\ 
\rowcolor{gray!15}
                              & Laion-GPT4V (en)~\cite{laion_gpt4v_dataset}, 
                              TextOCR-GPT4V (en)~\cite{textocr_gpt4v_dataset}, 
                              WildVision-GPT4o (en)~\cite{lu2024wildvision}, \\
\rowcolor{gray!15}
\multirow{-3}{*}{Conversation}& \textcolor{gray}{Synthetic Real-World Conversations (en \& zh)}  \\
                              & PMC-VQA (en)~\cite{zhang2023pmc}, 
                              VQA-RAD (en)~\cite{lau2018dataset}, 
                              ImageCLEF (en)~\cite{garcia2015overview}, 
                              PMC (en)~\cite{wu2023towards}, 
                              SLAKE (en \& zh)~\cite{liu2021slake}, \\
                              & GMAI-VL (en \& zh)~\cite{li2024gmai},
                              VQA-Med (en)~\cite{ben2019vqa},
                              Medical-Diff-VQA (en)~\cite{hu2023medical},
                              PathVQA (en)~\cite{he2020pathvqa}, \\
\multirow{-3}{*}{Medical}
                              & PMC-CaseReport (en)~\cite{pmccase}\\
\rowcolor{gray!15}
                             & Screen2Words (en)~\cite{wang2021screen2words}, 
                              WebSight (en)~\cite{laurenccon2024unlocking},
                              Widget-Caption (en)~\cite{li2020widget},
                              RICOSCA (en)~\cite{deka2017rico}, \\
\rowcolor{gray!15}
                              & Seeclick (en)~\cite{cheng2024seeclick},
                              ScreenQA (en)~\cite{hsiao2022screenqa},
                              AMEX (en)~\cite{chai2024amex},
                              AITW (en)~\cite{rawles2024androidinthewild},
                              Odyssey (en)~\cite{lu2024gui}, \\
\rowcolor{gray!15}
\multirow{-3}{*}{GUI}
                              & UIBert (en)~\cite{bai2021uibert},
                              AndroidControl (en)~\cite{lieffects},
                              Mind2Web (en)~\cite{deng2024mind2web},
                              OmniACT (en)~\cite{kapoor2025omniact}, 
                              WaveUI (en)~\cite{agentsea_wave_ui} \\
\hline
\multicolumn{2}{l}{\emph{Type: Multi-Image Datasets}}    \\
                              & Img-Diff (en)~\cite{jiao2024img}, 
                              Birds-to-Words (en)~\cite{jiang2024mantis}, 
                              Spot-the-Diff (en)~\cite{jiang2024mantis}, 
                              MultiVQA (en)~\cite{jiang2024mantis}, 
                              NLVR2 (en)~\cite{suhr2018corpus}, \\ 
\multirow{-2}{*}{General QA}   
                              & ContrastiveCaption (en)~\cite{jiang2024mantis}, 
                              DreamSim (en)~\cite{jiang2024mantis}, 
                              InternVL-SA-1B-Caption (en \& zh)~\cite{chen2023internvl} \\ 
\rowcolor{gray!15}
Document   
                              & MP-DocVQA (en)~\cite{tito2023hier}, 
                              MP-Docmatix (en)~\cite{2024docmatrix} \\
\hline
\multicolumn{2}{l}{\emph{Type: Video Datasets}}    \\
                              & Vript (en \& zh)~\cite{yang2024vript},
                              OpenVid (en)~\cite{nan2024openvid},
                              Mementos (en)~\cite{wang2024mementos}, 
                              ShareGPT4o-Video (en \& zh)~\cite{chen2024far}, \\ 
\multirow{-2}{*}{Captioning} 
                              & 
                              ShareGPT4Video (en \& zh)~\cite{chen2024sharegpt4video},
                              VideoGPT+ (en)~\cite{Maaz2024VideoGPT+} \\
\rowcolor{gray!15}
                              & VideoChat2-IT (en \& zh)~\cite{li2023videochat,li2024mvbench}, 
                              EgoTaskQA (en)~\cite{jia2022egotaskqa}, 
                              NTU RGB+D (en)~\cite{liu2020ntu},
                              CLEVRER (en)~\cite{yi2019clevrer}, \\                              
\rowcolor{gray!15}
                              & LLaVA-Video (en)~\cite{zhang2024video}, 
                              FineVideo (en)~\cite{FineVideo}, 
                              PerceptionTest (en)~\cite{patraucean2024perception},
                              HiREST (en)~\cite{zala2023hierarchical},
                              STAR (en)~\cite{wu2024star},
                              \\
\rowcolor{gray!15}
\multirow{-3}{*}{General QA} 
                             & EgoSchema (en)~\cite{mangalam2023egoschema},
                             ScanQA (en)~\cite{azuma2022scanqa},
                             LSMDC (en)~\cite{rohrbach2015dataset}  \\ 
GUI 
                             & GUI-World (en)~\cite{chen2024gui}  \\

\hline
\multicolumn{2}{l}{\emph{Type: Text Datasets}}    \\

                              & UltraFeedback (en)~\cite{cui2023ultrafeedback}, 
                              UltraChat (en)~\cite{ding2023enhancing}, 
                              Unnatural-Instructions (en)~\cite{honovich2022unnatural}, 
                              NoRobots (en)~\cite{no_robots}, \\
                              & MOSS (en)~\cite{sun2023moss}, 
                              LIMA (en)~\cite{zhou2024lima},  
                              SlimOrca (en)~\cite{SlimOrca}, 
                              WizardLM-Evol-Instruct-70K (en)~\cite{xu2024wizardlm}, \\
                              & Llama-3-Magpie-Pro (en)~\cite{xu2024magpie}, 
                              Magpie-Qwen2-Pro (en \& zh)~\cite{xu2024magpie}, 
                              KOpen-HQ-Hermes-2.5-60K (ko)~\cite{MarkrAI_KOpen_HQ_Hermes_2.5_60K},  \\
                              & Firefly (zh)~\cite{Firefly}, 
                              Dolly (en)~\cite{conover2023free}, 
                              OpenAI-Summarize-TLDR (en)~\cite{CarperAI_openai_summarize_tldr},
                              Know-Saraswati-CoT (en)~\cite{knowrohit07_know_saraswati_cot}, \\
\multirow{-5}{*}{General QA} 
                              &  
                              FLAN (en)~\cite{wei2021finetuned}, 
                              FLANv2 (en \& zh)~\cite{chung2024scaling} \\

\rowcolor{gray!15}
                              & Code-Feedback (en)~\cite{opencodeinterpreter}, 
                              Glaive-Code-Assistant (en)~\cite{glaive_code_assistant_v3}, 
                              XCoder-80K (en)~\cite{wang2024xcoder80k},
                              LeetCode (en \& zh), \\
\rowcolor{gray!15}
\multirow{-2}{*}{Code}        
                              & Evol-Instruct-Code (en)~\cite{luo2023wizardcoder}, 
                              InternLM2-Code (en \& zh)~\cite{cai2024internlm2} \\

\multirow{2}{*}{Long Context}
                              & Long-Instruction-with-Paraphrasing (en \& zh)~\cite{yu2023paraphrasing}, 
                              LongCite (en \& zh)~\cite{zhang2024longcite}, 
                              LongQLoRA (en)~\cite{yang2023longqlora}, \\ & 
                              LongAlpaca (en)~\cite{long-alpaca}  \\

\rowcolor{gray!15}
                              & GSM8K-Socratic (en)~\cite{cobbe2021training}, 
                              NuminaMath-CoT/TIR (en)~\cite{li2024numinamath}, 
                              Orca-Math (en)~\cite{mitra2024orcamath}, 
                              MathQA (en)~\cite{amini2019mathqa},  \\
\rowcolor{gray!15}
\multirow{-2}{*}{Mathematics}
                              & InfinityMATH (en)~\cite{zhang2024inifinitymath},
                              InternLM2-Math (en \& zh)~\cite{cai2024internlm2} \\  

Knowledge
                              & \textcolor{gray}{Synthetic Multidisciplinary Knowledge / QA (en)} \\

\end{tabular}
\caption{
\textbf{Summary of the fine-tuning data mixture of InternVL 2.5.}
We expanded our fine-tuning data mixture through extensive collection of open-source datasets and self-synthesized data. This mixture is predominantly in English (en) and Chinese (zh), with smaller portions in other languages, including Korean (ko), Japanese (ja), Italian (it), Russian (ru), German (de), French (fr), Thai (th), Arabic (ar), and Vietnamese (vi).
}
\label{tab:finetuning_datasets}
}
\end{table*}

\subsection{Fine-tuning Data Mixture}

\label{sec:sft_data}

As shown in Figure~\ref{fig:dataset_analysis}, from InternVL 1.5 to 2.0 and then to 2.5, the dataset has undergone iterative improvements in scale, quality, and diversity.
In terms of data scale, the number of samples grows from 5.1M in InternVL 1.5 to 7.3M in InternVL 2.0, and further doubles to 16.3M in InternVL 2.5. 
For diversity, our training data spans multiple domains, including general QA, charts, documents, OCR, science, medical, GUI, code, mathematics, \etal, while covering multiple modalities such as single-image, multi-image, video, and text.

In InternVL 2.5, single-image data constituted the majority with 45.92\% of tokens, while multi-image data accounted for 9.37\%, video data contributed 39.79\%, and pure-text data made up 4.92\%. Compared to earlier versions, multi-image and video data achieved the most notable increases, leading to the enhanced multi-image and long video comprehension abilities of InternVL 2.5.
Quality improvements were achieved through unifying conversation templates, using language models to score and refine data, removing repetitive patterns, applying heuristic rules to filter low-quality samples, and rewriting short responses into high-quality and longer interactions. This ensured a robust dataset for model training.

\section{Evaluation on Multimodal Capability}

To comprehensively evaluate InternVL's performance on multimodal tasks, we employ a diverse set of benchmarks, including both well-established classic datasets and newly introduced ones provided by VLMEvalKit~\cite{duan2024vlmevalkit}. These benchmarks span a wide range of categories, aiming to provide a thorough and balanced assessment of InternVL's capabilities across various multimodal tasks.

\subsection{Multimodal Reasoning and Mathematics}

\begin{table*}[t!]
\centering
{\fontsize{8}{10}\selectfont 
\renewcommand{\arraystretch}{0.94}
\setlength\tabcolsep{3.9pt}
\newcommand{\MMMU}{\makecell{MMMU\\(val)}}
\newcommand{\MMMUT}{\makecell{MMMU\\(test)}}
\newcommand{\MMMUPRO}{\makecell{MMMU-Pro\\(std10 / vision / overall)}}
\newcommand{\MathVista}{\makecell{MathVista\\(mini)}}
\newcommand{\MathVision}{\makecell{MATH-Vision\\(mini / full)}}
\newcommand{\MathVe}{\makecell{MathVerse\\(mini)}}
\newcommand{\MMMATH}{\makecell{MM-MATH\\(mini)}}
\newcommand{\Olympiad}{\makecell{Olympiad\\Bench}}
\newcommand{\TODO}{\textcolor{red}{TODO}}
\newcommand{\lsp}{--\ \ \ \ \ \ }
\newcommand{\rsp}{\ \ \ \ \ --~}
\begin{tabular}{l|ccc|cccc}
Model Name                                 & \MMMU      & \MMMUT & \MMMUPRO           & \MathVista & \MathVision & \MathVe & \Olympiad \\
\hline
LLaVA-OneVision-0.5B~\cite{li2024llavaov}  & 31.4       & --     & --                 & 34.8       & --          & 17.9    & --        \\
InternVL2-1B~\cite{chen2024far}            & 36.7       & 32.8   & 16.0 / 13.6 / 14.8 & 37.7       & 12.2 / 11.1 & 18.4    & 0.3       \\
\rowcolor{gray!15}
InternVL2.5-1B                             & 40.9       & 35.8   & 23.3 / 15.5 / 19.4 & 43.2       & 16.8 / 14.4 & 28.0    & 1.7       \\
Qwen2-VL-2B~\cite{wang2024qwen2vl}         & 41.1       & --     & 25.3 / 17.2 / 21.2 & 43.0       & 19.7 / 12.4 & 21.0    & --        \\
Aquila-VL-2B~\cite{gu2024aquilavl}         & 47.4       & --     & --                 & 59.0       & 21.1 / 18.4 & 26.2    & --        \\
InternVL2-2B~\cite{chen2024far}            & 36.3       & 34.7   & 21.6 / 14.9 / 18.2 & 46.3       & 15.8 / 12.1 & 25.3    & 0.4       \\
\rowcolor{gray!15}
InternVL2.5-2B                             & 43.6       & 38.2   & 27.3 / 20.1 / 23.7 & 51.3       & 13.5 / 14.7 & 30.6    & 2.0       \\
\hline
Phi-3.5-Vision-4B~\cite{abdin2024phi3}     & 43.0       & --     & 26.3 / 13.1 / 19.7 & 43.9       & 17.4 / 15.5 & 24.1    & --        \\
InternVL2-4B~\cite{chen2024far}            & 47.9       & 41.4   & 28.2 / 21.3 / 24.7 & 58.6       & 17.8 / 16.5 & 32.0    & 1.1       \\
\rowcolor{gray!15}
InternVL2.5-4B                             & 52.3       & 46.3   & 36.4 / 29.0 / 32.7 & 60.5      & 21.7 / 20.9 & 37.1    & 3.0        \\
\hline
Ovis1.6-Gemma2-9B~\cite{lu2024ovis}        & 55.0       & --     & --                 & 67.2       & \rsp / 18.8 & --      & --        \\
MiniCPM-V2.6~\cite{yao2024minicpm}         & 49.8       & --     & 30.2 / 24.2 / 27.2 & 60.6       & 16.1 / 17.5 & 25.7    & --        \\
Qwen2-VL-7B~\cite{wang2024qwen2vl}         & 54.1       & --     & 34.1 / 27.0 / 30.5 & 58.2       & 22.0 / 16.3 & 31.9    & --        \\
InternVL2-8B~\cite{chen2024far}            & 52.6       & 44.3   & 32.5 / 25.4 / 29.0 & 58.3       & 20.4 / 18.4 & 37.0    & 1.9       \\
\rowcolor{gray!15}
InternVL2.5-8B                             & 56.0       & 48.9   & 38.2 / 30.4 / 34.3 & 64.4       & 22.0 / 19.7 & 39.5    & 4.9       \\
\hline
InternVL-Chat-V1.5~\cite{chen2024far}      & 46.8       & 41.0   & 29.5 / 19.9 / 24.7 & 53.5       & 15.8 / 15.0 & 28.4    & 0.6       \\
InternVL2-26B~\cite{chen2024far}           & 51.2       & 43.8   & 32.8 / 27.1 / 30.0 & 59.4       & 23.4 / 17.0 & 31.1    & 3.5       \\
\rowcolor{gray!15}
InternVL2.5-26B                            & 60.0       & 51.8  & 41.6 / 32.6 / 37.1 & 67.7        & 28.0 / 23.1 & 40.1    & 8.8       \\
Cambrian-34B~\cite{tong2024cambrian}       & 49.7       & --     & --                 & 53.2       & --          & --      & --        \\
VILA-1.5-40B~\cite{lin2024vila}            & 55.1       & 46.9   & 35.9 / 14.1 / 25.0 & 49.5       & --          & --      & --        \\
InternVL2-40B~\cite{chen2024far}           & 55.2       & 49.3   & 36.3 / 32.1 / 34.2 & 63.7       & 21.4 / 16.9 & 36.3    & 3.9       \\
\rowcolor{gray!15}
InternVL2.5-38B                            & 63.9       & 57.6   & 48.0 / 44.0 / 46.0 & 71.9       & 32.2 / 31.8 & 49.4    & 12.1      \\
\hline
GPT-4V~\cite{gpt4v}                        & 63.1       & --     & --                 & 58.1       & \rsp / 24.0 & 32.8    & 18.0      \\
GPT-4o-20240513~\cite{gpt4v}               & 69.1       & --     & 54.0 / 49.7 / 51.9 & 63.8       & \rsp / 30.4 & 50.2    & 25.9      \\
Claude-3.5-Sonnet~\cite{claude3series2024} & 68.3       & --     & 55.0 / 48.0 / 51.5 & 67.7       & --          & --      & --        \\
Gemini-1.5-Pro~\cite{reid2024gemini1_5}    & 62.2       & --     & 49.4 / 44.4 / 46.9 & 63.9       & \rsp / 19.2 & --      & --        \\
LLaVA-OneVision-72B~\cite{li2024llavaov}   & 56.8       & --     & 38.0 / 24.0 / 31.0 & 67.5       & --          & 39.1    & --        \\
NVLM-D-72B~\cite{dai2024nvlm}              & 59.7       & 54.6   & --                 & 66.6       & --          & --      & --        \\
Molmo-72B~\cite{deitke2024molmo}           & 54.1       & --     & --                 & 58.6       & --          & --      & --        \\
Qwen2-VL-72B~\cite{wang2024qwen2vl}        & 64.5       & --     & 49.2 / 43.3 / 46.2 & 70.5       & \rsp / 25.9 & --      & 11.2        \\
InternVL2-Llama3-76B~\cite{chen2024far}    & 62.7       & 55.1   & 41.9 / 38.0 / 40.0 & 65.5       & 23.7 / 23.6 & 42.8    & 5.5       \\
\rowcolor{gray!15}
InternVL2.5-78B                            & 70.1       & 61.8   & 51.4 / 45.9 / 48.6 & 72.3       & 34.9 / 32.2 & 51.7    & 11.6      \\
\end{tabular}
}
\caption{\textbf{Comparison of multimodal reasoning and mathematical performance.}
MMMU~\cite{yue2023mmmu} and MMMU-Pro~\cite{yue2024mmmu} are multidisciplinary reasoning benchmarks, while MathVista~\cite{lu2023mathvista}, MATH-Vision~\cite{wang2024measuring}, MathVerse~\cite{zhang2025mathverse}, and OlympiadBench~\cite{he2024olympiadbench} are mathematics benchmarks.
Part of results are collected from \cite{deitke2024molmo, claude3series2024, yue2024mmmu, wang2024measuring, zhang2025mathverse, he2024olympiadbench} and the OpenCompass leaderboard~\cite{opencompass2023}.
}
\label{tab:benchmark_math_reasoning}
\end{table*}

\subsubsection{Benchmarks}

\label{sec:math_reasoning_datasets}

We evaluate InternVL's multimodal mathematical and reasoning capabilities through a comprehensive assessment across various discipline-related benchmarks.

\textbf{MMMU}~\cite{yue2023mmmu}: MMMU is a benchmark evaluating MLLMs on college-level tasks across six disciplines, testing expert-level reasoning and advanced perception in specific fields. 
We report the maximum accuracy achieved across both direct-answer and CoT reasoning approaches on the MMMU validation and test sets.

\textbf{MMMU-Pro}~\cite{yue2024mmmu}: MMMU-Pro is an upgraded version of the MMMU benchmark, designed to more accurately and rigorously evaluate the multimodal understanding and reasoning capabilities of models in a wide range of academic disciplines. We report three metrics: standard (10 options), vision, and overall (the average of standard and vision). Here, ``standard'' and ``vision'' are the maximum scores from the CoT and direct-answer settings, consistent with the original paper.

\textbf{MathVista}~\cite{lu2023mathvista}: MathVista is a benchmark for evaluating MLLMs' mathematical reasoning in visual contexts, encompassing reasoning types such as algebra, geometry, and statistics. We report the scores on the testmini set.

\textbf{MATH-Vision}~\cite{wang2024measuring}: MATH-Vision is a high-quality dataset of 3,040 visually contextualized math problems sourced from real competitions. We report performance on both the testmini and full sets.

\textbf{MathVerse}~\cite{zhang2025mathverse}: MathVerse is a visual math benchmark for evaluating MLLMs in solving diagram-based math problems. It comprises 2,612 high-quality, multi-subject math problems, each transformed into six distinct versions with varying degrees of visual and textual information. We report performance on the testmini set.

\textbf{OlympiadBench}~\cite{he2024olympiadbench}: OlympiadBench is a bilingual, multimodal benchmark with high-difficulty math and physics problems from Olympiad competitions and Gaokao. Each problem is annotated with expert-level step-by-step reasoning, enabling detailed assessment of logical deduction and problem-solving abilities. This benchmark is challenging, and a well-defined CoT prompt can significantly improve performance.

\subsubsection{Evaluation Results}

\begin{figure*}[t!]
    \centering
    \includegraphics[width=1\linewidth]{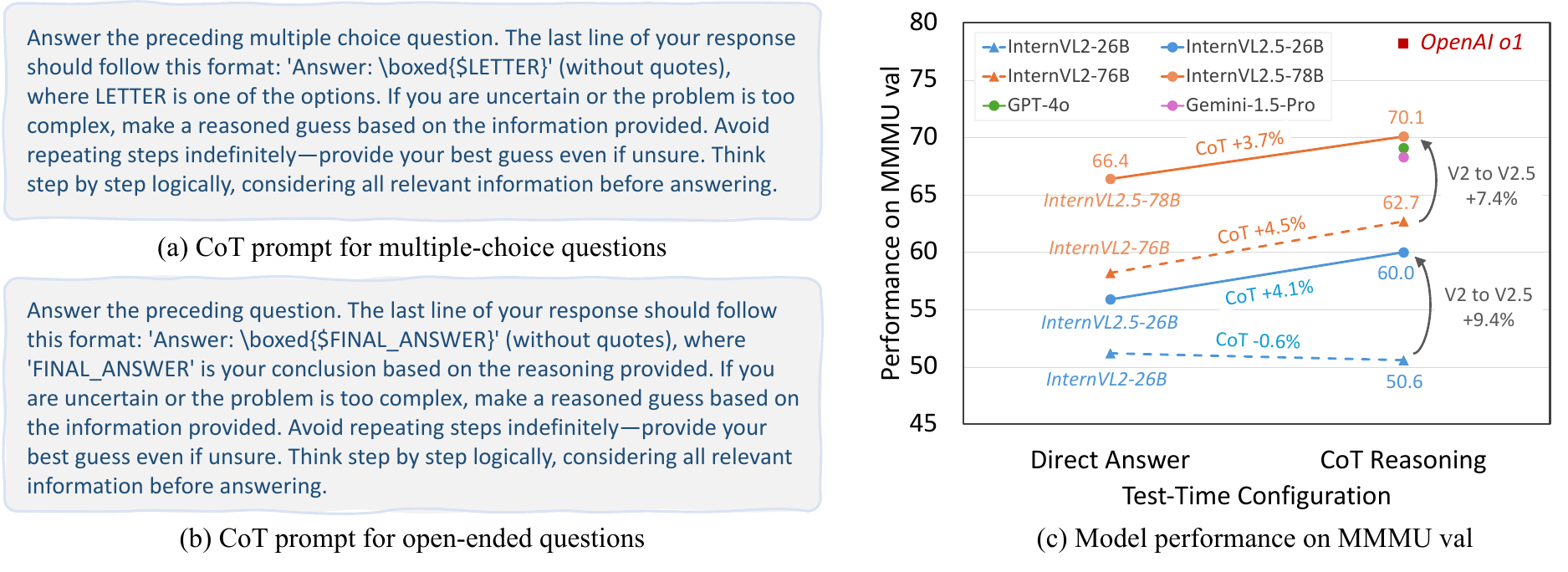}
    \caption{\textbf{CoT prompts used in our model testing.}
    By leveraging these prompts for CoT reasoning, we can scale up testing time, significantly enhancing the performance of  InternVL 2.5 models on MMMU~\cite{yue2023mmmu}.
    }
    \label{fig:cot_prompt}
\end{figure*}

Multidisciplinary reasoning ability reflects a model's capacity to comprehend, process, and manipulate abstract concepts, which is crucial for complex problem-solving and decision-making tasks.
In the left section of Table~\ref{tab:benchmark_math_reasoning}, we provide a comparison of InternVL 2.5's performance on multidisciplinary reasoning-related benchmarks, including MMMU~\cite{yue2023mmmu} and MMMU-Pro~\cite{yue2024mmmu}. 

Here, we test both direct-answer and CoT reasoning performance, reporting the higher score. 
The results suggest that our model achieves encouraging improvements over existing open-source models, such as LLaVA-OneVision~\cite{li2024llavaov}, NVLM~\cite{dai2024nvlm}, VILA 1.5~\cite{lin2024vila}, and Qwen2-VL~\cite{wang2024qwen2vl}, as well as notable progress compared to earlier versions of the InternVL2 series. 
Specifically, InternVL2.5-78B achieves a score exceeding 70 on the MMMU validation set, representing a 7.4-point improvement over InternVL2-Llama3-76B. These results indicate that our model’s performance is moving closer to that of some advanced closed-source models, such as GPT-4o~\cite{gpt4v}, Claude-3.5-Sonnet~\cite{claude3series2024}, and Gemini-1.5-Pro~\cite{reid2024gemini1_5}.
Additionally, through majority voting, the score of InternVL2-Llama3-76B on the MMMU benchmark is improved from 62.7 to 65.3 when using CoT. We observe a similar phenomenon in InternVL 2.5 as well, which demonstrates that test-time scaling can improve the CoT reasoning of MLLMs.

Mathematical reasoning reflects a higher-level reasoning capability and enhances the potential of MLLMs in scientific and engineering applications. In the right-hand section of Table~\ref{tab:benchmark_math_reasoning}, we present InternVL 2.5's performance across four multimodal mathematical benchmarks. These results demonstrate significant progress over InternVL 2.0. Notably, InternVL2.5-78B achieved an accuracy of 72.3\% on the MathVista test-mini set~\cite{lu2023mathvista}. Additionally, on the challenging OlympiadBench~\cite{he2024olympiadbench}, the InternVL 2.5 series showed an overall improvement compared to the 2.0 series. We attribute part of this advancement to our data filtering pipeline. Specifically, we observed that the 2.0 models frequently encountered deadlocks during CoT reasoning, failing to reach correct final answers, while this issue has been mitigated in the 2.5 series.

\subsection{OCR, Chart, and Document Understanding}

\subsubsection{Benchmarks}

\label{sec:ocr_datasets}

\begin{table*}[t!]
\centering
{\fontsize{8}{10}\selectfont 
\renewcommand{\arraystretch}{0.95}
\setlength\tabcolsep{2.7pt}
\newcommand{\TextVQA}{\makecell{TextVQA\\(val)}}
\newcommand{\ChartQA}{\makecell{ChartQA\\(test avg)}}
\newcommand{\DocVQA}{\makecell{DocVQA\\(test)}}
\newcommand{\InfoVQA}{\makecell{InfoVQA\\(test)}}
\newcommand{\TableVQA}{\makecell{TableVQA\\Bench}}
\newcommand{\CharXiv}{\makecell{CharXiv\\(RQ / DQ)}}
\newcommand{\SEEDTP}{\makecell{SEED-2\\Plus}}
\newcommand{\VCREN}{\makecell{VCR-EN-Easy\\(EM / Jaccard)}}
\newcommand{\VCRZH}{\makecell{VCR-ZH\\(EM/Jaccard)}}
\newcommand{\OCRBench}{\makecell{OCR\\Bench}}
\newcommand{\AITD}{\makecell{AI2D\\(w / wo M)}}
\newcommand{\lsp}{--\ \ \ \ \ \ }
\newcommand{\rsp}{\ \ \ \ \ --~}
\newcommand{\TODO}{\textcolor{red}{TODO}}
\newcommand{\newicon}{\includegraphics[height=1em]{figure/new.pdf}}
\begin{tabular}{l|ccccccccc}
Model Name                                 & \AITD       & \ChartQA & \TextVQA & \DocVQA & \InfoVQA & \OCRBench & \SEEDTP & \CharXiv    & \VCREN       \\
\hline
LLaVA-OneVision-0.5B~\cite{li2024llavaov}  & 57.1 / \lsp & 61.4     & --       & 70.0    & 41.8     & 565       & --      & --          & --           \\
InternVL2-1B~\cite{chen2024far}            & 64.1 / 70.5 & 72.9     & 70.5     & 81.7    & 50.9     & 754       & 54.3    & 18.1 / 30.7 & 21.5 / 48.4  \\
\rowcolor{gray!15}
InternVL2.5-1B                             & 69.3 / 77.8 & 75.9     & 72.0     & 84.8    & 56.0     & 785       & 59.0    & 19.0 / 38.4 & 91.5 / 97.0  \\
Qwen2-VL-2B~\cite{wang2024qwen2vl}         & 74.7 / 84.6 & 73.5     & 79.7     & 90.1    & 65.5     & 809       & 62.4    & --          & 81.5 / \lsp  \\
Aquila-VL-2B~\cite{gu2024aquilavl}         & 75.0 / \lsp & 76.5     & 76.4     & 85.0    & 58.3     & 772       & 63.0    & --          & 70.0 / \lsp  \\
InternVL2-2B~\cite{chen2024far}            & 74.1 / 82.3 & 76.2     & 73.4     & 86.9    & 58.9     & 784       & 60.0    & 21.0 / 40.6 & 32.9 / 59.2  \\
\rowcolor{gray!15}
InternVL2.5-2B                             & 74.9 / 83.5 & 79.2     & 74.3     & 88.7    & 60.9     & 804       & 60.9    & 21.3 / 49.7 & 93.2 / 97.6  \\
\hline
Phi-3.5-Vision-4B~\cite{abdin2024phi3}     & 77.8 / 87.6 & 81.8     & 72.0     & 69.3    & 36.6     & 599       & 62.2    & --          & 39.3 / 60.4  \\
InternVL2-4B~\cite{chen2024far}            & 78.9 / 87.8 & 81.5     & 74.4     & 89.2    & 67.0     & 788       & 63.9    & 24.5 / 48.0 & 33.7 / 61.1  \\
\rowcolor{gray!15}
InternVL2.5-4B                             & 81.4 / 90.5 & 84.0     & 76.8     & 91.6    & 72.1     & 828       & 66.9    & 24.9 / 61.7 & 93.7 / 97.8  \\
\hline
Ovis1.6-Gemma2-9B~\cite{lu2024ovis}        & 84.4 / \lsp & --       & --       & --      & --       & 830       & --      & --          & --           \\
MiniCPM-V2.6~\cite{yao2024minicpm}         & 82.1 / \lsp & 82.4     & 80.1     & 90.8    & --       & 852       & 65.7    & 31.0 / 57.1 & 73.9 / 85.7  \\
Molmo-7B-D~\cite{deitke2024molmo}          & \rsp / 93.2 & 84.1     & 81.7     & 92.2    & 72.6     & 694       & --      & --          & --           \\
Qwen2-VL-7B~\cite{wang2024qwen2vl}         & 83.0 / 92.1 & 83.0     & 84.3     & 94.5    & 76.5     & 866       & 69.0    & --          & 89.7 / 93.8  \\
InternVL2-8B~\cite{chen2024far}            & 83.8 / 91.7 & 83.3     & 77.4     & 91.6    & 74.8     & 794       & 67.5    & 31.2 / 56.1 & 37.9 / 61.5  \\
\rowcolor{gray!15} 
InternVL2.5-8B                             & 84.5 / 92.8 & 84.8     & 79.1     & 93.0    & 77.6     & 822       & 69.7    & 32.9 / 68.6 & 92.6 / 97.4  \\
\hline
InternVL-Chat-V1.5~\cite{chen2024far}      & 80.7 / 89.8 & 83.8     & 80.6     & 90.9    & 72.5     & 724       & 66.3    & 29.2 / 58.5 & 14.7 / 51.4  \\
InternVL2-26B~\cite{chen2024far}           & 84.5 / 92.5 & 84.9     & 82.3     & 92.9    & 75.9     & 825       & 67.6    & 33.4 / 62.4 & 74.5 / 86.7  \\
\rowcolor{gray!15}
InternVL2.5-26B                            & 86.4 / 94.4 & 87.2     & 82.4     & 94.0    & 79.8     & 852       & 70.8    & 35.9 / 73.5 & 94.4 / 98.0  \\
Cambrian-34B~\cite{tong2024cambrian}       & 79.5 / \lsp & 75.6     & 76.7     & 75.5    & 46.0     & 600       & --      & 27.3 / 59.7 & 79.7 / 89.3  \\
VILA-1.5-40B~\cite{lin2024vila}            & 69.9 / \lsp & 67.2     & 73.6     & --      & --       & 460       & --      & 24.0 / 38.7 & --           \\
InternVL2-40B~\cite{chen2024far}           & 86.6 / 94.5 & 86.2     & 83.0     & 93.9    & 78.7     & 837       & 69.2    & 32.3 / 66.0 & 84.7 / 92.6  \\
\rowcolor{gray!15}
InternVL2.5-38B                            & 87.6 / 95.1 & 88.2     & 82.7     & 95.3    & 83.6     & 842       & 71.2    & 42.4 / 79.6 & 94.7 / 98.2  \\
\hline
GPT-4V~\cite{gpt4v}                        & 78.2 / 89.4 & 78.5     & 78.0     & 88.4    & 75.1     & 645       & 53.8    & 37.1 / 79.9 & 52.0 / 65.4  \\
GPT-4o-20240513~\cite{gpt4v}               & 84.6 / 94.2 & 85.7     & 77.4     & 92.8    & 79.2     & 736       & 72.0    & 47.1 / 84.5 & 91.6 / 96.4  \\
Claude-3-Opus~\cite{claude3series2024}     & 70.6 / 88.1 & 80.8     & 67.5     & 89.3    & 55.6     & 694       & 44.2    & 30.2 / 71.6 & 62.0 / 77.7  \\
Claude-3.5-Sonnet~\cite{claude3series2024} & 81.2 / 94.7 & 90.8     & 74.1     & 95.2    & 74.3     & 788       & 71.7    & 60.2 / 84.3 & 63.9 / 74.7  \\
Gemini-1.5-Pro~\cite{reid2024gemini1_5}    & 79.1 / 94.4 & 87.2     & 78.8     & 93.1    & 81.0     & 754       & --      & 43.3 / 72.0 & 62.7 / 77.7  \\
LLaVA-OneVision-72B~\cite{li2024llavaov}   & 85.6 / \lsp & 83.7     & 80.5     & 91.3    & 74.9     & 741       & --      & --          & --           \\
NVLM-D-72B~\cite{dai2024nvlm}              & 85.2 / 94.2 & 86.0     & 82.1     & 92.6    & --       & 853       & --      & --          & --           \\
Molmo-72B~\cite{deitke2024molmo}           & \rsp / 96.3 & 87.3     & 83.1     & 93.5    & 81.9     & --        & --      & --          & --           \\
Qwen2-VL-72B~\cite{wang2024qwen2vl}        & 88.1 / \lsp & 88.3     & 85.5     & 96.5    & 84.5     & 877       & --      & --          & 91.3 / 94.6  \\
InternVL2-Llama3-76B~\cite{chen2024far}    & 87.6 / 94.8 & 88.4     & 84.4     & 94.1    & 82.0     & 839       & 69.7    & 38.9 / 75.2 & 83.2 / 91.3  \\
\rowcolor{gray!15}
InternVL2.5-78B                            & 89.1 / 95.7 & 88.3     & 83.4     & 95.1    & 84.1    & 854       & 71.3    & 42.4 / 82.3 & 95.7 / 94.5  \\
\end{tabular}
}
\caption{\textbf{Comparison of OCR, chart, and document understanding performance.} We evaluate OCR-related capabilities across 9 benchmarks, including AI2D~\cite{kembhavi2016ai2d}, ChartQA~\cite{masry2022chartqa}, TextVQA~\cite{singh2019textvqa}, DocVQA~\cite{mathew2021docvqa}, InfoVQA~\cite{mathew2022infographicvqa}, OCRBench~\cite{liu2023ocrbench}, SEED-2-Plus~\cite{li2024seedbench2plus}, CharXiv~\cite{wang2024charxiv}, and VCR~\cite{zhang2024vcr}.
Part of results are collected from \cite{dubey2024llama3, deitke2024molmo, claude3series2024, wang2024charxiv, zhang2024vcr} and the OpenCompass leaderboard~\cite{opencompass2023}.
}
\vspace{-0.5em}
\label{tab:benchmark_ocr}
\end{table*}

We assess InternVL's OCR, chart, and document understanding capabilities through a comprehensive evaluation on a variety of OCR-related datasets.

\textbf{AI2D}~\cite{kembhavi2016ai2d}: AI2D is a dataset of over 5,000 elementary school science diagrams, each with detailed annotations and corresponding multiple-choice questions. For a fair comparison, we report results for both ``mask'' and ``no mask'' settings on the test set.

\textbf{ChartQA}~\cite{masry2022chartqa}: ChartQA is a dataset focused on assessing models' abilities to interpret and reason with data visualizations such as charts and graphs. Our evaluation metric is the average relaxed accuracy across both human and augmented test sets in ChartQA.

\textbf{TextVQA}~\cite{singh2019textvqa}: TextVQA is a dataset designed to benchmark visual reasoning based on text within images. It requires models to read and interpret text in images to accurately answer related questions. We report the VQA accuracy on the TextVQA validation set.

\textbf{DocVQA}~\cite{clark2017docqa}: DocVQA is a dataset aimed at evaluating models' ability to comprehend and retrieve information from text within document images. Performance is reported on the test set using the ANLS metric, which captures answer accuracy by measuring text similarity.

\textbf{InfoVQA}~\cite{mathew2022infographicvqa}: InfographicVQA is a dataset aimed at evaluating models' ability to interpret and reason with complex infographics that combine text, graphics, and visual elements. Performance is measured using the ANLS metric on the test set.

\textbf{OCRBench}~\cite{liu2023ocrbench}: OCRBench evaluates the OCR capabilities of MLLMs across five tasks: text recognition, scene text VQA, document VQA, key information extraction, and handwritten math expression recognition, with a maximum score of 1000.

\textbf{SEEDBench-2-Plus}~\cite{li2024seedbench2plus}: SEED-Bench-2-Plus evaluates MLLMs on text-rich visual tasks, with 2,300 human-annotated questions across charts, maps, and webs. We report the average accuracy on this dataset.

\textbf{CharXiv}~\cite{wang2024charxiv}: CharXiv is a comprehensive evaluation suite featuring 2,323 charts from scientific papers. It includes two types of questions: reasoning questions (RQ) requiring synthesis of complex visual information, and descriptive questions (DQ) assessing basic chart element understanding.

\textbf{VCR}~\cite{zhang2024vcr}: Visual Caption Restoration (VCR) is a task that involves restoring partially hidden text within images by understanding both the visual content and the text. We report the Exact Match (EM) score and Jaccard similarity on the VCR-EN-Easy subset.

\subsubsection{Evaluation Results}

Table~\ref{tab:benchmark_ocr} provides a detailed comparison of InternVL 2.5 with its predecessor InternVL 2.0, other representative open-source models (\eg, Qwen2-VL~\cite{wang2024qwen2vl}, LLaVA-OneVision~\cite{li2024llavaov}), and closed-source models (\eg, GPT-4o~\cite{gpt4v}, Claude-3.5-Sonnet~\cite{claude3series2024}) on OCR-related tasks. Across most benchmarks, InternVL 2.5 achieves significant improvements over InternVL 2.0 at all model scales and demonstrates performance comparable to the current state-of-the-art model, Qwen2-VL-72B~\cite{wang2024qwen2vl}, reflecting the effectiveness of the improvements in training strategies and data quality.

However, at the 2B scale, InternVL2.5-2B underperforms compared to Qwen2-VL-2B on benchmarks such as TextVQA~\cite{singh2019textvqa}, DocVQA~\cite{mathew2021docvqa}, and InfoVQA~\cite{mathew2022infographicvqa}. We suspect that, in addition to differences in data and training strategies, model architecture may also play a significant role. Specifically, Qwen2-VL-2B features a 600M vision encoder and a 1.5B language model, whereas InternVL2.5-2B employs a smaller 300M vision encoder paired with a 1.8B language model. It appears that, for a smaller-scale MLLM (\eg, 2B), the size of the vision encoder plays a relatively important role in OCR performance, given the same total parameter budget.

Additionally, InternVL 2.5 demonstrates exceptional performance on the visual caption restoration (VCR) task~\cite{zhang2024vcr}. The 2.5 series achieves a significant improvement over InternVL 2.0 on this task, with the 2B model reaching EM/Jaccard scores of 93.2/97.6, far surpassing the previous generation’s 32.9/59.2. This improvement can be attributed to the introduction of a small portion of the VCR training set (approximately 22K samples). We find that the model’s poor performance on VCR tasks was not due to inadequate OCR capabilities but rather to its insufficient instruction-following ability for task-specific directives. By leveraging these few but focused samples, InternVL 2.5 exhibits a remarkable enhancement in its instruction-following ability for the VCR task, resulting in a substantial performance boost.

\subsection{Multi-Image Understanding}

\subsubsection{Benchmarks}

\label{sec:multi_image_datasets}

We assess InternVL's capabilities in multi-image relation perception and understanding across various multi-image benchmarks.

\textbf{BLINK}~\cite{fu2024blink}: The BLINK benchmark evaluates the core visual perception capabilities of MLLMs through 14 tasks inspired by classic computer vision challenges. Over half of the questions involve multiple images. Our results are reported on the validation set.

\textbf{Mantis-Eval}~\cite{jiang2024mantis}: Mantis-Eval is a meticulously curated small-scale benchmark for evaluating MLLMs' reasoning capabilities across multiple images. It comprises 217 challenging, human-annotated problems covering topics such as size perception and weight comparison.

\textbf{MMIU}~\cite{meng2024mmiu}: MMIU is an extensive benchmark suite developed to rigorously assess the performance of MLLMs in multi-image tasks. It encompasses 7 distinct types of multi-image relationships and spans 52 diverse tasks, providing a comprehensive framework for evaluation.

\textbf{MuirBench}~\cite{wang2024muirbench}: MuirBench is a comprehensive benchmark for evaluating MLLMs capabilities in multi-image understanding. It spans 12 tasks and 10 types of multi-image relations and enhances model assessment with unanswerable instance variants.

\textbf{MMT-Bench}~\cite{mmtbench}: MMT-Bench evaluates MLLMs on multimodal tasks like driving and navigation, focusing on recognition, reasoning, and planning, with many sub-tasks requiring multi-image understanding. To speed up testing, results are reported on the validation set.

\textbf{MIRB}~\cite{zhao2024mirb}: MIRB is a benchmark designed to evaluate the ability of MLLMs to understand and reason across multiple images. It contains four task categories: perception, visual world knowledge, reasoning, and multi-hop reasoning. The reported performance is the average score across these four categories.

\begin{table*}[t!]
\scriptsize
\centering
{\fontsize{8}{10}\selectfont 
\renewcommand{\arraystretch}{0.95}
\setlength\tabcolsep{3.5pt}
\newcommand{\MMIU}{\makecell{MMIU}}
\newcommand{\Muir}{\makecell{Muir\\Bench}} 
\newcommand{\BLINK}{\makecell{BLINK\\(val)}}
\newcommand{\Mantis}{\makecell{Mantis\\Eval}} 
\newcommand{\RWQA}{\makecell{RealWorld\\QA}} 
\newcommand{\MMERW}{\makecell{MME-RW\\(EN)}} 
\newcommand{\RBench}{\makecell{R-Bench\\(test)}}
\newcommand{\TaskMe}{\makecell{TaskMe-\\Anything}}
\newcommand{\MMT}{\makecell{MMT\\(val)}}
\newcommand{\WILDV}{\makecell{WildVision\\(win rate)}}
\newcommand{\MIRB}{\makecell{MIRB\\(avg)}}
\newcommand{\RB}{\makecell{R-Bench\\(dis)}}
\newcommand{\TODO}{\textcolor{red}{TODO}}
\newcommand{\lsp}{--\ \ \ \ \ \ }
\newcommand{\rsp}{\ \ \ \ \ --~}
\begin{tabular}{l|cccccc|cccc}
Model Name                                 & \BLINK & \Mantis & \MMIU & \Muir & \MMT & \MIRB &\RWQA  & \MMERW & \WILDV  & \RB  \\
\hline
LLaVA-OneVision-0.5B~\cite{li2024llavaov}  & 52.1   & 39.6    & --    & 25.5  & --   & --    & 55.6  & --     & --      & --   \\
InternVL2-1B~\cite{chen2024far}            & 38.6   & 46.1    & 37.3  & 29.3  & 49.5 & 31.5  & 50.3  & 40.2   & 17.8    & 55.6 \\
\rowcolor{gray!15}
InternVL2.5-1B                             & 42.0   & 51.2    & 38.5  & 29.9  & 50.3 & 35.6  & 57.5  & 44.2   & 43.4    & 59.0 \\
Qwen2-VL-2B~\cite{wang2024qwen2vl}         & 44.4   & --      & --    & --    & 55.1 & --    & 62.6  & --     & --      & --   \\
InternVL2-2B~\cite{chen2024far}            & 43.8   & 48.4    & 39.8  & 32.5  & 50.4 & 32.1  & 57.3  & 47.3   & 31.8    & 56.8 \\
\rowcolor{gray!15}
InternVL2.5-2B                             & 44.0   & 54.8    & 43.5  & 40.6  & 54.5 & 36.4  & 60.1  & 48.8   & 44.2    & 62.2 \\
\hline
Phi-3.5-Vision-4B~\cite{abdin2024phi3}     & 58.3   & --      & --    & --    & 53.6 & --    & 53.6  & --     & --      & 55.5 \\
InternVL2-4B~\cite{chen2024far}            & 46.1   & 61.3    & 43.3  & 40.5  & 55.7 & 39.9  & 60.7  & 52.1   & 44.2    & 64.5 \\
\rowcolor{gray!15}
InternVL2.5-4B                             & 50.8   & 62.7    & 43.8  & 45.2  & 62.4 & 51.7  & 64.3  & 55.3   & 49.4    & 66.1 \\
\hline
Qwen2-VL-7B~\cite{wang2024qwen2vl}         & 53.2   & --      & --    & --    & 64.0 & --    & 70.1  & 56.5   & --      & 64.0 \\
MiniCPM-V2.6~\cite{yao2024minicpm}         & 53.0   & 69.0    & --    & --    & 60.8 & --    & 65.0  & --     & --      & --   \\
InternVL2-8B~\cite{chen2024far}            & 50.9   & 65.4    & 42.0  & 48.7  & 60.0 & 50.0  & 64.4  & 53.5   & 54.4    & 67.9 \\
\rowcolor{gray!15}
InternVL2.5-8B                             & 54.8   & 67.7    & 46.7  & 51.1  & 62.3 & 52.5  & 70.1  & 59.1   & 62.0    & 70.1 \\
\hline
InternVL-Chat-V1.5~\cite{chen2024far}      & 46.6   & 66.8    & 37.4  & 38.5  & 58.0 & 50.3  & 66.0  & 49.4   & 56.6    & 67.9 \\
InternVL2-26B~\cite{chen2024far}           & 56.2   & 69.6    & 42.6  & 50.6  & 60.6 & 53.7  & 68.3  & 58.7   & 62.2    & 70.1 \\
\rowcolor{gray!15}
InternVL2.5-26B                            & 61.8   & 75.6    & 49.4  & 61.1  & 66.9 & 55.7  & 74.5  & 61.8   & 65.2    & 72.9 \\
Cambrian-34B~\cite{tong2024cambrian}       & --     & --      & --    & --    & --   & --    & 67.8  & 44.1   & --      & --   \\
InternVL2-40B~\cite{chen2024far}           & 57.2   & 71.4    & 47.9  & 54.4  & 66.2 & 55.2  & 71.8  & 61.8   & 63.2    & 73.3 \\
\rowcolor{gray!15}
InternVL2.5-38B                            & 63.2   & 78.3    & 55.3  & 62.7  & 70.0 & 61.2  & 73.5  & 64.0   & 66.4    & 72.1 \\
\hline
GPT-4V~\cite{gpt4v}                        & 54.6   & 62.7    & --    & 62.3  & 64.3 & 53.1  & 61.4  & --     & 71.8    & 65.6 \\
GPT-4o-20240513~\cite{gpt4v}               & 68.0   & --      & 55.7  & 68.0  & 65.4 & --    & 75.4  & 45.2   & 80.6    & 77.7 \\
Claude-3.5-Sonnet~\cite{claude3series2024} & --     & --      & 53.4  & --    & --   & --    & 60.1  & 51.6   & --      & --   \\
Gemini-1.5-Pro~\cite{reid2024gemini1_5}    & --     & --      & 53.4  & --    & 64.5 & --    & 67.5  & 38.2   & --      & --   \\
LLaVA-OneVision-72B~\cite{li2024llavaov}   & 55.4   & 77.6    & --    & 54.8  & --   & --    & 71.9  & --     & --      & --   \\
Qwen2-VL-72B~\cite{wang2024qwen2vl}        & --     & --      & --    & --    & 71.8 & --    & 77.8  & --     & --      & --   \\
InternVL2-Llama3-76B~\cite{chen2024far}    & 56.8   & 73.7    & 44.2  & 51.2  & 67.4 & 58.2  & 72.2  & 63.0   & 65.8    & 74.1 \\
\rowcolor{gray!15}
InternVL2.5-78B                            & 63.8   & 77.0    & 55.8  & 63.5  & 70.8 & 61.1  & 78.7  & 62.9   & 71.4    & 77.2 \\
\end{tabular}
}
\caption{\textbf{Comparison of multi-image and real-world understanding performance. }
Multi-image benchmarks include BLINK~\cite{fu2024blink}, Mantis-Eval~\cite{jiang2024mantis}, MMIU~\cite{meng2024mmiu}, MuirBench~\cite{wang2024muirbench}, MMT-Bench~\cite{mmtbench}, and MIRB~\cite{zhao2024mirb}.
Real-world benchmarks encompass RealWorldQA~\cite{realworldqa}, MME-RealWorld~\cite{zhang2024mme}, WildVision~\cite{lu2024wildvision}, and R-Bench~\cite{li2024r}.
Part of the results are sourced from the benchmark papers and the OpenCompass leaderboard~\cite{opencompass2023}.
}
\label{tab:benchmark_multi_image_real_world}
\end{table*}

\subsubsection{Evaluation Results}

As multi-image content becomes an increasingly common form of information exchange on the internet, it is essential for models to possess the ability to simultaneously understand and analyze relationships between multiple images. In the left part of Table~\ref{tab:benchmark_multi_image_real_world}, we evaluate the multi-image understanding capabilities of InternVL 2.5 across six diverse benchmarks: BLINK~\cite{fu2024blink}, Mantis-Eval~\cite{jiang2024mantis}, MMIU~\cite{meng2024mmiu}, MuirBench~\cite{wang2024muirbench}, MMT-Bench~\cite{mmtbench}, and MIRB~\cite{zhao2024mirb}. These benchmarks test a range of skills, including reasoning across images, integrating information, and addressing task-specific requirements.

InternVL 2.5 achieves consistent improvements over InternVL 2.0 across all model scales, reflecting enhanced reasoning ability and better integration of multi-image information. For instance, at the 2B scale, InternVL2.5-2B delivers significant gains on Mantis-Eval (54.8 \emph{vs.} 48.4) and MuirBench (40.6 \emph{vs.} 32.5). These advancements can be largely attributed to the inclusion of additional multi-image datasets, as detailed in Section~\ref{sec:sft_data}. These datasets, which were carefully curated and of high quality, played a critical role in improving the model’s ability to understand and reason across multiple visual inputs. 

At larger scales, InternVL 2.5 demonstrates substantial progress and achieves competitive performance with advanced closed-source models. For example, InternVL2.5-78B scores 55.8 on MMIU, closely matching GPT-4o's 55.7, and achieves a score of 70.8 on MMT-Bench, surpassing GPT-4o's 65.4. These results highlight the importance of scaling model size and incorporating high-quality training data specifically tailored for multi-image tasks. 
However, on BLINK and MuirBench, our model still exhibits a performance gap of around 5 points compared to GPT-4o~\cite{gpt4v}, suggesting that further improvements are needed, potentially through the inclusion of additional high-quality multi-image training data.

\subsection{Real-World Comprehension}

\subsubsection{Benchmarks}

\label{sec:real_world_datasets}

We assess InternVL's performance on a suite of real-world benchmarks designed to evaluate its capabilities on realistic and complex tasks.

\textbf{RealWorldQA}~\cite{realworldqa}: RealWorldQA is a benchmark designed to evaluate the real-world spatial understanding capabilities of MLLMs. It contains more than 700 images, each accompanied by a question and a verifiable answer, from various real-world scenarios.

\textbf{MME-RealWorld}~\cite{zhang2024mme}: MME-RealWorld is a benchmark for evaluating MLLMs on complex, high-resolution image tasks across 43 real-world scenarios in 5 domains. Here, we test the English full set of the dataset.

\textbf{WildVision}~\cite{lu2024wildvision}: WildVision-Bench is a benchmark designed to evaluate MLLMs in the wild with human preferences. It comprises 500 high-quality samples meticulously curated from real-world user QA interactions. The benchmark uses a win rate metric to quantify the performance of models, providing insights into their ability to meet human expectations in practical applications.

\textbf{R-Bench}~\cite{li2024r}: R-Bench is a benchmark designed to evaluate the robustness of MLLMs against real-world image distortions, measuring their resilience in handling corrupted images in practical scenarios. We report the absolute robustness overall score for the MCQ task, which is the average score across low, mid, and high difficulty levels, corresponding to ``R-Bench-Dis'' in VLMEvalKit.

\subsubsection{Evaluation Results}

Given the complexity and dynamic nature of real-world environments, models must be robust enough to handle a wide range of challenging conditions. As shown in the right part of Table~\ref{tab:benchmark_multi_image_real_world}, InternVL 2.5 achieves leading performance across four real-world understanding benchmarks, including RealWorldQA~\cite{realworldqa}, MME-RealWorld~\cite{zhang2024mme}, WildVision~\cite{lu2024wildvision}, and R-Bench~\cite{li2024r}, and significantly outperforms the previous version, InternVL 2.0. This indicates that InternVL 2.5 has a stronger potential for practical application in complex and ever-changing real-world scenarios.

In benchmarks like RealWorldQA, MME-RealWorld, and R-Bench, which involve multiple-choice questions, InternVL 2.5 demonstrates strong real-world perceptual and understanding abilities. Differently, the WildVision benchmark uses GPT-4o~\cite{gpt4v} as the judge model to evaluate the performance of various MLLM against the reference model, Claude-3-Sonnet~\cite{claude3series2024}. In this benchmark, the model's output quality and user experience are key metrics. Although InternVL2.5-78B performs well in providing concise answers, it still shows a gap when generating longer responses to match human preferences. Specifically, InternVL2.5-78B scores 71.4, while GPT-4o scores 80.6, indicating a notable difference in user experience.

These results indicate that, while InternVL 2.5 delivers accurate and concise responses across most tasks, there is potential for improvement in generating more personalized and detailed answers. Future work will focus on enhancing the model's performance in open-ended tasks and complex interactions, aiming to better align with human preferences, bridge the gap in user experience with GPT-4o.

\subsection{Comprehensive Multimodal Evaluation}

\subsubsection{Benchmarks}

We evaluate InternVL's comprehensive multimodal capabilities through a range of benchmarks, including:

\noindent \textbf{MME}~\cite{fu2023mme}: MME is the first comprehensive evaluation benchmark designed for MLLMs. It assesses models' perception and cognitive abilities across 14 subtasks, including object presence, counting, position, color recognition, as well as commonsense reasoning, numerical computation, text translation, and code reasoning. We report the overall score across all tasks.

\noindent \textbf{MMBench}~\cite{liu2023mmbench}: MMBench evaluates the multimodal understanding of MLLMs through nearly 3,000 multiple-choice questions spanning 20 dimensions. It supports both English and Chinese versions, and we present the model's performance scores on the test set.

\noindent \textbf{MMBench v1.1}~\cite{liu2023mmbench}: Compared to MMBench, MMBench v1.1 features a refined dataset with a small number of noisy or low-quality questions removed, resulting in a subtle improvement in overall data quality. We report the model's performance on the English version of the test set.

\noindent \textbf{MMVet}~\cite{yu2023mmvet}: MMVet is a benchmark designed to assess the integrated capabilities of MLLMs on complex tasks. It evaluates six core competencies: recognition, knowledge, spatial awareness, language generation, OCR, and mathematics, across 16 integrated tasks. Note that VLMEvalKit uses GPT-4-Turbo as the scoring model for this benchmark, which yields slightly lower scores compared to the official evaluation server.

\noindent \textbf{MMVet v2}~\cite{yu2024mmvetv2}: Expanding on MMVet, MMVet v2 introduces an enhanced benchmark with a new capability: image-text sequence understanding, allowing for the assessment of models' ability to process interleaved content. Here, we utilize the official evaluation server for scoring, which employs GPT-4-0613 as the scoring model.

\noindent \textbf{MMStar}~\cite{chen2024mmstar}: MMStar is a benchmark for evaluating the multimodal capabilities of MLLMs. It includes 1,500 carefully curated samples focusing on advanced visual and language understanding, minimizing data leakage, and emphasizing visual dependency.

\subsubsection{Evaluation Results}

Comprehensive multimodal evaluation benchmarks, such as MME~\cite{fu2023mme}, the MMBench series~\cite{liu2023mmbench}, the MMVet series~\cite{yu2023mmvet, yu2024mmvetv2}, and MMStar~\cite{chen2024mmstar}, provide valuable and widely adopted frameworks for assessing model performance across a diverse set of multimodal tasks.

As shown in the left section of Table~\ref{tab:benchmark_multimodal_hallucination}, the InternVL 2.5 models consistently outperform the previous InternVL 2.0 series across various model sizes, especially for smaller models with 1B-8B parameters. For example, in the MMBench v1.0 benchmark, which evaluates tasks in both English and Chinese, the InternVL 2.5 models show significant improvements. The InternVL2.5-4B achieves a score of 81.1/79.3, surpassing the InternVL2-4B’s 78.6/73.9, while the InternVL2.5-8B reaches 84.6/82.6, outperforming the InternVL2-8B’s 81.7/81.2.

It is also noteworthy that, while we have significantly improved the performance of smaller models on the MMVet series benchmarks, our largest model, InternVL2.5-78B, still does not surpass the Qwen2-VL-72B~\cite{wang2024qwen2vl}. Currently, the state-of-the-art models on MMVet v2 remain closed-source models like GPT-4o~\cite{gpt4v} and Claude-3.5-Sonnet~\cite{claude3series2024}. This highlights the gap between open-source models and closed-source ones in multimodal integrated capability. We recognize this as an important direction for future development.

\begin{table*}[t!]
\centering
\renewcommand{\arraystretch}{0.95}
{\fontsize{8}{10}\selectfont 
\setlength\tabcolsep{2.6pt}
\newcommand{\MME}{\makecell{MME\\(sum)}}
\newcommand{\MMB}{\makecell{MMB\\(EN / CN)}}
\newcommand{\MMBV}{\makecell{MMBv1.1\\(EN)}}
\newcommand{\MMVet}{\makecell{MMVet\\(turbo)}}
\newcommand{\MMVetV}{\makecell{MMVetv2\\(0613)}}
\newcommand{\MMStar}{\makecell{MMStar}}
\newcommand{\POPE}{\makecell{POPE\\(avg)}}
\newcommand{\HallB}{\makecell{HallBench\\(avg)}}
\newcommand{\AMBER}{\makecell{AMBER\\(generative / discriminative / overall)}}
\newcommand{\ObjectHalBench}{\makecell{ObjectHal}}
\newcommand{\MMHal}{\makecell{MMHal\\(score)}}
\newcommand{\CRPE}{\makecell{CRPE\\(relation)}}
\begin{tabular}{l|cccccc|cccc}
Model Name                                & \MME   & \MMB        & \MMBV & \MMVet & \MMVetV & \MMStar & \HallB & \MMHal & \CRPE & \POPE \\
\hline
LLaVA-OneVision-0.5B~\cite{li2024llavaov} & 1438.0 & 61.6 / 55.5 & 59.6  & 32.2   & --      & 37.7    & 27.9   & --     & --    & --    \\
InternVL2-1B~\cite{chen2024far}           & 1794.4 & 65.4 / 60.7 & 61.6  & 32.7   & 36.1    & 45.7    & 34.0   & 2.25   & 57.5  & 87.3  \\
\rowcolor{gray!15}
InternVL2.5-1B                            & 1950.5 & 70.7 / 66.3 & 68.4  & 48.8   & 43.2    & 50.1    & 39.0   & 2.49   & 60.9  & 89.9  \\
Qwen2-VL-2B~\cite{wang2024qwen2vl}        & 1872.0 & 74.9 / 73.5 & 72.2  & 49.5   & --      & 48.0    & 41.7   & --     & --    & --    \\
InternVL2-2B~\cite{chen2024far}           & 1876.8 & 73.2 / 70.9 & 70.2  & 39.5   & 39.6    & 50.1    & 37.9   & 2.52   & 66.3  & 88.3  \\
\rowcolor{gray!15}
InternVL2.5-2B                            & 2138.2 & 74.7 / 71.9 & 72.2  & 60.8   & 52.3    & 53.7    & 42.6   & 2.94   & 70.2  & 90.6  \\
\hline
Phi-3.5-Vision-4B~\cite{abdin2024phi3}    & --     & 76.0 / 66.1 & 72.1  & 43.2   & --      & 47.5    & 40.5   & --     & --    & --    \\
InternVL2-4B~\cite{chen2024far}           & 2059.8 & 78.6 / 73.9 & 75.8  & 51.0   & 46.6    & 54.3    & 41.9   & 2.75   & 71.1  & 87.2  \\
\rowcolor{gray!15}
InternVL2.5-4B                            & 2337.5 & 81.1 / 79.3 & 79.3  & 60.6   & 55.4    & 58.3    & 46.3   & 3.31   & 75.5  & 90.9  \\
\hline
Qwen2-VL-7B~\cite{wang2024qwen2vl}        & 2326.8 & 83.0 / 80.5 & 80.7  & 62.0   & --      & 60.7    & 50.6   & 3.40   & 74.4  & 88.1  \\
MiniCPM-V2.6~\cite{yao2024minicpm}        & 2348.4 & 81.5 / 79.3 & 78.0  & 60.0   & --      & 57.5    & 48.1   & 3.60   & 75.2  & 87.3  \\
InternVL2-8B~\cite{chen2024far}           & 2210.3 & 81.7 / 81.2 & 79.5  & 54.2   & 52.3    & 62.0    & 45.2   & 3.33   & 75.8  & 86.9  \\
\rowcolor{gray!15}
InternVL2.5-8B                            & 2344.1 & 84.6 / 82.6 & 83.2  & 62.8   & 58.1    & 62.8    & 50.1   & 3.65   & 78.4  & 90.6  \\
\hline
InternVL-Chat-V1.5~\cite{chen2024far}     & 2194.2 & 82.2 / 82.0 & 80.3  & 61.5   & 51.5    & 57.3    & 50.3   & 3.11   & 75.4  & 88.4  \\
InternVL2-26B~\cite{chen2024far}          & 2260.7 & 83.4 / 82.0 & 81.5  & 62.1   & 57.2    & 61.2    & 50.7   & 3.55   & 75.6  & 88.0  \\
\rowcolor{gray!15}
InternVL2.5-26B                           & 2373.3 & 85.4 / 85.5 & 84.2  & 65.0   & 60.8    & 66.5    & 55.0   & 3.70   & 79.1  & 90.6  \\
Cambrian-34B~\cite{tong2024cambrian}      & --     & 80.4 / 79.2 & 78.3  & 53.2   & --      & 54.2    & 41.6   & --     & --    & --    \\
InternVL2-40B~\cite{chen2024far}          & 2307.5 & 86.8 / 86.5 & 85.1  & 65.5   & 63.8    & 65.4    & 56.9   & 3.75   & 77.6  & 88.4  \\
\rowcolor{gray!15}
InternVL2.5-38B                           & 2455.8 & 86.5 / 86.3 & 85.5  & 68.8   & 62.1    & 67.9    & 56.8   & 3.71   & 78.3  & 90.7  \\
\hline
GPT-4V~\cite{gpt4v}                       & 1926.6 & 81.0 / 80.2 & 80.0  & 67.5   & 66.3    & 56.0    & 46.5   & --     & --    & --    \\
GPT-4o-20240513~\cite{gpt4v}              & --     & 83.4 / 82.1 & 83.1  & 69.1   & 71.0    & 64.7    & 55.0   & 4.00   & 76.6  & 86.9  \\
Claude-3-Opus~\cite{claude3series2024}    & 1586.8 & 63.3 / 59.2 & 60.1  & 51.7   & 55.8    & 45.7    & 37.8   & --     & --    & --    \\
Claude-3.5-Sonnet~\cite{claude3series2024}& --     & 82.6 / 83.5 & 80.9  & 70.1   & 71.8    & 65.1    & 55.5   & --     & --    & --    \\
Gemini-1.5-Pro~\cite{reid2024gemini1_5}   & --     & 73.9 / 73.8 & 74.6  & 64.0   & 66.9    & 59.1    & 45.6   & --     & --    & --    \\
LLaVA-OneVision-72B~\cite{li2024llavaov}  & 2261.0 & 85.8 / 85.3 & 85.0  & 60.6   & --      & 65.8    & 49.0   & --     & --    & --    \\
Qwen2-VL-72B~\cite{wang2024qwen2vl}       & 2482.7 & 86.5 / 86.6 & 85.9  & 74.0   & 66.9    & 68.3    & 58.1   & --     & --    & --    \\
InternVL2-Llama3-76B~\cite{chen2024far}   & 2414.7 & 86.5 / 86.3 & 85.5  & 65.7   & 68.4    & 67.4    & 55.2   & 3.83   & 77.6  & 89.0  \\
\rowcolor{gray!15}
InternVL2.5-78B                           & 2494.5 & 88.3 / 88.5 & 87.4  & 72.3   & 65.5    & 69.5    & 57.4   & 3.89   & 78.8  & 90.8  \\
\end{tabular}
}

\caption{\textbf{Comparison of comprehensive multimodal understanding and hallucination performance.}
Comprehensive multimodal benchmarks include MME~\cite{fu2023mme}, MMBench series~\cite{liu2023mmbench}, MMVet series~\cite{yu2023mmvet, yu2024mmvetv2}, and MMStar~\cite{chen2024mmstar}.
Hallucination benchmarks encompass HallusionBench~\cite{guan2023hallusionbench}, MMHal~\cite{sun2023aligning}, CRPE~\cite{wang2024allseeingv2}, and POPE~\cite{li2023pope}.
Part of the results are sourced from the benchmark papers and the OpenCompass leaderboard~\cite{opencompass2023}.
}
\label{tab:benchmark_multimodal_hallucination}
\end{table*}

\subsection{Multimodal Hallucination Evaluation}

\subsubsection{Benchmarks}

\label{sec:hall_datasets}

We evaluate InternVL's tendency toward hallucinations across four different benchmarks, including:

\noindent \textbf{HallusionBench}~\cite{guan2023hallusionbench}: HallusionBench is a benchmark for evaluating image-context reasoning in MLLMs through a Yes/No judgment question format, focusing on challenges such as language hallucination and visual illusion. We report performance using the average scores of its three metrics: aAcc, fAcc, and qAcc.

\noindent \textbf{MMHal-Bench}~\cite{sun2023aligning}: MMHal-Bench is a benchmark designed to evaluate hallucinations in MLLMs. It includes 96 challenging questions derived from images in the OpenImages dataset, along with their corresponding ground-truth answers and image content. Scoring is conducted using GPT-4o, with scores ranging from 0 to 6.

\noindent \textbf{CRPE}~\cite{wang2024allseeingv2}: CRPE is a benchmark that measures the hallucination level of the relation between objects using multiple-choice questions. We report accuracy on the relation subset for this benchmark.

\noindent \textbf{POPE}~\cite{li2023pope}: POPE is a benchmark for evaluating object hallucination in MLLMs, utilizing binary questions to quantify and analyze hallucination tendencies. We report the average F1 score across three categories: random, popular, and adversarial.

\subsubsection{Evaluation Results}

We evaluate the performance of InternVL on four key hallucination evaluation benchmarks: HallusionBench~\cite{guan2023hallusionbench}, MMHal~\cite{sun2023aligning}, CRPE~\cite{wang2024allseeingv2}, and POPE~\cite{li2023pope}. These benchmarks assess the frequency of hallucinations, or factual inaccuracies, across multimodal tasks, providing a measure of model reliability in handling complex inputs like text and images.

The InternVL 2.5 models show significant progress over the InternVL 2.0 series, particularly in smaller models (\eg, 1B-8B parameters). For instance, InternVL2.5-1B and InternVL2.5-2B demonstrate improved scores on all hallucination benchmarks, with the 1B model achieving a 39.0 score on HallusionBench, up from 34.0 in the earlier version. Similarly, the 2B model improved to 42.6, outperforming the previous 2B model by nearly 5 points. These results indicate substantial gains in reducing hallucinations while handling multimodal data.

The largest model, InternVL2.5-78B, also shows improvements, reducing hallucinations compared to both prior versions and other leading models. It scores 57.4 on HallusionBench, competing with top models like Qwen2-VL-72B (58.1) and GPT-4o (55.0). 
Although InternVL2.5-78B demonstrates relatively low hallucination rates on these hallucination evaluation benchmarks, some hallucinations are still inevitably present when generating long responses in practical use. This is a challenge we plan to tackle in future work.

\subsection{Visual Grounding}

\begin{table*}[t] 
\centering 
\renewcommand{\arraystretch}{0.95}
\setlength\tabcolsep{9.2pt}
{\fontsize{8}{10}\selectfont 
\begin{tabular}{l|ccc|ccc|cc|c}
\multirow{2}{*}{{Model Name}}           & \multicolumn{3}{c|}{{RefCOCO}} & \multicolumn{3}{c|}{{RefCOCO+}} & \multicolumn{2}{c|}{{RefCOCOg}} & \multirow{2}{*}{{avg.}} \\
                                        & val    & test-A    & test-B    & val    & test-A    & test-B     & val     & test                 &         \\
\hline
Grounding-DINO-L~\citep{grounding_dino} & 90.6   & 93.2      & 88.2      & 82.8   & 89.0      & 75.9       & 86.1    & 87.0                 & 86.6    \\
UNINEXT-H~\citep{uninext}               & 92.6   & 94.3      & 91.5      & 85.2   & 89.6      & 79.8       & 88.7    & 89.4                 & 88.9    \\
ONE-PEACE~\citep{one-peace}             & 92.6   & 94.2      & 89.3      & 88.8   & 92.2      & 83.2       & 89.2    & 89.3                 & 89.8    \\
\hline
Shikra-7B~\citep{chen2023shikra}        & 87.0   & 90.6      & 80.2      & 81.6   & 87.4      & 72.1       & 82.3    & 82.2                 & 82.9    \\
Ferret-v2-13B~\citep{ferretv2}          & 92.6   & 95.0      & 88.9      & 87.4   & 92.1      & 81.4       & 89.4    & 90.0                 & 89.6    \\
CogVLM-Grounding-17B~\citep{wang2023cogvlm} 
                                        & 92.8   & 94.8      & 89.0      & 88.7   & 92.9      & 83.4       & 89.8    & 90.8                 & 90.3    \\
MM1.5~\cite{zhang2024mm1_5}             & --     & 92.5      & 86.7      & --     & 88.7      & 77.8       & --      & 87.1                 & --      \\
Qwen2-VL-7B~\cite{wang2024qwen2vl}      & 91.7   & 93.6      & 87.3      & 85.8   & 90.5      & 79.5       & 87.3    & 87.8                 & 87.9    \\
TextHawk2~\citep{yu2024texthawk2}       & 91.9   & 93.0      & 87.6      & 86.2   & 90.0      & 80.4       & 88.2    & 88.1                 & 88.2    \\
InternVL2-8B~\cite{chen2024far}         & 87.1   & 91.1      & 80.7      & 79.8   & 87.9      & 71.4       & 82.7    & 82.7                 & 82.9    \\
\rowcolor{gray!15}
InternVL2.5-8B                          & 90.3   & 94.5      & 85.9      & 85.2   & 91.5      & 78.8       & 86.7    & 87.6                 & 87.6    \\
\hline
Qwen2-VL-72B~\cite{wang2024qwen2vl}     & 93.2   & 95.3      & 90.7      & 90.1   & 93.8      & 85.6       & 89.9    & 90.4                 & 91.1    \\
InternVL2-Llama3-76B~\cite{chen2024far} & 92.2   & 94.8      & 88.4      & 88.8   & 93.1      & 82.8       & 89.5    & 90.3                 & 90.0    \\
\rowcolor{gray!15}
InternVL2.5-78B                         & 93.7   & 95.6      & 92.5      & 90.4   & 94.7      & 86.9       & 92.7    & 92.2                 & 92.3    \\
\end{tabular}
}
\caption{\textbf{Comparison of visual grounding performance.}
We evaluate InternVL's visual grounding capability on RefCOCO, RefCOCO+, and RefCOCOg datasets~\cite{kazemzadeh2014referitgame, mao2016generation}. Parts of the results are collected from \cite{wang2024qwen2vl}.
} 
\label{tab:benchmark-grounding}
\end{table*}

\subsubsection{Benchmarks}
We evaluate InternVL's visual grounding capability via referring expression comprehension (REC) on the RefCOCO, RefCOCO+, and RefCOCOg datasets, where the model identifies target objects in images from given descriptions.

\textbf{RefCOCO}~\cite{kazemzadeh2014referitgame}: 
Built on COCO, this dataset contains 19,994 images with 142,210 referring expressions for 50,000 objects, split into subsets like test A (people-focused) and test B (other objects) for REC tasks.

\textbf{RefCOCO+}~\cite{kazemzadeh2014referitgame}:
Similar to RefCOCO but emphasizing attribute-based descriptions without absolute location cues. It includes 19,992 images and 141,564 expressions, requiring models to focus on descriptive attributes.

\textbf{RefCOCOg}~\cite{mao2016generation}:
With 25,799 images and 95,010 expressions, this dataset features longer, more complex expressions, and challenging models to manage intricate language in REC tasks.

\subsubsection{Evaluation Results}

Visual grounding is critical for connecting textual descriptions with visual content, enabling accurate multimodal interaction. Table~\ref{tab:benchmark-grounding} compares InternVL 2.5 with its predecessor, InternVL 2.0, at the 8B and 78B scales, alongside other leading MLLMs (\eg, CogVLM-Grounding-17B~\cite{wang2023cogvlm}, Qwen2-VL~\cite{wang2024qwen2vl}) and specialized grounding models (\eg, Grounding-DINO-L~\cite{grounding_dino}, UNINEXT-H~\cite{uninext}, ONE-PEACE~\cite{one-peace}), evaluated on the RefCOCO~\cite{kazemzadeh2014referitgame}, RefCOCO+~\cite{kazemzadeh2014referitgame}, and RefCOCOg~\cite{mao2016generation} datasets.

InternVL2.5-8B improves its predecessor’s performance, with the average score rising from 82.9 to 87.6, achieving comparable results to Qwen2-VL-7B (87.6 \emph{vs.} 87.9), though slightly behind Ferret-v2-13B~\cite{ferretv2} and CogVLM-Grounding-17B~\cite{wang2023cogvlm}, which benefit from fine-tuning for grounding and larger model sizes. At the larger scale, InternVL2.5-78B achieves state-of-the-art performance with an average score of 92.3, a 2.3-point improvement over InternVL2-Llama3-76B, surpassing Qwen2-VL-72B~\cite{wang2024qwen2vl}. These gains highlight the effectiveness of our data and training optimizations, significantly enhancing localization capabilities.

\subsection{Multimodal Multilingual Understanding}

\begin{table*}[t!]
\scriptsize
\centering
\renewcommand{\arraystretch}{0.95}
{\fontsize{8}{10}\selectfont 
\setlength\tabcolsep{5.2pt}
\newcommand{\MMMB}{\makecell{MMMB}}
\newcommand{\MultiMMB}{\makecell{Multilingual MMBench}}
\begin{tabular}{l|cccccc|cccccc|c}
\multirow{2}{*}{Model Name}               & \multicolumn{6}{c|}{\MMMB}              & \multicolumn{6}{c|}{\MultiMMB}          & MTVQA   \\
                                          & en   & zh   & pt   & ar   & tr   & ru   & en   & zh   & pt   & ar   & tr   & ru   & (avg)   \\
\hline
InternVL2-1B~\cite{chen2024far}           & 73.2 & 67.4 & 55.5 & 53.5 & 43.8 & 55.2 & 67.9 & 61.2 & 50.8 & 43.3 & 31.8 & 52.7 & 12.6    \\
\rowcolor{gray!15}
InternVL2.5-1B                            & 78.8 & 70.2 & 61.5 & 55.0 & 45.3 & 61.1 & 72.5 & 64.7 & 57.0 & 43.0 & 37.8 & 53.2 & 21.4    \\
Qwen2-VL-2B~\cite{wang2024qwen2vl}        & 78.3 & 74.2 & 72.6 & 68.3 & 61.8 & 72.8 & 72.1 & 71.1 & 69.9 & 61.1 & 54.4 & 69.3 & 20.0    \\
InternVL2-2B~\cite{chen2024far}           & 79.4 & 71.6 & 54.0 & 43.5 & 46.4 & 48.1 & 73.8 & 69.6 & 51.4 & 29.8 & 31.3 & 42.3 & 10.9    \\
\rowcolor{gray!15}
InternVL2.5-2B                            & 81.4 & 74.4 & 58.2 & 48.3 & 46.4 & 53.2 & 76.5 & 71.6 & 55.9 & 37.3 & 33.9 & 44.8 & 21.8    \\
\hline
InternVL2-4B~\cite{chen2024far}           & 82.0 & 76.1 & 75.6 & 54.3 & 51.2 & 67.4 & 77.3 & 72.4 & 72.6 & 43.6 & 46.5 & 61.2 & 15.3    \\
\rowcolor{gray!15}
InternVL2.5-4B                            & 83.7 & 81.0 & 79.7 & 76.0 & 70.5 & 79.9 & 82.3 & 81.1 & 78.9 & 73.4 & 68.1 & 76.2 & 28.4    \\
\hline
mPLUG-Owl2~\cite{ye2023mplug2}            & 67.3 & 61.0 & 59.7 & 45.8 & 45.4 & 62.6 & 66.2 & 59.4 & 58.2 & 37.9 & 47.7 & 60.4 & --      \\
Qwen2-VL-7B~\cite{wang2024qwen2vl}        & 83.9 & 82.4 & 81.2 & 79.0 & 74.7 & 82.4 & 81.8 & 81.6 & 79.1 & 75.6 & 74.5 & 79.3 & 25.6    \\
InternVL2-8B~\cite{chen2024far}           & 83.4 & 81.5 & 76.1 & 66.3 & 69.2 & 75.7 & 82.9 & 81.8 & 76.0 & 60.5 & 66.0 & 74.4 & 20.9    \\ 
\rowcolor{gray!15}
InternVL2.5-8B                            & 84.3 & 83.1 & 78.6 & 69.3 & 71.5 & 79.5 & 83.8 & 83.2 & 79.4 & 64.3 & 67.8 & 77.3 & 27.6    \\
\hline
InternVL-Chat-V1.5~\cite{chen2024far}     & 82.6 & 80.8 & 76.3 & 65.2 & 68.6 & 74.0 & 81.1 & 80.2 & 76.9 & 56.2 & 66.7 & 71.0 & 20.5    \\
InternVL2-26B~\cite{chen2024far}          & 83.8 & 81.7 & 78.0 & 68.8 & 69.3 & 76.3 & 82.7 & 81.8 & 77.8 & 61.9 & 69.6 & 74.4 & 17.7    \\
\rowcolor{gray!15}
InternVL2.5-26B                           & 86.2 & 83.8 & 81.6 & 73.3 & 73.7 & 82.8 & 86.1 & 85.5 & 80.7 & 67.5 & 75.0 & 79.6 & 28.5    \\
InternVL2-40B~\cite{chen2024far}          & 85.3 & 84.1 & 81.1 & 70.3 & 74.2 & 81.4 & 86.2 & 85.8 & 82.8 & 64.0 & 74.2 & 81.8 & 20.6    \\
\rowcolor{gray!15}
InternVL2.5-38B                           & 86.4 & 85.1 & 84.1 & 84.3 & 82.8 & 84.9 & 87.5 & 88.6 & 85.3 & 84.5 & 84.0 & 85.9 & 31.7    \\
\hline
GPT-4V~\cite{gpt4v}                       & 75.0 & 74.2 & 71.5 & 73.5 & 69.0 & 73.1 & 77.6 & 74.4 & 72.5 & 72.3 & 70.5 & 74.8 & 22.0    \\
GPT-4o~\cite{gpt4v}                       & --   & --   & --   & --   & --   & --   & --   & --   & --   & --   & --   & --   & 27.8    \\
Gemini-1.0-Pro~\cite{team2023gemini}      & 75.0 & 71.9 & 70.6 & 69.9 & 69.6 & 72.7 & 73.6 & 72.1 & 70.3 & 61.1 & 69.8 & 70.5 & --      \\
Qwen2-VL-72B~\cite{wang2024qwen2vl}       & 86.8 & 85.3 & 85.2 & 84.8 & 84.2 & 85.3 & 86.9 & 87.2 & 85.8 & 83.5 & 84.4 & 85.3 & 30.9    \\
InternVL2-Llama3-76B~\cite{chen2024far}   & 85.3 & 85.1 & 82.8 & 82.8 & 83.0 & 83.7 & 87.8 & 87.3 &	85.9 & 83.1 & 85.0 & 85.7 & 22.0    \\
\rowcolor{gray!15}
InternVL2.5-78B                           & 86.3 & 85.6 & 85.1 & 84.8 & 83.1 & 85.4 & 90.0 & 89.7 & 87.4 & 83.3 & 84.9 & 86.3 & 31.9    \\
\end{tabular}
}
\caption{\textbf{Comparison of multimodal multilingual performance. }
We evaluate multilingual capabilities across 3 benchmarks, including MMMB~\cite{sun2024parrot}, Multilingual MMBench~\cite{sun2024parrot} and MTVQA~\cite{tang2024mtvqa}.
The languages evaluated are English (en), Chinese (zh), Portuguese (pt), Arabic (ar), Turkish (tr), and Russian (ru).
}
\label{tab:benchmark_multilingual}
\end{table*}

\subsubsection{Benchmarks}

\label{sec:multilingual_datasets}

We assess InternVL's multimodal multilingual understanding capabilities using three representative benchmarks: 

\textbf{MMMB and Multilingual MMBench}~\cite{sun2024parrot}: MMMB is a large-scale multilingual multimodal benchmark with 6 languages, 15 categories, and 12,000 questions. 
The languages evaluated are English (en), Chinese (zh), Portuguese (pt), Arabic (ar), Turkish (tr), and Russian (ru).
Multilingual MMBench extends MMBench~\cite{liu2023mmbench} to these 6 languages using GPT-4 translation for multilingual understanding evaluation.

\textbf{MTVQA}~\cite{tang2024mtvqa}: MTVQA is a multilingual benchmark tailored for text-centric visual question answering. It includes high-quality, expert human annotations across nine languages, specifically addressing the ``visual-text misalignment'' challenge in multilingual contexts. We report the average score of MTVQA. 

\subsubsection{Evaluation Results}

Multilingual ability is critical for MLLMs as it expands their application and improves cross-language communication. For global deployment, MLLMs must effectively handle both high-resource and low-resource languages. As shown in Table~\ref{tab:benchmark_multilingual}, we evaluated our model's performance on three multilingual benchmarks: MMMB~\cite{sun2024parrot}, Multilingual MMBench~\cite{sun2024parrot}, and MTVQA~\cite{tang2024mtvqa}.

A comparison between InternVL2.5-78B and Qwen2-VL-72B~\cite{wang2024qwen2vl} reveals that, despite differences in training data, model architecture, and training strategies, their multilingual performance is quite similar. This suggests that the multilingual capabilities of MLLMs are largely inherited from the underlying language model. Both models share the same LLM, indicating that a strong multilingual LLM forms the foundation for effective multilingual performance in MLLMs.

\subsection{Video Understanding}

\begin{table*}[t!]
\centering
\renewcommand{\arraystretch}{0.95}
\setlength\tabcolsep{5pt}
\newcommand{\VMME}{\makecell{Video-MME\\(wo / w sub)}}
\newcommand{\MVB}{\makecell{MVBench}}
\newcommand{\MMBV}{\makecell{MMBench-Video\\(val)}}
\newcommand{\MLVU}{\makecell{MLVU\\(M-Avg)}}
\newcommand{\LVB}{\makecell{LongVideoBench\\(val total)}}
\newcommand{\CG}{\makecell{CG-Bench\\(long / clue acc.)}}
\newcommand{\lsp}{--\ \ \ \ \ \ }
\newcommand{\rsp}{\ \ \ \ \ --~}
{\fontsize{8}{10}\selectfont 
\begin{tabular}{l|cccccc}
Model Name                                 &    \VMME    & \MVB & \MMBV & \MLVU  & \LVB & \CG         \\
\hline
InternVL2-1B~\cite{chen2024far}            & 42.9 / 45.4 & 57.5 & 1.14  & 51.6  & 43.3  & --          \\
\rowcolor{gray!15}
InternVL2.5-1B                             & 50.3 / 52.3 & 64.3 & 1.36  & 57.3  & 47.9  & --          \\
Qwen2-VL-2B~\cite{wang2024qwen2vl}         & 55.6 / 60.4 & 63.2 & --    & --    & --    & --          \\
InternVL2-2B~\cite{chen2024far}            & 46.2 / 49.1 & 60.2 & 1.30  & 54.3  & 46.0  & --          \\
\rowcolor{gray!15}
InternVL2.5-2B                             & 51.9 / 54.1 & 68.8 & 1.44  & 61.4  & 52.0  & --          \\
\hline
InternVL2-4B~\cite{chen2024far}            & 53.9 / 57.0 & 64.0 & 1.45  & 59.9  & 53.0  & --          \\
\rowcolor{gray!15}
InternVL2.5-4B                             & 62.3 / 63.6 & 71.6 & 1.73  & 68.3  & 55.2  & --          \\
\hline
VideoChat2-HD~\cite{li2023videochat}       & 45.3 / 55.7 & 62.3 & 1.22  & 47.9  & --    & --          \\
MiniCPM-V-2.6~\cite{yao2024minicpm}        & 60.9 / 63.6 & --   & 1.70  & --    & 54.9  & --          \\
LLaVA-OneVision-7B~\cite{li2024llavaov}    & 58.2 / \lsp & 56.7 & --    & --    & --    & --          \\
Qwen2-VL-7B~\cite{wang2024qwen2vl}         & 63.3 / 69.0 & 67.0 & 1.44  & --    & 55.6  & --          \\
InternVL2-8B~\cite{chen2024far}            & 56.3 / 59.3 & 65.8 & 1.57  & 64.0  & 54.6  & --          \\
\rowcolor{gray!15}
InternVL2.5-8B                             & 64.2 / 66.9 & 72.0 & 1.68  & 68.9  & 60.0  & --          \\
\hline
InternVL2-26B~\cite{chen2024far}           & 57.0 / 60.2 & 67.5 & 1.67  & 64.2  & 56.1  & --          \\
\rowcolor{gray!15}
InternVL2.5-26B                            & 66.9 / 69.2 & 75.2 & 1.86  & 72.3  & 59.9  & --          \\
Oryx-1.5-32B~\cite{liu2024oryx}            & 67.3 / 74.9 & 70.1 & 1.52  & 72.3  & --    & --          \\  
VILA-1.5-40B~\cite{lin2024vila}            & 60.1 / 61.1 & --   & 1.61  & 56.7  & --    & --          \\
InternVL2-40B~\cite{chen2024far}           & 66.1 / 68.6 & 72.0 & 1.78  & 71.0  & 60.6  & --          \\
\rowcolor{gray!15}
InternVL2.5-38B                            & 70.7 / 73.1 & 74.4 & 1.82  & 75.3  & 63.3  & --          \\
\hline
GPT-4V/4T~\cite{openai2023gpt4}            & 59.9 / 63.3 & 43.7 & 1.53  & 49.2  & 59.1  & --          \\
GPT-4o-20240513~\cite{gpt4v}               & 71.9 / 77.2 & --   & 1.63  & 64.6  & 66.7  & --          \\
GPT-4o-20240806~\cite{gpt4v}               & --          & --   & 1.87  & --    & --    & 41.8 / 58.3 \\
Gemini-1.5-Pro~\cite{reid2024gemini1_5}    & 75.0 / 81.3 & --   & 1.30  & --    & 64.0  & 40.1 / 56.4 \\
VideoLLaMA2-72B~\cite{cheng2024videollama2}& 61.4 / 63.1 & 62.0 & --    & --    & --    & --          \\
LLaVA-OneVision-72B~\cite{li2024llavaov}   & 66.2 / 69.5 & 59.4 & --    & 66.4  & 61.3  & --          \\
Qwen2-VL-72B~\cite{wang2024qwen2vl}        & 71.2 / 77.8 & 73.6 & 1.70  & --    & --    & 41.3 / 56.2 \\  
InternVL2-Llama3-76B~\cite{chen2024far}    & 64.7 / 67.8 & 69.6 & 1.71  & 69.9  & 61.1  & --          \\
\rowcolor{gray!15}
InternVL2.5-78B                            & 72.1 / 74.0 & 76.4 & 1.97  & 75.7  & 63.6  & 42.2 / 58.5 \\ 
\end{tabular}
}
\caption{\textbf{Comparison of video understanding performance.}
We evaluate InternVL's video understanding capabilities across 6 benchmarks.
For Video-MME~\cite{fu2024video}, MMBench-Video~\cite{fang2024mmbench}, MLVU~\cite{MLVU}, and LongVideoBench~\cite{wu2024longvideobench}, we test with four different settings: 16, 32, 48, and 64 frames, and report the maximum results. For MVBench~\cite{li2024mvbench}, we conduct testing using 16 frames. For CG-Bench~\cite{anonymous2024cgbench}, we use 32 frames.
}
\label{tab:benchmark_video}
\end{table*}

\subsubsection{Benchmarks}

\textbf{Video-MME}~\cite{fu2024video}: Video-MME is a benchmark for evaluating MLLMs in full-spectrum video analysis. It features a wide variety of video types across multiple domains and durations, with multimodal inputs including video, subtitles, and audio. 
For this benchmark, we test with four settings: 16, 32, 48, and 64 frames, and report the maximum results. We report results for both ``with subtitle'' and ``without subtitle'' settings.

\textbf{MVBench}~\cite{li2024mvbench}: MVBench is a video understanding benchmark designed to comprehensively evaluate the temporal awareness of MLLMs in the open world. It covers 20 challenging video tasks, ranging from perception to cognition, which cannot be effectively solved using a single frame. We test this benchmark using 16 frames.

\textbf{MMBench-Video}~\cite{fang2024mmbench}: MMBench-Video is a quantitative benchmark for evaluating MLLMs' video understanding and temporal reasoning skills, covering diverse domains, multi-shot long videos, and features like hallucination, commonsense reasoning, and temporal reasoning. For this benchmark, we test with four different settings: 16, 32, 48, and 64 frames, and report the maximum scores.

\textbf{MLVU}~\cite{MLVU}: MLVU is a comprehensive benchmark designed to evaluate MLLMs in long video understanding tasks, featuring videos ranging from 3 minutes to 2 hours.
It includes nine different evaluation tasks divided into three categories: holistic understanding, single-detail understanding, and multi-detail understanding. We evaluate four settings: 16, 32, 48, and 64 frames, and report the highest ``M-Avg'' results.

\textbf{LongVideoBench}~\cite{wu2024longvideobench}: LongVideoBench focuses on referring reasoning tasks that involve long-frame inputs, requiring the model to accurately retrieve and reason about detailed multimodal information based on referring queries. We test four settings—16, 32, 48, and 64 frames—and report the best results on the validation set.

\textbf{CG-Bench}~\cite{anonymous2024cgbench}: CG-Bench is a benchmark for evaluating long video understanding in MLLMs. Unlike existing benchmarks, it focuses on models' ability to retrieve relevant clues for answering questions. It includes 1,219 curated videos and over 12,000 question-answer pairs. Two novel clue-based evaluation methods are introduced to assess genuine video understanding. We test this benchmark using 32 frames.

\subsubsection{Evaluation Results}

Video understanding is vital for assessing MLLMs' ability to process temporal and multimodal information. To evaluate this comprehensively, we tested six benchmarks: Video-MME~\cite{fu2024video}, MVBench~\cite{li2024mvbench}, MMBench-Video~\cite{fang2024mmbench}, MLVU~\cite{MLVU}, LongVideoBench~\cite{wu2024longvideobench}, and CG-Bench~\cite{anonymous2024cgbench}, covering diverse tasks from short video comprehension to long video reasoning.

\begin{wrapfigure}{r}{0.44\linewidth}
    \centering
    \vspace{-1.5em}
    \includegraphics[width=0.97\linewidth]{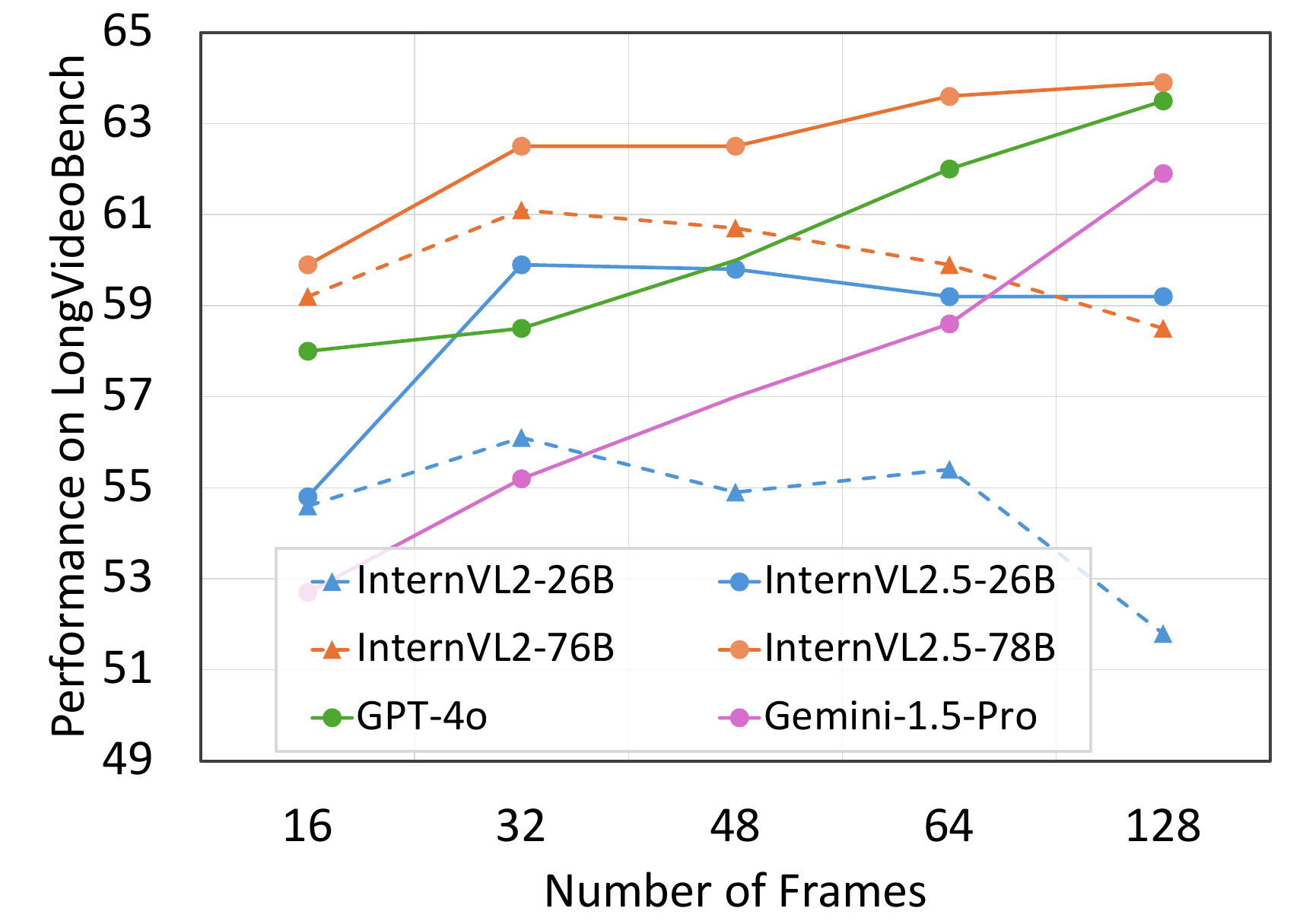}
    \caption{{Performance on LongVideoBench with varying input video frames.}}
    \label{fig:video_frame_scaling}
\end{wrapfigure}

As shown in Table~\ref{tab:benchmark_video}, InternVL 2.5 achieves consistent improvements over InternVL 2.0 across all benchmarks. For example, our smallest model, InternVL2.5-1B improves Video-MME scores from 42.9/45.4 to 50.3/52.3 and MVBench from 57.5 to 64.3. 
Moreover, we find that \emph{InternVL 2.5 demonstrates better scalability when handling increasing input frames compared to its predecessor}, as shown in Figure~\ref{fig:video_frame_scaling}.
We attribute these improvements to two key enhancements:
(1) The inclusion of more high-quality video data, which has significantly enhanced the model’s video understanding capabilities.
(2) Adjusting the training frame sampling strategy from 4–24 to 8–32 frames (as shown in Figure~\ref{fig:dataset_configuration}(c)) enhanced the model's ability to process richer video information. Consequently, while InternVL 2.0 models typically perform best at 16 or 32 frames but degrade with more input frames, InternVL 2.5 could benefit from increasing input frames, showing better scalability for long video understanding.

Our largest model, InternVL2.5-78B, achieves leading performance among open-source models and approaches the performance of closed-source systems. Compared to open-source models, InternVL2.5-78B surpasses Qwen2-VL-72B on MVBench (76.4 \emph{vs.} 73.6) and MMBench-Video (1.97 \emph{vs.} 1.70), though its Video-MME score with subtitles is slightly lower (74.0 \emph{vs.} 77.8).  
Against closed-source models like GPT-4o~\cite{gpt4v} and Gemini-1.5-Pro~\cite{li2024miniGemini}, InternVL2.5-78B demonstrates competitive performance. On Video-MME, it scores 72.1/74.0, closely matching GPT-4o (71.9/77.2) and Gemini-1.5-Pro (75.0/81.3). However, on LongVideoBench, it achieves 63.6, slightly trailing Gemini-1.5-Pro (64.0) and GPT-4o (66.7). This highlights the remaining challenges in long video understanding for open-source models, indicating room for further improvement.

\section{Evaluation on Language Capability}

\begin{table*}[!t]
\centering
{\fontsize{8}{10}\selectfont 
\renewcommand{\arraystretch}{0.95}
\setlength\tabcolsep{6pt}
\newcommand{\GPT}{\rotatebox{90}{\makecell{GPT-4o}}}
\newcommand{\Claude}{\rotatebox{90}{\makecell{Claude 3.5 Sonnet}}}
\newcommand{\QwenS}{\rotatebox{90}{\makecell{Qwen-7B-Chat}}}
\newcommand{\QwenF}{\rotatebox{90}{\makecell{Qwen-14B-Chat}}}
\newcommand{\ILMTTB}{\rotatebox{90}{\makecell{InternLM2-1.8B-Chat}}}
\newcommand{\ILMTDFTB}{\rotatebox{90}{\makecell{InternLM2.5-1.8B-Chat}}}
\newcommand{\ILMTS}{\rotatebox{90}{\makecell{InternLM2-7B-Chat}}}
\newcommand{\ILMTDFS}{\rotatebox{90}{\makecell{InternLM2.5-7B-Chat}}}
\newcommand{\ILMTTTB}{\rotatebox{90}{\makecell{InternLM2-20B-Chat}}}
\newcommand{\ILMTDFTTB}{\rotatebox{90}{\makecell{InternLM2.5-20B-Chat}}}
\newcommand{\IVLTTB}{\rotatebox{90}{\makecell{InternVL2-2B}}}
\newcommand{\IVLTDFTB}{\rotatebox{90}{\makecell{InternVL2.5-2B}}}
\newcommand{\IVLTEB}{\rotatebox{90}{\makecell{InternVL2-8B}}}
\newcommand{\IVLTDFEB}{\rotatebox{90}{\makecell{InternVL2.5-8B}}}
\newcommand{\IVLTTSB}{\rotatebox{90}{\makecell{InternVL2-26B}}}
\newcommand{\IVLTDFTSB}{\rotatebox{90}{\makecell{InternVL2.5-26B}}}
\newcommand{\lsp}{--\ \ \ \ \ \ }
\newcommand{\rsp}{\ \ \ \ \ --~}
\begin{tabular}{lc|cccc|ccc|cccc}
Dataset          & Settings & \ILMTTB & \IVLTTB & \ILMTDFTB & \IVLTDFTB & \ILMTDFS & \IVLTEB & \IVLTDFEB & \ILMTTTB & \IVLTTSB & \ILMTDFTTB & \IVLTDFTSB \\
\hline
MMLU             & 5-shot   & 47.3    & 46.4    & 50.5      & 52.6      & 72.8     & 73.2    & 74.6      & 66.5     & 68.2     & 73.3       & 76.6       \\
CMMLU            & 5-shot   & 46.1    & 47.1    & 62.7      & 57.0      & 78.2     & 79.2    & 78.7      & 64.7     & 68.1     & 79.4       & 81.9       \\
C-Eval           & 5-shot   & 48.6    & 48.6    & 60.4      & 56.2      & 77.9     & 80.1    & 79.7      & 61.8     & 67.7     & 80.2       & 83.8       \\
GAOKAO           & 0-shot   & 33.1    & 32.3    & 54.7      & 52.6      & 78.7     & 75.0    & 77.3      & 63.5     & 62.3     & 81.0       & 86.9       \\
\hline
TriviaQA         & 0-shot   & 37.3    & 31.5    & 32.3      & 31.2      & 64.0     & 62.0    & 63.4      & 61.8     & 61.8     & 67.3       & 69.0       \\
NaturalQuestions & 0-shot   & 15.3    & 13.2    & 10.1      & 11.8      & 21.1     & 28.1    & 29.4      & 23.6     & 28.8     & 21.3       & 36.1       \\
C3               & 0-shot   & 75.8    & 76.9    & 61.4      & 78.0      & 88.1     & 94.2    & 94.7      & 92.2     & 93.2     & 94.0       & 95.8       \\
RACE-High        & 0-shot   & 74.0    & 72.6    & 78.5      & 77.4      & 90.5     & 90.8    & 90.8      & 86.2     & 86.5     & 91.3       & 92.2       \\
\hline
WinoGrande       & 0-shot   & 56.5    & 58.7    & 56.9      & 59.1      & 84.9     & 85.9    & 83.5      & 76.4     & 79.9     & 86.4       & 87.9       \\
HellaSwag        & 0-shot   & 57.9    & 53.7    & 76.2      & 68.2      & 94.8     & 94.9    & 94.1      & 85.3     & 87.5     & 95.9       & 95.8       \\
BBH              & 0-shot   & 37.9    & 36.3    & 43.4      & 40.9      & 73.1     & 72.7    & 73.4      & 70.1     & 69.8     & 78.4       & 78.9       \\
\hline
GSM8K            & 4-shot   & 42.7    & 40.7    & 53.3      & 55.1      & 85.1     & 75.6    & 77.8      & 80.7     & 80.0     & 88.5       & 82.9       \\
MATH             & 4-shot   & 11.0    & 7.0     & 39.5      & 33.5      & 60.6     & 39.5    & 49.9      & 34.9     & 35.5     & 54.7       & 53.7       \\
TheoremQA        & 0-shot   & 13.9    & 12.3    & 11.4      & 12.0      & 23.4     & 15.6    & 23.8      & 22.1     & 15.3     & 23.9       & 15.4       \\
\hline
HumanEval        & 4-shot   & 34.8    & 32.3    & 41.5      & 52.4      & 74.4     & 69.5    & 75.0      & 71.3     & 67.1     & 69.5       & 68.9       \\
MBPP             & 3-shot   & 40.9    & 33.1    & 42.8      & 50.6      & 63.0     & 58.8    & 68.5      & 70.8     & 66.2     & 70.0       & 72.0       \\
MBPP-CN          & 0-shot   & 28.2    & 23.4    & 33.8      & 34.2      & 51.6     & 48.2    & 55.2      & 55.8     & 54.2     & 61.0       & 61.6       \\
\hline
Average          &   --     & 41.3    & 39.2    & 47.6      & 48.4      & 69.5     & 67.2    & 70.0      & 64.0     & 64.2     & 71.5       & 72.9       \\
Gain             &   --     & --      & \cellcolor{red!15}(-2.1)  & --        & \cellcolor{green!15}(+0.8)    & --       & \cellcolor{red!15}(-2.3)  & \cellcolor{green!15}(+0.5)    & --       & \cellcolor{green!15}(+0.2)   & --         & \cellcolor{green!15}(+1.4)     \\

\end{tabular}}
\vspace{2mm}
\caption{\textbf{Comparison of language capabilities across multiple benchmarks.}
These results were obtained using the OpenCompass toolkit for testing.
Training InternVL 2.0 models led to a decline in pure language capabilities. InternVL 2.5 addresses this by collecting more high-quality open-source data and filtering out low-quality data, achieving better preservation of pure language performance.
}
\label{tab:language_model_comparison}
\end{table*}

To thoroughly assess the language capabilities of LLMs and MLLMs, we evaluate their performance across five core dimensions using a diverse set of datasets. These benchmarks encompass tasks like comprehensive examination, language and knowledge, reasoning, mathematics, and coding.

\subsection{Benchmarks}

\textbf{Comprehensive Examination}.  
We conduct a thorough evaluation of LLMs and MLLMs using various exam-related datasets: 
(1)~\textbf{MMLU}~\cite{hendrycks2020measuring} includes 57 subtasks covering diverse topics such as humanities, social sciences, and STEM, evaluated with a 5-shot approach. 
(2)~\textbf{CMMLU}~\cite{li2023cmmlu}, focused on a Chinese context, features 67 subtasks spanning general and Chinese-specific domains, also tested in a 5-shot setting.
(3)~\textbf{C-Eval}~\cite{huang2023ceval} contains 52 subtasks across four difficulty levels, evaluated in a 5-shot setting.
(4)~\textbf{GAOKAO-Bench}~\cite{Zhang2023gaokao}, derived from Chinese college entrance exams, offers comprehensive coverage of both subjective and objective question types, with objective questions evaluated in a 0-shot setting.

\textbf{Language and Knowledge}.  
For language and knowledge-based assessments, we use a range of datasets designed to test the capabilities: 
(1)~\textbf{TriviaQA}~\cite{joshi2017triviaqa}, which includes both reading comprehension and open-domain QA tasks with multiple answers per question, evaluated in a 0-shot setting.
(2)~\textbf{NaturalQuestions}~\cite{naturalquestion}, featuring user-generated questions validated by experts, also evaluated in a 0-shot manner.
(3)~\textbf{C3}~\cite{sun2019investigating}, a free-form multiple-choice Chinese machine reading comprehension dataset, with 0-shot results reported.
(4)~\textbf{RACE}~\cite{lai2017race}, a reading comprehension dataset containing English exam questions for Chinese middle and high school students aged 12 to 18, with results reported for the high school subset in a 0-shot setting.

\textbf{Reasoning}.  
To measure reasoning capabilities, we use datasets like 
(1)~\textbf{WinoGrande}~\cite{sakaguchi2020winogrande}, which tests commonsense reasoning through 44,000 multiple-choice questions requiring pronoun disambiguation, evaluated in a 0-shot setting. 
(2)~\textbf{HellaSwag}~\cite{zellers2019hellaswag} challenges models with natural language inference scenarios and four outcome options, demanding selection of the most logical conclusion, also evaluated in a 0-shot manner. 
(3)~\textbf{BigBench Hard (BBH)}~\cite{suzgun2023bigbench} comprises 23 tasks specifically chosen for their difficulty in surpassing human performance, further evaluating reasoning depth, with 0-shot results reported.

\textbf{Mathematics}.  
In the domain of mathematics, 
(1)~\textbf{GSM8K-Test}~\cite{cobbe2021training} offers approximately 1,300 elementary-level situational problems, evaluated in a 4-shot setting. 
(2)~\textbf{MATH}~\cite{DBLP:conf/nips/HendrycksBKABTS21} presents 12,500 high school competition-level problems across subjects like algebra and calculus, each with detailed solutions, also evaluated in a 4-shot manner. 
(3)~\textbf{TheoremQA}~\cite{DBLP:conf/emnlp/ChenYKLWMXWX23} introduces 800 STEM-focused problems requiring theorem application in fields like mathematics, physics, and finance, with 0-shot results reported.

\textbf{Coding}.  
To evaluate coding capabilities, we employ the following benchmarks:
(1)~\textbf{HumanEval}~\cite{chen2021evaluating}: This benchmark includes 164 Python programming tasks, each paired with detailed specifications, serving as a standard for assessing coding performance. It is evaluated in a 4-shot setting.
(2)~\textbf{MBPP}~\cite{austin2021program}: Comprising 974 entry-level programming tasks, MBPP covers a wide range of challenges, from simple arithmetic problems to more complex sequence definitions, evaluated in a 3-shot setting. 
(3)~\textbf{MBPP-CN}~\cite{austin2021program}: A Chinese adaptation of MBPP designed to assess multilingual programming capabilities. This extension broadens the evaluation scope to include linguistic and contextual diversity, with 0-shot results reported.

\subsection{Evaluation Results}

In the development of MLLMs, maintaining strong pure language capabilities remains critically important. Following the approach of InternLM2~\cite{cai2024internlm2}, we conducted a comprehensive evaluation of our models' performance across 17 pure language benchmarks using the OpenCompass toolkit~\cite{opencompass2023}. These benchmarks are categorized into five major groups, providing a thorough assessment of the models' pure language abilities.

The results show that InternVL 2.0 demonstrates a slight decline in pure language performance compared to its foundational LLM counterparts. For example, InternVL2-2B achieved an average score of 39.2, a decrease of 2.1 points compared to InternLM2-1.8B-Chat. Similarly, InternVL2-8B scored an average of 67.2, 2.3 points lower than InternLM2.5-7B-Chat. 

To address this issue, we curated a large collection of high-quality open-source pure language instruction data and applied rigorous filtering pipelines to eliminate low-quality samples, thereby enhancing the overall data quality. These improvements in InternVL 2.5 have effectively mitigated the decline in language performance, enabling the model to match or even surpass the original LLM in several tasks. This demonstrates that supplementing and optimizing with high-quality language data can not only preserve MLLM's pure language capabilities but also establish a stronger foundation for multimodal tasks.

\section{Evaluation on Vision Capability}

In this section, we present a comprehensive evaluation of the vision encoder’s performance across various domains and tasks. The evaluation is divided into two key categories: 
(1) \emph{image classification}, representing global-view semantic quality, and 
(2) \emph{semantic segmentation}, capturing local-view semantic quality. This approach allows us to assess the representation quality of InternViT across its successive version updates.

\subsection{Image Classification}

\subsubsection{Benchmarks}

We assess the global-view semantic quality of InternViT through a comprehensive evaluation on diverse image classification datasets.

\textbf{ImageNet-1K} \cite{deng2009imagenet}: A widely-used large-scale dataset containing over 1 million images across 1,000 classes, commonly used for benchmarking image classification models.

\textbf{ImageNet-ReaL} \cite{beyer2020imagenetreal}: A re-labeled version of ImageNet's validation set, providing multi-label annotations that are more accurate and robust, following an enhanced labeling protocol.

\textbf{ImageNet-V2} \cite{recht2019imagenetv2}: A dataset designed to evaluate the robustness of models trained on ImageNet-1K, featuring new test images collected using the original ImageNet methodology.

\textbf{ImageNet-A} \cite{hendrycks2021imagenet_a}: A challenging dataset of naturally occurring, unmodified images that are often misclassified by ResNet models. It highlights the limitations of models when exposed to adversarially difficult examples in real-world settings.

\textbf{ImageNet-R} \cite{hendrycks2021imagenet_r}: A rendition dataset with 30K images across 200 ImageNet classes, composed of art, sketches, toys, sculptures, and other creative representations. It assesses the robustness of models in recognizing abstract renditions of common objects.

\textbf{ImageNet-Sketch} \cite{wang2019imagenet_sketch}: This dataset contains 51K sketch images, with approximately 50 sketches per ImageNet class. It is constructed via Google Image queries using the class name followed by ``sketch of,'' testing a model’s ability to generalize to abstract, hand-drawn representations.

\subsubsection{Settings}

In this study, two evaluation methods, linear probing~\cite{chen2020simple} and attention pooling probing, are employed to assess the performance of the InternViT models:

\setlist[itemize]{left=0pt}
\begin{itemize}

\item \textbf{Linear Probing}~\cite{chen2020simple}: This method involves freezing the pre-trained model and training only a linear classifier on top. It evaluates the quality of the learned features without updating the backbone, providing insights into how effectively the pre-trained model captures semantic information usable by a simple linear classifier in downstream tasks like image classification.

\item \textbf{Attention Pooling Probing}: In contrast, attention pooling probing evaluates the model by adding an attention pooling layer on top of the frozen features. This approach allows the vision encoder to retain richer information in the final layer, as attention pooling can dynamically select task-relevant features for classification without interference from unrelated information. 

\end{itemize}

For both experiments, we use ImageNet-1K~\cite{deng2009imagenet} as the training set and evaluate the models on the ImageNet-1K validation set along with several ImageNet variants (\ie, ImageNet-ReaL~\cite{beyer2020imagenetreal}, ImageNet-V2~\cite{recht2019imagenetv2}, ImageNet-A~\cite{hendrycks2021imagenet_a}, ImageNet-R~\cite{hendrycks2021imagenet_r}, and ImageNet-Sketch~\cite{wang2019imagenet_sketch}) to benchmark their domain generalization capabilities.

The models are trained using SGD as the optimizer, with a peak learning rate of 0.2, a momentum of 0.9, and no weight decay. A cosine learning rate decay schedule is applied over 10 training epochs, with 1 warmup epoch. We use input resolutions of 448$\times$448, with a patch size of 14 and a total batch size of 1024. Data augmentation techniques, such as random resized cropping and horizontal flipping, are employed during training. The code and logs of these classification experiments will be released on our GitHub repository\footnote{https://github.com/OpenGVLab/InternVL/tree/main/classification}.

\begin{table*}[t!]
\centering
{\fontsize{8}{10}\selectfont 
\renewcommand{\arraystretch}{1.1}
\setlength\tabcolsep{1.3pt}
\begin{tabular}{l|c|ccccccc|ccccccc|c}
                          &                        & \multicolumn{7}{c|}{Linear Probing} & \multicolumn{7}{c|}{Attention Pooling Probing} &   \\

\multirow{-2}{*}{Model Name}& \multirow{-2}{*}{res.} & IN-1K & IN-ReaL & IN-V2 & IN-A & IN-R & IN-Ske & avg.                      & IN-1K & IN-ReaL & IN-V2 & IN-A & IN-R & IN-Ske & avg.                     & \multirow{-2}{*}{$\Delta$} \\
\hline
\textcolor{gray}{InternViT-6B-224px}        & \textcolor{gray}{224}                    & \textcolor{gray}{88.2}  & \textcolor{gray}{90.4}    & \textcolor{gray}{79.9}  & \textcolor{gray}{77.5} & \textcolor{gray}{89.8} & \textcolor{gray}{69.1}   & \textcolor{gray}{82.5}                      & \textcolor{gray}{89.2}  & \textcolor{gray}{91.1}    & \textcolor{gray}{82.3}  & \textcolor{gray}{84.7} & \textcolor{gray}{93.1} & \textcolor{gray}{72.7}   & \textcolor{gray}{85.5}                     & \textcolor{gray}{3.0}    \\
InternViT-6B-224px        & 448                    & 87.8  & 90.2    & 79.8  & 77.2 & 87.1 & 65.8   & \cellcolor{gray!15} 81.3  & 88.8  & 91.0    & 82.0  & 85.4 & 91.3 & 70.5   & \cellcolor{gray!15} 84.8 & 3.5    \\
InternViT-6B-448px-V1.0   & 448                    & 87.0  & 90.0    & 78.8  & 77.2 & 85.5 & 65.1   & \cellcolor{red!15} 80.6 & 88.7  & 91.0    & 82.0  & 88.7 & 92.8 & 72.0   & \cellcolor{green!15} 85.9  & 5.3    \\
InternViT-6B-448px-V1.2   & 448                    & 87.0  & 89.9    & 78.5  & 77.1 & 83.9 & 59.7   & \cellcolor{red!15} 79.4 & 88.6  & 91.1    & 82.0  & 88.7 & 92.7 & 71.6   & \cellcolor{green!15} 85.8  & 6.4    \\
InternViT-6B-448px-V1.5   & 448                    & 86.5  & 89.9    & 78.1  & 69.8 & 82.9 & 60.1   & \cellcolor{red!15} 77.9 & 88.4  & 91.2    & 81.6  & 86.0 & 92.2 & 70.9   & \cellcolor{green!15} 85.1  & 7.2    \\
InternViT-6B-448px-V2.5   & 448                    & 86.6  & 90.1    & 77.8  & 73.7 & 82.7 & 60.0   & \cellcolor{red!15} 78.5 & 88.3  & 91.2    & 81.3  & 86.9 & 92.4 & 70.8   & \cellcolor{green!15} 85.2  & 6.7    \\
\end{tabular}
}
\caption{\textbf{Image classification performance across different versions of InternViT.}
We use IN-1K~\cite{deng2009imagenet} for training and evaluate on the IN-1K validation set as well as multiple ImageNet variants, including IN-ReaL~\cite{beyer2020imagenetreal}, IN-V2~\cite{recht2019imagenetv2}, IN-A~\cite{hendrycks2021imagenet_a}, IN-R~\cite{hendrycks2021imagenet_r}, and IN-Sketch~\cite{wang2019imagenet_sketch}.
Results are reported for both linear probing and attention pooling probing methods, with average accuracy for each method. 
$\Delta$ represents the performance gap between attention pooling probing and linear probing, where a larger $\Delta$ suggests a shift from learning simple linear features to capturing more complex, nonlinear semantic representations.
    }
\label{tab:benchmark_in_cls}
\end{table*}

\subsubsection{Evaluation Results}

As shown in Table~\ref{tab:benchmark_in_cls}, the results reveal an interesting trend across the version updates of InternViT: as the model progresses, the performance of linear probing declines substantially, with all versions showing an average below the gray baseline. 
In contrast, attention pooling probing consistently outperforms the gray baseline despite some fluctuations. 
This results in a growing trend in the average score difference (from 3.5 to 6.7), denoted as $\Delta$, across successive InternViT versions.

This suggests that features in the model’s final layer become less linearly separable, likely as representations evolve to capture more complex, open-ended semantic information.
The attention pooling mechanism effectively selects relevant features from this enriched representation space, offsetting challenges from reduced linear separability.
Additionally, these findings imply that InternViT maintains key pre-training attributes through iterative updates without catastrophic forgetting. With each version, its representations grow more diverse, capturing open-set semantics and enhancing generalization—an advantage particularly valuable for MLLMs requiring high abstraction for real-world tasks.

\subsection{Semantic Segmentation}

\subsubsection{Benchmarks}

We evaluate the local-view semantic quality of InternViT using two representative semantic segmentation datasets, ADE20K and COCO-Stuff-164K.

\textbf{ADE20K}~\cite{zhou2017ade20k}: A comprehensive dataset containing over 20,000 images with annotations across 150 object and background categories, widely used for scene parsing. It provides detailed pixel-level labels for both objects and parts, facilitating a range of fine-grained segmentation tasks.

\textbf{COCO-Stuff-164K}~\cite{caesar2018cocostuff}: An extension of the original COCO images with pixel-level annotations, adding 91 ``stuff'' classes (like grass and sky) to 80 ``thing'' categories (like people and cars), covering a total of 172 classes. With these comprehensive labels, the dataset supports tasks in scene parsing and semantic segmentation, enabling richer context understanding in image analysis.

\subsubsection{Settings}

In this study, three evaluation methods—linear probing, head tuning, and full tuning—are employed to assess the performance of the InternViT models on semantic segmentation tasks:

\setlist[itemize]{left=0pt}
\begin{itemize}
\item \textbf{Linear Probing}: Linear probing applies a frozen backbone with a linear segmentation head, offering insight into the linear separability of learned features. This method provides a baseline for evaluating pixel-level semantic information with minimal adaptation, though it may not fully capture the encoder’s capacity for complex features.

\item \textbf{Head Tuning}: In head tuning, the InternViT is frozen while the UperNet~\cite{xiao2018upernet} head remains trainable, allowing the model to utilize a stronger head to reduce its dependence on linear separability.  This setup mitigates the decline in linear separability caused by the complex, open-ended features, enabling a more precise evaluation of the vision encoder's capabilities.

\item \textbf{Full Tuning}: Full tuning involves making both the InternViT backbone and the UperNet~\cite{xiao2018upernet} segmentation head trainable, allowing the model to adapt all layers for the target task and minimizing reliance on pre-existing linear separability.  This setup provides an alternative perspective for evaluating the vision encoder’s capacity to extract visual features.

\end{itemize}

We use AdamW~\cite{loshchilov2017adamw} with a peak learning rate of 4e-5 and a polynomial decay schedule. Layer-wise learning rate decay (0.95) is applied in full tuning. Weight decay is set to 0.05 for both head and full tuning, and none for linear probing. The input resolution is 504$\times$504, with a patch size of 14 and a batch size of 16. Training consists of 1.5K warmup iterations and 80K total iterations. A drop path rate of 0.4 is applied in full tuning. We utilize default data augmentation from MMSegmentation~\cite{contributors2020mmsegmentation}.
All the code and logs related to these experiments will be released on GitHub\footnote{https://github.com/OpenGVLab/InternVL-MMDetSeg}.

\begin{table*}[t!]
\centering
{\fontsize{8}{10}\selectfont 
\renewcommand{\arraystretch}{1.1}
\setlength\tabcolsep{5.2pt}
\begin{tabular}{l|ccc|ccc|ccc|cc}
                          & \multicolumn{3}{c|}{Linear Probing} & \multicolumn{3}{c|}{Head Tuning (UperNet)} & \multicolumn{3}{c|}{Full Tuning (UperNet)} &   &   \\

\multirow{-2}{*}{Model Name} & ADE20K & COCO & avg.                     & ADE20K & COCO & avg.                    & ADE20K & COCO & avg. & \multirow{-2}{*}{$\Delta_1$} & \multirow{-2}{*}{$\Delta_2$} \\
\hline
InternViT-6B-224px           & 47.2   & 42.8 & \cellcolor{gray!15}45.0  & 54.9   & 48.9 & \cellcolor{gray!15}51.9 & 58.9   & 51.6 & 55.3 & 6.9                          & 10.2                             \\
InternViT-6B-448px-V1.0      & 43.6   & 38.5 & \cellcolor{red!15}41.0 & 55.4   & 49.4 & \cellcolor{green!15}52.4  & 58.1   & 51.7 & 54.9 & 11.3                         & 13.9                             \\
InternViT-6B-448px-V1.2      & 40.7   & 36.1 & \cellcolor{red!15}38.4 & 55.2   & 48.8 & \cellcolor{green!15}52.0  & 58.8   & 51.7 & 55.2 & 13.6                         & 16.8                             \\
InternViT-6B-448px-V1.5      & 40.9   & 36.3 & \cellcolor{red!15}38.6 & 55.0   & 49.1 & \cellcolor{green!15}52.0  & 58.8   & 51.5 & 55.2 & 13.4                         & 16.6                             \\
InternViT-6B-448px-V2.5      & 39.4   & 35.6 & \cellcolor{red!15}37.5 & 55.4   & 49.7 & \cellcolor{green!15}52.6  & 58.6   & 51.8 & 55.2 & 15.1                         & 17.7                             \\
\end{tabular}
}
\caption{\textbf{Semantic segmentation performance across different versions of InternViT.}
The models are evaluated on ADE20K~\cite{zhou2017ade20k} and COCO-Stuff-164K~\cite{caesar2018cocostuff} using three configurations: linear probing, head tuning, and full tuning. The table shows the mIoU scores for each configuration and their averages. $\Delta_1$ represents the gap between head tuning and linear probing, while $\Delta_2$ shows the gap between full tuning and linear probing. A larger $\Delta$ value indicates a shift from simple linear features to more complex, nonlinear representations.
    }
\label{tab:benchmark_sem_seg}
\end{table*}

\subsubsection{Evaluation Results}

As shown in Table~\ref{tab:benchmark_sem_seg}, the semantic segmentation performance of InternViT models is evaluated across three configurations—linear probing, head tuning, and full tuning—on ADE20K~\cite{zhou2017ade20k} and COCO-Stuff-164K~\cite{caesar2018cocostuff}. The results reveal distinct trends in how the models' feature representations evolve across version updates.

Linear probing results show a decline in mIoU scores as the model versions progress, with average scores dropping from 45.0 in InternViT-6B-224px to 37.5 in InternViT-6B-448px-V2.5. This indicates that as InternViT updates, the features become less linearly separable, reflecting a shift toward capturing more complex and open-ended information.

In head tuning, the models display a different trend compared to linear probing. All other versions of InternViT surpass the baseline InternViT-6B-224px’s mIoU score of 51.9, showing no performance decline. This leads to increasing $\Delta_1$ values, growing from 6.9 in InternViT-6B-224px to 15.1 in InternViT-6B-448px-V2.5. The rise in $\Delta_1$ suggests that while the features become less linearly separable, their quality remains intact, effectively capturing complex information. Similarly, full tuning yields consistent results, as seen in the $\Delta_2$ values. The increase in $\Delta_2$ from 10.2 in InternViT-6B-224px to 17.7 in InternViT-6B-448px-V2.5 further supports this trend.

Overall, the increasing values of $\Delta_1$ and $\Delta_2$ across model versions highlight the shift from simple, linearly separable features to more complex, nonlinear representations. This evolution aligns with InternViT's growing capability to extract visual information as its versions progress within the development of InternVL. It demonstrates the effectiveness of our ViT incremental learning strategy in enhancing the vision encoder's ability to extract open-ended features.

\section{Conclusion}

\label{sec:conclusion}

In this work, we introduce InternVL 2.5, an advanced open-source multimodal large language model (MLLM) series that builds upon the architecture of InternVL 2.0 with significant improvements in training, testing strategies, and data quality. We systematically explore the relationship between model scaling and performance, analyzing vision encoders, language models, dataset sizes, and test-time configurations. Extensive evaluations on diverse benchmarks demonstrate that InternVL 2.5 achieves competitive performance across tasks such as multi-discipline reasoning, document understanding, video understanding, multilingual processing, \etc. Notably, it is the first open-source MLLM to surpass 70\% on the MMMU benchmark, narrowing the gap between open-source and commercial models like OpenAI o1. By sharing InternVL 2.5 with the community, we hope to contribute a powerful tool for advancing multimodal AI research and applications, and we look forward to seeing future developments building upon this work.

\section*{Acknowledgement}

This work is supported by the National Key R\&D Program of China (No. 2022ZD0160102, 2022ZD0161300), the National Natural Science Foundation of China (No. 62376134, 62372223, U24A20330), the Youth PhD Student Research Project under the National Natural Science Foundation (No. 623B2050), and the China Mobile Zijin Innovation Institute (No. NR2310J7M).

{
    \small
    \bibliographystyle{plain}
    \bibliography{main}

\begin{thebibliography}{100}

\bibitem{abdin2024phi3}
Marah Abdin, Sam~Ade Jacobs, Ammar~Ahmad Awan, Jyoti Aneja, Ahmed Awadallah, Hany Awadalla, Nguyen Bach, Amit Bahree, Arash Bakhtiari, Harkirat Behl, et~al.
\newblock Phi-3 technical report: A highly capable language model locally on your phone.
\newblock {\em arXiv preprint arXiv:2404.14219}, 2024.

\bibitem{acharya2019tallyqa}
Manoj Acharya, Kushal Kafle, and Christopher Kanan.
\newblock Tallyqa: Answering complex counting questions.
\newblock In {\em Proceedings of the AAAI Conference on Artificial Intelligence}, volume~33, pages 8076--8084, 2019.

\bibitem{openai2023gpt4}
Josh Achiam, Steven Adler, Sandhini Agarwal, Lama Ahmad, Ilge Akkaya, Florencia~Leoni Aleman, Diogo Almeida, Janko Altenschmidt, Sam Altman, Shyamal Anadkat, et~al.
\newblock Gpt-4 technical report.
\newblock {\em arXiv preprint arXiv:2303.08774}, 2023.

\bibitem{agentsea_wave_ui}
AgentSea.
\newblock Wave-ui.
\newblock \url{https://huggingface.co/datasets/agentsea/wave-ui}, 2024.

\bibitem{alayrac2022flamingo}
Jean-Baptiste Alayrac, Jeff Donahue, Pauline Luc, Antoine Miech, Iain Barr, Yana Hasson, Karel Lenc, Arthur Mensch, Katherine Millican, Malcolm Reynolds, et~al.
\newblock Flamingo: a visual language model for few-shot learning.
\newblock {\em Advances in Neural Information Processing Systems}, 35:23716--23736, 2022.

\bibitem{amini2019mathqa}
Aida Amini, Saadia Gabriel, Peter Lin, Rik Koncel-Kedziorski, Yejin Choi, and Hannaneh Hajishirzi.
\newblock Mathqa: Towards interpretable math word problem solving with operation-based formalisms.
\newblock {\em arXiv preprint arXiv:1905.13319}, 2019.

\bibitem{anonymous2024cgbench}
Anonymous.
\newblock {CG}-bench: Clue-grounded question answering benchmark for long video understanding.
\newblock In {\em Submitted to The Thirteenth International Conference on Learning Representations}, 2024.
\newblock under review.

\bibitem{claude3series2024}
{Anthropic}.
\newblock The claude 3 model family: Opus, sonnet, haiku.
\newblock \url{https://www.anthropic.com}, 2024.

\bibitem{austin2021program}
Jacob Austin, Augustus Odena, Maxwell Nye, Maarten Bosma, Henryk Michalewski, David Dohan, Ellen Jiang, Carrie Cai, Michael Terry, Quoc Le, et~al.
\newblock Program synthesis with large language models.
\newblock {\em arXiv preprint arXiv:2108.07732}, 2021.

\bibitem{azuma2022scanqa}
Daichi Azuma, Taiki Miyanishi, Shuhei Kurita, and Motoaki Kawanabe.
\newblock Scanqa: 3d question answering for spatial scene understanding.
\newblock In {\em Proceedings of the IEEE/CVF Conference on Computer Vision and Pattern Recognition}, pages 19129--19139, 2022.

\bibitem{ba2016layer}
Jimmy~Lei Ba, Jamie~Ryan Kiros, and Geoffrey~E Hinton.
\newblock Layer normalization.
\newblock {\em arXiv preprint arXiv:1607.06450}, 2016.

\bibitem{bai2021uibert}
Chongyang Bai, Xiaoxue Zang, Ying Xu, Srinivas Sunkara, Abhinav Rastogi, Jindong Chen, et~al.
\newblock Uibert: Learning generic multimodal representations for ui understanding.
\newblock {\em arXiv preprint arXiv:2107.13731}, 2021.

\bibitem{bai2023qwenvl}
Jinze Bai, Shuai Bai, Shusheng Yang, Shijie Wang, Sinan Tan, Peng Wang, Junyang Lin, Chang Zhou, and Jingren Zhou.
\newblock Qwen-vl: A frontier large vision-language model with versatile abilities.
\newblock {\em arXiv preprint arXiv:2308.12966}, 2023.

\bibitem{chinese-ocr}
Ltd. Beijing Anjie Zhihe Technology~Co.
\newblock Chinese-ocr.
\newblock \url{https://huggingface.co/datasets/longmaodata/Chinese-OCR}, 2024.

\bibitem{ben2019vqa}
Asma Ben~Abacha, Sadid~A Hasan, Vivek~V Datla, Dina Demner-Fushman, and Henning M{\"u}ller.
\newblock Vqa-med: Overview of the medical visual question answering task at imageclef 2019.
\newblock In {\em Proceedings of CLEF (Conference and Labs of the Evaluation Forum) 2019 Working Notes}, 2019.

\bibitem{beyer2020imagenetreal}
Lucas Beyer, Olivier~J H{\'e}naff, Alexander Kolesnikov, Xiaohua Zhai, and A{\"a}ron van~den Oord.
\newblock Are we done with imagenet?
\newblock {\em arXiv preprint arXiv:2006.07159}, 2020.

\bibitem{biten2019stvqa}
Ali~Furkan Biten, Ruben Tito, Andres Mafla, Lluis Gomez, Mar{\c{c}}al Rusinol, Ernest Valveny, CV~Jawahar, and Dimosthenis Karatzas.
\newblock Scene text visual question answering.
\newblock In {\em Proceedings of the IEEE/CVF International Conference on Computer Vision}, pages 4291--4301, 2019.

\bibitem{caesar2018cocostuff}
Holger Caesar, Jasper Uijlings, and Vittorio Ferrari.
\newblock Coco-stuff: Thing and stuff classes in context.
\newblock In {\em Proceedings of the IEEE/CVF Conference on Computer Vision and Pattern Recognition}, pages 1209--1218, 2018.

\bibitem{cai2024internlm2}
Zheng Cai, Maosong Cao, Haojiong Chen, Kai Chen, Keyu Chen, Xin Chen, Xun Chen, Zehui Chen, Zhi Chen, Pei Chu, et~al.
\newblock Internlm2 technical report.
\newblock {\em arXiv preprint arXiv:2403.17297}, 2024.

\bibitem{cao2022geoqa_plus}
Jie Cao and Jing Xiao.
\newblock An augmented benchmark dataset for geometric question answering through dual parallel text encoding.
\newblock In {\em Proceedings of the 29th International Conference on Computational Linguistics}, pages 1511--1520, 2022.

\bibitem{CarperAI_openai_summarize_tldr}
CarperAI.
\newblock openai summarize tldr dataset.
\newblock \url{https://huggingface.co/datasets/CarperAI/openai\_summarize\_tldr}, 2023.

\bibitem{chai2024amex}
Yuxiang Chai, Siyuan Huang, Yazhe Niu, Han Xiao, Liang Liu, Dingyu Zhang, Peng Gao, Shuai Ren, and Hongsheng Li.
\newblock Amex: Android multi-annotation expo dataset for mobile gui agents.
\newblock {\em arXiv preprint arXiv:2407.17490}, 2024.

\bibitem{chang2022mapqa}
Shuaichen Chang, David Palzer, Jialin Li, Eric Fosler-Lussier, and Ningchuan Xiao.
\newblock Mapqa: A dataset for question answering on choropleth maps.
\newblock {\em arXiv preprint arXiv:2211.08545}, 2022.

\bibitem{chen2024gui}
Dongping Chen, Yue Huang, Siyuan Wu, Jingyu Tang, Liuyi Chen, Yilin Bai, Zhigang He, Chenlong Wang, Huichi Zhou, Yiqiang Li, et~al.
\newblock Gui-world: A dataset for gui-oriented multimodal llm-based agents.
\newblock {\em arXiv preprint arXiv:2406.10819}, 2024.

\bibitem{chen2024allava}
Guiming~Hardy Chen, Shunian Chen, Ruifei Zhang, Junying Chen, Xiangbo Wu, Zhiyi Zhang, Zhihong Chen, Jianquan Li, Xiang Wan, and Benyou Wang.
\newblock Allava: Harnessing gpt4v-synthesized data for a lite vision-language model.
\newblock {\em arXiv preprint arXiv:2402.11684}, 2024.

\bibitem{chen2022unigeo}
Jiaqi Chen, Tong Li, Jinghui Qin, Pan Lu, Liang Lin, Chongyu Chen, and Xiaodan Liang.
\newblock Unigeo: Unifying geometry logical reasoning via reformulating mathematical expression.
\newblock {\em arXiv preprint arXiv:2212.02746}, 2022.

\bibitem{chen2023shikra}
Keqin Chen, Zhao Zhang, Weili Zeng, Richong Zhang, Feng Zhu, and Rui Zhao.
\newblock Shikra: Unleashing multimodal llm's referential dialogue magic.
\newblock {\em arXiv preprint arXiv:2306.15195}, 2023.

\bibitem{chen2024mmstar}
Lin Chen, Jinsong Li, Xiaoyi Dong, Pan Zhang, Yuhang Zang, Zehui Chen, Haodong Duan, Jiaqi Wang, Yu~Qiao, Dahua Lin, et~al.
\newblock Are we on the right way for evaluating large vision-language models?
\newblock {\em arXiv preprint arXiv:2403.20330}, 2024.

\bibitem{chen2023sharegpt4v}
Lin Chen, Jisong Li, Xiaoyi Dong, Pan Zhang, Conghui He, Jiaqi Wang, Feng Zhao, and Dahua Lin.
\newblock Sharegpt4v: Improving large multi-modal models with better captions.
\newblock {\em arXiv preprint arXiv:2311.12793}, 2023.

\bibitem{chen2024sharegpt4video}
Lin Chen, Xilin Wei, Jinsong Li, Xiaoyi Dong, Pan Zhang, Yuhang Zang, Zehui Chen, Haodong Duan, Bin Lin, Zhenyu Tang, et~al.
\newblock Sharegpt4video: Improving video understanding and generation with better captions.
\newblock {\em arXiv preprint arXiv:2406.04325}, 2024.

\bibitem{chen2021evaluating}
Mark Chen, Jerry Tworek, Heewoo Jun, Qiming Yuan, Henrique Ponde de~Oliveira Pinto, Jared Kaplan, Harri Edwards, Yuri Burda, Nicholas Joseph, Greg Brockman, et~al.
\newblock Evaluating large language models trained on code.
\newblock {\em arXiv preprint arXiv:2107.03374}, 2021.

\bibitem{chen2020simple}
Ting Chen, Simon Kornblith, Mohammad Norouzi, and Geoffrey Hinton.
\newblock A simple framework for contrastive learning of visual representations.
\newblock In {\em The International Conference on Learning Representations}, pages 1597--1607. PMLR, 2020.

\bibitem{DBLP:conf/emnlp/ChenYKLWMXWX23}
Wenhu Chen, Ming Yin, Max Ku, Pan Lu, Yixin Wan, Xueguang Ma, Jianyu Xu, Xinyi Wang, and Tony Xia.
\newblock Theoremqa: {A} theorem-driven question answering dataset.
\newblock In Houda Bouamor, Juan Pino, and Kalika Bali, editors, {\em Proceedings of the 2023 Conference on Empirical Methods in Natural Language Processing, {EMNLP} 2023, Singapore, December 6-10, 2023}, pages 7889--7901. Association for Computational Linguistics, 2023.

\bibitem{long-alpaca}
Yukang Chen, Shengju Qian, Haotian Tang, Xin Lai, Zhijian Liu, Song Han, and Jiaya Jia.
\newblock Longlora: Efficient fine-tuning of long-context large language models.
\newblock {\em arXiv preprint arXiv:2309.12307}, 2023.

\bibitem{chen2024far}
Zhe Chen, Weiyun Wang, Hao Tian, Shenglong Ye, Zhangwei Gao, Erfei Cui, Wenwen Tong, Kongzhi Hu, Jiapeng Luo, Zheng Ma, et~al.
\newblock How far are we to gpt-4v? closing the gap to commercial multimodal models with open-source suites.
\newblock {\em arXiv preprint arXiv:2404.16821}, 2024.

\bibitem{chen2023internvl}
Zhe Chen, Jiannan Wu, Wenhai Wang, Weijie Su, Guo Chen, Sen Xing, Muyan Zhong, Qinglong Zhang, Xizhou Zhu, Lewei Lu, et~al.
\newblock Internvl: Scaling up vision foundation models and aligning for generic visual-linguistic tasks.
\newblock In {\em Proceedings of the IEEE/CVF Conference on Computer Vision and Pattern Recognition}, pages 24185--24198, 2024.

\bibitem{cheng2024seeclick}
Kanzhi Cheng, Qiushi Sun, Yougang Chu, Fangzhi Xu, Yantao Li, Jianbing Zhang, and Zhiyong Wu.
\newblock Seeclick: Harnessing gui grounding for advanced visual gui agents.
\newblock {\em arXiv preprint arXiv:2401.10935}, 2024.

\bibitem{cheng2024videollama2}
Zesen Cheng, Sicong Leng, Hang Zhang, Yifei Xin, Xin Li, Guanzheng Chen, Yongxin Zhu, Wenqi Zhang, Ziyang Luo, Deli Zhao, et~al.
\newblock Videollama 2: Advancing spatial-temporal modeling and audio understanding in video-llms.
\newblock {\em arXiv preprint arXiv:2406.07476}, 2024.

\bibitem{chi2019complicated}
Zewen Chi, Heyan Huang, Heng-Da Xu, Houjin Yu, Wanxuan Yin, and Xian-Ling Mao.
\newblock Complicated table structure recognition.
\newblock {\em arXiv preprint arXiv:1908.04729}, 2019.

\bibitem{chng2019art}
Chee~Kheng Chng, Yuliang Liu, Yipeng Sun, Chun~Chet Ng, Canjie Luo, Zihan Ni, ChuanMing Fang, Shuaitao Zhang, Junyu Han, Errui Ding, et~al.
\newblock Icdar2019 robust reading challenge on arbitrary-shaped text-rrc-art.
\newblock In {\em International Conference on Document Analysis and Recognition}, pages 1571--1576, 2019.

\bibitem{chung2024scaling}
Hyung~Won Chung, Le~Hou, Shayne Longpre, Barret Zoph, Yi~Tay, William Fedus, Yunxuan Li, Xuezhi Wang, Mostafa Dehghani, Siddhartha Brahma, et~al.
\newblock Scaling instruction-finetuned language models.
\newblock {\em Journal of Machine Learning Research}, 25(70):1--53, 2024.

\bibitem{clark2017docqa}
Christopher Clark and Matt Gardner.
\newblock Simple and effective multi-paragraph reading comprehension.
\newblock In {\em Proceedings of the Annual Meeting of the Association for Computational Linguistics}, pages 845--855, 2018.

\bibitem{cobbe2021training}
Karl Cobbe, Vineet Kosaraju, Mohammad Bavarian, Mark Chen, Heewoo Jun, Lukasz Kaiser, Matthias Plappert, Jerry Tworek, Jacob Hilton, Reiichiro Nakano, et~al.
\newblock Training verifiers to solve math word problems.
\newblock {\em arXiv preprint arXiv:2110.14168}, 2021.

\bibitem{conover2023free}
Mike Conover, Matt Hayes, Ankit Mathur, Jianwei Xie, Jun Wan, Sam Shah, Ali Ghodsi, Patrick Wendell, Matei Zaharia, and Reynold Xin.
\newblock Free dolly: Introducing the world’s first truly open instruction-tuned llm.
\newblock {\em Company Blog of Databricks}, 2023.

\bibitem{contributors2020mmsegmentation}
MMSegmentation Contributors.
\newblock Mmsegmentation: Openmmlab semantic segmentation toolbox and benchmark.
\newblock \url{https://github.com/open-mmlab/mmsegmentation}, 2020.

\bibitem{opencompass2023}
OpenCompass Contributors.
\newblock Opencompass: A universal evaluation platform for foundation models.
\newblock \url{https://github.com/open-compass/opencompass}, 2023.

\bibitem{realworldqa}
X.AI Corp.
\newblock Grok-1.5 vision preview: Connecting the digital and physical worlds with our first multimodal model.
\newblock \url{https://x.ai/blog/grok-1.5v}, 2024.

\bibitem{cui2023ultrafeedback}
Ganqu Cui, Lifan Yuan, Ning Ding, Guanming Yao, Wei Zhu, Yuan Ni, Guotong Xie, Zhiyuan Liu, and Maosong Sun.
\newblock Ultrafeedback: Boosting language models with high-quality feedback.
\newblock {\em arXiv preprint arXiv:2310.01377}, 2023.

\bibitem{dai202415m}
Dawei Dai, YuTang Li, YingGe Liu, Mingming Jia, Zhang YuanHui, and Guoyin Wang.
\newblock 15m multimodal facial image-text dataset.
\newblock {\em arXiv preprint arXiv:2407.08515}, 2024.

\bibitem{dai2024nvlm}
Wenliang Dai, Nayeon Lee, Boxin Wang, Zhuolin Yang, Zihan Liu, Jon Barker, Tuomas Rintamaki, Mohammad Shoeybi, Bryan Catanzaro, and Wei Ping.
\newblock Nvlm: Open frontier-class multimodal llms.
\newblock {\em arXiv preprint arXiv:2409.11402}, 2024.

\bibitem{das2017visdial}
Abhishek Das, Satwik Kottur, Khushi Gupta, Avi Singh, Deshraj Yadav, Jos{\'e}~MF Moura, Devi Parikh, and Dhruv Batra.
\newblock Visual dialog.
\newblock In {\em Proceedings of the IEEE/CVF Conference on Computer Vision and Pattern Recognition}, pages 326--335, 2017.

\bibitem{davis2019deep}
Brian Davis, Bryan Morse, Scott Cohen, Brian Price, and Chris Tensmeyer.
\newblock Deep visual template-free form parsing.
\newblock In {\em International Conference on Document Analysis and Recognition}, pages 134--141, 2019.

\bibitem{dehghani2023vit22b}
Mostafa Dehghani, Josip Djolonga, Basil Mustafa, Piotr Padlewski, Jonathan Heek, Justin Gilmer, Andreas~Peter Steiner, Mathilde Caron, Robert Geirhos, Ibrahim Alabdulmohsin, et~al.
\newblock Scaling vision transformers to 22 billion parameters.
\newblock In {\em International Conference on Machine Learning}, pages 7480--7512, 2023.

\bibitem{deitke2024molmo}
Matt Deitke, Christopher Clark, Sangho Lee, Rohun Tripathi, Yue Yang, Jae~Sung Park, Mohammadreza Salehi, Niklas Muennighoff, Kyle Lo, Luca Soldaini, et~al.
\newblock Molmo and pixmo: Open weights and open data for state-of-the-art multimodal models.
\newblock {\em arXiv preprint arXiv:2409.17146}, 2024.

\bibitem{deka2017rico}
Biplab Deka, Zifeng Huang, Chad Franzen, Joshua Hibschman, Daniel Afergan, Yang Li, Jeffrey Nichols, and Ranjitha Kumar.
\newblock Rico: A mobile app dataset for building data-driven design applications.
\newblock In {\em Proceedings of the 30th Annual ACM Symposium on User Interface Software and Technology}, pages 845--854, 2017.

\bibitem{deng2009imagenet}
Jia Deng, Wei Dong, Richard Socher, Li-Jia Li, Kai Li, and Li~Fei-Fei.
\newblock Imagenet: A large-scale hierarchical image database.
\newblock In {\em Proceedings of the IEEE/CVF Conference on Computer Vision and Pattern Recognition}, pages 248--255, 2009.

\bibitem{deng2024mind2web}
Xiang Deng, Yu~Gu, Boyuan Zheng, Shijie Chen, Sam Stevens, Boshi Wang, Huan Sun, and Yu~Su.
\newblock Mind2web: Towards a generalist agent for the web.
\newblock {\em Advances in Neural Information Processing Systems}, 36, 2024.

\bibitem{ding2023enhancing}
Ning Ding, Yulin Chen, Bokai Xu, Yujia Qin, Zhi Zheng, Shengding Hu, Zhiyuan Liu, Maosong Sun, and Bowen Zhou.
\newblock Enhancing chat language models by scaling high-quality instructional conversations.
\newblock {\em arXiv preprint arXiv:2305.14233}, 2023.

\bibitem{doan2024vintern}
Khang~T Doan, Bao~G Huynh, Dung~T Hoang, Thuc~D Pham, Nhat~H Pham, Quan Nguyen, Bang~Q Vo, and Suong~N Hoang.
\newblock Vintern-1b: An efficient multimodal large language model for vietnamese.
\newblock {\em arXiv preprint arXiv:2408.12480}, 2024.

\bibitem{dong2024xc24khd}
Xiaoyi Dong, Pan Zhang, Yuhang Zang, Yuhang Cao, Bin Wang, Linke Ouyang, Songyang Zhang, Haodong Duan, Wenwei Zhang, Yining Li, et~al.
\newblock Internlm-xcomposer2-4khd: A pioneering large vision-language model handling resolutions from 336 pixels to 4k hd.
\newblock {\em arXiv preprint arXiv:2404.06512}, 2024.

\bibitem{dosovitskiy2020image}
Alexey Dosovitskiy, Lucas Beyer, Alexander Kolesnikov, Dirk Weissenborn, Xiaohua Zhai, Thomas Unterthiner, Mostafa Dehghani, Matthias Minderer, Georg Heigold, Sylvain Gelly, et~al.
\newblock An image is worth 16x16 words: Transformers for image recognition at scale.
\newblock In {\em The International Conference on Learning Representations}, 2020.

\bibitem{du2022glm}
Zhengxiao Du, Yujie Qian, Xiao Liu, Ming Ding, Jiezhong Qiu, Zhilin Yang, and Jie Tang.
\newblock Glm: General language model pretraining with autoregressive blank infilling.
\newblock In {\em Proceedings of the Annual Meeting of the Association for Computational Linguistics}, pages 320--335, 2022.

\bibitem{duan2024vlmevalkit}
Haodong Duan, Junming Yang, Yuxuan Qiao, Xinyu Fang, Lin Chen, Yuan Liu, Xiaoyi Dong, Yuhang Zang, Pan Zhang, Jiaqi Wang, et~al.
\newblock Vlmevalkit: An open-source toolkit for evaluating large multi-modality models.
\newblock In {\em Proceedings of the 32nd ACM International Conference on Multimedia}, pages 11198--11201, 2024.

\bibitem{dubey2024llama3}
Abhimanyu Dubey, Abhinav Jauhri, Abhinav Pandey, Abhishek Kadian, Ahmad Al-Dahle, Aiesha Letman, Akhil Mathur, Alan Schelten, Amy Yang, Angela Fan, et~al.
\newblock The llama 3 herd of models.
\newblock {\em arXiv preprint arXiv:2407.21783}, 2024.

\bibitem{fang2024mmbench}
Xinyu Fang, Kangrui Mao, Haodong Duan, Xiangyu Zhao, Yining Li, Dahua Lin, and Kai Chen.
\newblock Mmbench-video: A long-form multi-shot benchmark for holistic video understanding.
\newblock {\em arXiv preprint arXiv:2406.14515}, 2024.

\bibitem{fang2022eva}
Yuxin Fang, Wen Wang, Binhui Xie, Quan Sun, Ledell Wu, Xinggang Wang, Tiejun Huang, Xinlong Wang, and Yue Cao.
\newblock Eva: Exploring the limits of masked visual representation learning at scale.
\newblock In {\em Proceedings of the IEEE/CVF Conference on Computer Vision and Pattern Recognition}, pages 19358--19369, 2023.

\bibitem{FineVideo}
Miquel Farré, Andi Marafioti, Lewis Tunstall, Leandro Von~Werra, and Thomas Wolf.
\newblock Finevideo.
\newblock \url{https://huggingface.co/datasets/HuggingFaceFV/finevideo}, 2024.

\bibitem{fu2023mme}
Chaoyou Fu, Peixian Chen, Yunhang Shen, Yulei Qin, Mengdan Zhang, Xu~Lin, Zhenyu Qiu, Wei Lin, Jinrui Yang, Xiawu Zheng, et~al.
\newblock Mme: A comprehensive evaluation benchmark for multimodal large language models.
\newblock {\em arXiv preprint arXiv:2306.13394}, 2023.

\bibitem{fu2024video}
Chaoyou Fu, Yuhan Dai, Yondong Luo, Lei Li, Shuhuai Ren, Renrui Zhang, Zihan Wang, Chenyu Zhou, Yunhang Shen, Mengdan Zhang, et~al.
\newblock Video-mme: The first-ever comprehensive evaluation benchmark of multi-modal llms in video analysis.
\newblock {\em arXiv preprint arXiv:2405.21075}, 2024.

\bibitem{fu2024blink}
Xingyu Fu, Yushi Hu, Bangzheng Li, Yu~Feng, Haoyu Wang, Xudong Lin, Dan Roth, Noah~A Smith, Wei-Chiu Ma, and Ranjay Krishna.
\newblock Blink: Multimodal large language models can see but not perceive.
\newblock {\em arXiv preprint arXiv:2404.12390}, 2024.

\bibitem{gao2024mini_internvl}
Zhangwei Gao, Zhe Chen, Erfei Cui, Yiming Ren, Weiyun Wang, Jinguo Zhu, Hao Tian, Shenglong Ye, Junjun He, Xizhou Zhu, et~al.
\newblock Mini-internvl: A flexible-transfer pocket multimodal model with 5\% parameters and 90\% performance.
\newblock {\em arXiv preprint arXiv:2410.16261}, 2024.

\bibitem{garcia2015overview}
Alba Garcia Seco De~Herrera, Henning M{\"u}ller, and Stefano Bromuri.
\newblock Overview of the imageclef 2015 medical classification task.
\newblock In {\em Working Notes of CLEF 2015--Cross Language Evaluation Forum, CEUR}, volume 1391. CEUR Workshop Proceedings, 2015.

\bibitem{glaive_code_assistant_v3}
GlaiveAI.
\newblock Glaive code assistant v3 dataset.
\newblock \url{https://huggingface.co/datasets/glaiveai/glaive-code-assistant-v3}, 2024.

\bibitem{goyal2017vqav2}
Yash Goyal, Tejas Khot, Douglas Summers-Stay, Dhruv Batra, and Devi Parikh.
\newblock Making the v in vqa matter: Elevating the role of image understanding in visual question answering.
\newblock In {\em Proceedings of the IEEE/CVF Conference on Computer Vision and Pattern Recognition}, pages 6904--6913, 2017.

\bibitem{gu2022wukong}
Jiaxi Gu, Xiaojun Meng, Guansong Lu, Lu~Hou, Niu Minzhe, Xiaodan Liang, Lewei Yao, Runhui Huang, Wei Zhang, Xin Jiang, et~al.
\newblock Wukong: A 100 million large-scale chinese cross-modal pre-training benchmark.
\newblock {\em Advances in Neural Information Processing Systems}, 35:26418--26431, 2022.

\bibitem{gu2024aquilavl}
Shuhao Gu, Jialing Zhang, Siyuan Zhou, Kevin Yu, Zhaohu Xing, Liangdong Wang, Zhou Cao, Jintao Jia, Zhuoyi Zhang, Yixuan Wang, et~al.
\newblock Infinity-mm: Scaling multimodal performance with large-scale and high-quality instruction data.
\newblock {\em arXiv preprint arXiv:2410.18558}, 2024.

\bibitem{guan2023hallusionbench}
Tianrui Guan, Fuxiao Liu, Xiyang Wu, Ruiqi Xian, Zongxia Li, Xiaoyu Liu, Xijun Wang, Lichang Chen, Furong Huang, Yaser Yacoob, et~al.
\newblock Hallusionbench: An advanced diagnostic suite for entangled language hallucination \& visual illusion in large vision-language models.
\newblock {\em arXiv preprint arXiv:2310.14566}, 2023.

\bibitem{guo2019eaten}
He~Guo, Xiameng Qin, Jiaming Liu, Junyu Han, Jingtuo Liu, and Errui Ding.
\newblock Eaten: Entity-aware attention for single shot visual text extraction.
\newblock In {\em International Conference on Document Analysis and Recognition}, pages 254--259, 2019.

\bibitem{gupta2016synthtext}
Ankush Gupta, Andrea Vedaldi, and Andrew Zisserman.
\newblock Synthetic data for text localisation in natural images.
\newblock In {\em Proceedings of the IEEE/CVF Conference on Computer Vision and Pattern Recognition}, pages 2315--2324, 2016.

\bibitem{he2024olympiadbench}
Chaoqun He, Renjie Luo, Yuzhuo Bai, Shengding Hu, Zhen~Leng Thai, Junhao Shen, Jinyi Hu, Xu~Han, Yujie Huang, Yuxiang Zhang, et~al.
\newblock Olympiadbench: A challenging benchmark for promoting agi with olympiad-level bilingual multimodal scientific problems.
\newblock {\em arXiv preprint arXiv:2402.14008}, 2024.

\bibitem{he2023wanjuan}
Conghui He, Zhenjiang Jin, Chao Xu, Jiantao Qiu, Bin Wang, Wei Li, Hang Yan, Jiaqi Wang, and Dahua Lin.
\newblock Wanjuan: A comprehensive multimodal dataset for advancing english and chinese large models.
\newblock {\em arXiv preprint arXiv:2308.10755}, 2023.

\bibitem{he2018icpr2018}
Mengchao He, Yuliang Liu, Zhibo Yang, Sheng Zhang, Canjie Luo, Feiyu Gao, Qi~Zheng, Yongpan Wang, Xin Zhang, and Lianwen Jin.
\newblock Icpr2018 contest on robust reading for multi-type web images.
\newblock In {\em International Conference on Pattern Recognition}, pages 7--12, 2018.

\bibitem{he2020pathvqa}
Xuehai He, Yichen Zhang, Luntian Mou, Eric Xing, and Pengtao Xie.
\newblock Pathvqa: 30000+ questions for medical visual question answering.
\newblock {\em arXiv preprint arXiv:2003.10286}, 2020.

\bibitem{hendrycks2021imagenet_r}
Dan Hendrycks, Steven Basart, Norman Mu, Saurav Kadavath, Frank Wang, Evan Dorundo, Rahul Desai, Tyler Zhu, Samyak Parajuli, Mike Guo, et~al.
\newblock The many faces of robustness: A critical analysis of out-of-distribution generalization.
\newblock In {\em Proceedings of the IEEE/CVF International Conference on Computer Vision}, pages 8340--8349, 2021.

\bibitem{hendrycks2020measuring}
Dan Hendrycks, Collin Burns, Steven Basart, Andy Zou, Mantas Mazeika, Dawn Song, and Jacob Steinhardt.
\newblock Measuring massive multitask language understanding.
\newblock In {\em The International Conference on Learning Representations}, 2020.

\bibitem{DBLP:conf/nips/HendrycksBKABTS21}
Dan Hendrycks, Collin Burns, Saurav Kadavath, Akul Arora, Steven Basart, Eric Tang, Dawn Song, and Jacob Steinhardt.
\newblock Measuring mathematical problem solving with the {MATH} dataset.
\newblock In Joaquin Vanschoren and Sai{-}Kit Yeung, editors, {\em Proceedings of the Neural Information Processing Systems Track on Datasets and Benchmarks 1, NeurIPS Datasets and Benchmarks 2021, December 2021, virtual}, 2021.

\bibitem{hendrycks2021imagenet_a}
Dan Hendrycks, Kevin Zhao, Steven Basart, Jacob Steinhardt, and Dawn Song.
\newblock Natural adversarial examples.
\newblock In {\em Proceedings of the IEEE/CVF Conference on Computer Vision and Pattern Recognition}, pages 15262--15271, 2021.

\bibitem{hessel2023androids}
Jack Hessel, Ana Marasovi{\'c}, Jena~D. Hwang, Lillian Lee, Jeff Da, Rowan Zellers, Robert Mankoff, and Yejin Choi.
\newblock Do androids laugh at electric sheep? {Humor} ``understanding'' benchmarks from {The New Yorker Caption Contest}.
\newblock In {\em Proceedings of the Annual Meeting of the Association for Computational Linguistics}, 2023.

\bibitem{hezarai_parsynth_ocr_200k}
Hezarai.
\newblock Parsynth-ocr-200k.
\newblock \url{https://huggingface.co/datasets/hezarai/parsynth-ocr-200k}, 2024.

\bibitem{honovich2022unnatural}
Or~Honovich, Thomas Scialom, Omer Levy, and Timo Schick.
\newblock Unnatural instructions: Tuning language models with (almost) no human labor.
\newblock {\em arXiv preprint arXiv:2212.09689}, 2022.

\bibitem{hosu2020koniq}
Vlad Hosu, Hanhe Lin, Tamas Sziranyi, and Dietmar Saupe.
\newblock Koniq-10k: An ecologically valid database for deep learning of blind image quality assessment.
\newblock {\em IEEE Transactions on Image Processing}, 29:4041--4056, 2020.

\bibitem{hsiao2022screenqa}
Yu-Chung Hsiao, Fedir Zubach, Gilles Baechler, Victor Carbune, Jason Lin, Maria Wang, Srinivas Sunkara, Yun Zhu, and Jindong Chen.
\newblock Screenqa: Large-scale question-answer pairs over mobile app screenshots.
\newblock {\em arXiv preprint arXiv:2209.08199}, 2022.

\bibitem{hu2024mplug_docowl_1_5}
Anwen Hu, Haiyang Xu, Jiabo Ye, Ming Yan, Liang Zhang, Bo~Zhang, Chen Li, Ji~Zhang, Qin Jin, Fei Huang, et~al.
\newblock mplug-docowl 1.5: Unified structure learning for ocr-free document understanding.
\newblock {\em arXiv preprint arXiv:2403.12895}, 2024.

\bibitem{hu2023medical}
Xinyue Hu, L~Gu, Q~An, M~Zhang, L~Liu, K~Kobayashi, T~Harada, R~Summers, and Y~Zhu.
\newblock Medical-diff-vqa: a large-scale medical dataset for difference visual question answering on chest x-ray images.
\newblock {\em PhysioNet}, 2023.

\bibitem{huang2020movienet}
Qingqiu Huang, Yu~Xiong, Anyi Rao, Jiaze Wang, and Dahua Lin.
\newblock Movienet: A holistic dataset for movie understanding.
\newblock In {\em European Conference on Computer Vision}, 2020.

\bibitem{huang2023ceval}
Yuzhen Huang, Yuzhuo Bai, Zhihao Zhu, Junlei Zhang, Jinghan Zhang, Tangjun Su, Junteng Liu, Chuancheng Lv, Yikai Zhang, Yao Fu, et~al.
\newblock C-eval: A multi-level multi-discipline chinese evaluation suite for foundation models.
\newblock {\em Advances in Neural Information Processing Systems}, 36, 2024.

\bibitem{huang2019icdar2019}
Zheng Huang, Kai Chen, Jianhua He, Xiang Bai, Dimosthenis Karatzas, Shijian Lu, and CV~Jawahar.
\newblock Icdar2019 competition on scanned receipt ocr and information extraction.
\newblock In {\em International Conference on Document Analysis and Recognition}, pages 1516--1520, 2019.

\bibitem{hudson2019gqa}
Drew~A Hudson and Christopher~D Manning.
\newblock Gqa: A new dataset for real-world visual reasoning and compositional question answering.
\newblock In {\em Proceedings of the IEEE/CVF Conference on Computer Vision and Pattern Recognition}, pages 6700--6709, 2019.

\bibitem{jia2022egotaskqa}
Baoxiong Jia, Ting Lei, Song-Chun Zhu, and Siyuan Huang.
\newblock Egotaskqa: Understanding human tasks in egocentric videos.
\newblock {\em Advances in Neural Information Processing Systems}, 35:3343--3360, 2022.

\bibitem{jiang2024mantis}
Dongfu Jiang, Xuan He, Huaye Zeng, Cong Wei, Max Ku, Qian Liu, and Wenhu Chen.
\newblock Mantis: Interleaved multi-image instruction tuning.
\newblock {\em arXiv preprint arXiv:2405.01483}, 2024.

\bibitem{jiao2024img}
Qirui Jiao, Daoyuan Chen, Yilun Huang, Yaliang Li, and Ying Shen.
\newblock Img-diff: Contrastive data synthesis for multimodal large language models.
\newblock {\em arXiv preprint arXiv:2408.04594}, 2024.

\bibitem{textocr_gpt4v_dataset}
Jimmycarter.
\newblock Textocr gpt-4v dataset.
\newblock \url{https://huggingface.co/datasets/jimmycarter/textocr-gpt4v}, 2023.

\bibitem{joshi2017triviaqa}
Mandar Joshi, Eunsol Choi, Daniel~S Weld, and Luke Zettlemoyer.
\newblock Triviaqa: A large scale distantly supervised challenge dataset for reading comprehension.
\newblock {\em arXiv preprint arXiv:1705.03551}, 2017.

\bibitem{kafle2018dvqa}
Kushal Kafle, Brian Price, Scott Cohen, and Christopher Kanan.
\newblock Dvqa: Understanding data visualizations via question answering.
\newblock In {\em Proceedings of the IEEE/CVF Conference on Computer Vision and Pattern Recognition}, pages 5648--5656, 2018.

\bibitem{kahou2017figureqa}
Samira~Ebrahimi Kahou, Vincent Michalski, Adam Atkinson, {\'A}kos K{\'a}d{\'a}r, Adam Trischler, and Yoshua Bengio.
\newblock Figureqa: An annotated figure dataset for visual reasoning.
\newblock {\em arXiv preprint arXiv:1710.07300}, 2017.

\bibitem{kapoor2025omniact}
Raghav Kapoor, Yash~Parag Butala, Melisa Russak, Jing~Yu Koh, Kiran Kamble, Waseem AlShikh, and Ruslan Salakhutdinov.
\newblock Omniact: A dataset and benchmark for enabling multimodal generalist autonomous agents for desktop and web.
\newblock In {\em European Conference on Computer Vision}, pages 161--178. Springer, 2025.

\bibitem{kazemi2023geomverse}
Mehran Kazemi, Hamidreza Alvari, Ankit Anand, Jialin Wu, Xi~Chen, and Radu Soricut.
\newblock Geomverse: A systematic evaluation of large models for geometric reasoning.
\newblock {\em arXiv preprint arXiv:2312.12241}, 2023.

\bibitem{kazemzadeh2014referitgame}
Sahar Kazemzadeh, Vicente Ordonez, Mark Matten, and Tamara Berg.
\newblock Referitgame: Referring to objects in photographs of natural scenes.
\newblock In {\em Proceedings of the 2014 Conference on Empirical Methods in Natural Language Processing}, pages 787--798, 2014.

\bibitem{kembhavi2016ai2d}
Aniruddha Kembhavi, Mike Salvato, Eric Kolve, Minjoon Seo, Hannaneh Hajishirzi, and Ali Farhadi.
\newblock A diagram is worth a dozen images.
\newblock In {\em European Conference on Computer Vision}, pages 235--251, 2016.

\bibitem{kembhavi2017tqa}
Aniruddha Kembhavi, Minjoon Seo, Dustin Schwenk, Jonghyun Choi, Ali Farhadi, and Hannaneh Hajishirzi.
\newblock Are you smarter than a sixth grader? textbook question answering for multimodal machine comprehension.
\newblock In {\em Proceedings of the IEEE/CVF Conference on Computer Vision and Pattern Recognition}, pages 4999--5007, 2017.

\bibitem{kiela2020hateful}
Douwe Kiela, Hamed Firooz, Aravind Mohan, Vedanuj Goswami, Amanpreet Singh, Pratik Ringshia, and Davide Testuggine.
\newblock The hateful memes challenge: Detecting hate speech in multimodal memes.
\newblock {\em Advances in Neural Information Processing Systems}, 33:2611--2624, 2020.

\bibitem{kim2022synthdog}
Geewook Kim, Teakgyu Hong, Moonbin Yim, JeongYeon Nam, Jinyoung Park, Jinyeong Yim, Wonseok Hwang, Sangdoo Yun, Dongyoon Han, and Seunghyun Park.
\newblock Ocr-free document understanding transformer.
\newblock In {\em European Conference on Computer Vision}, pages 498--517. Springer, 2022.

\bibitem{kirillov2023segment}
Alexander Kirillov, Eric Mintun, Nikhila Ravi, Hanzi Mao, Chloe Rolland, Laura Gustafson, Tete Xiao, Spencer Whitehead, Alexander~C Berg, Wan-Yen Lo, et~al.
\newblock Segment anything.
\newblock In {\em Proceedings of the IEEE/CVF International Conference on Computer Vision}, pages 4015--4026, 2023.

\bibitem{knowrohit07_know_saraswati_cot}
knowrohit07.
\newblock know saraswati cot dataset.
\newblock \url{https://huggingface.co/datasets/knowrohit07/know-saraswati-cot}, 2023.

\bibitem{kuang2023visual}
Jianfeng Kuang, Wei Hua, Dingkang Liang, Mingkun Yang, Deqiang Jiang, Bo~Ren, and Xiang Bai.
\newblock Visual information extraction in the wild: practical dataset and end-to-end solution.
\newblock In {\em International Conference on Document Analysis and Recognition}, pages 36--53. Springer, 2023.

\bibitem{kuznetsova2020openimage}
Alina Kuznetsova, Hassan Rom, Neil Alldrin, Jasper Uijlings, Ivan Krasin, Jordi Pont-Tuset, Shahab Kamali, Stefan Popov, Matteo Malloci, Alexander Kolesnikov, et~al.
\newblock The open images dataset v4: Unified image classification, object detection, and visual relationship detection at scale.
\newblock {\em IJCV}, 128(7):1956--1981, 2020.

\bibitem{naturalquestion}
Tom Kwiatkowski, Jennimaria Palomaki, Olivia Redfield, Michael Collins, Ankur Parikh, Chris Alberti, Danielle Epstein, Illia Polosukhin, Jacob Devlin, Kenton Lee, et~al.
\newblock Natural questions: a benchmark for question answering research.
\newblock {\em Transactions of the Association for Computational Linguistics}, 7:453--466, 2019.

\bibitem{lai2017race}
Guokun Lai, Qizhe Xie, Hanxiao Liu, Yiming Yang, and Eduard Hovy.
\newblock Race: Large-scale reading comprehension dataset from examinations.
\newblock {\em arXiv preprint arXiv:1704.04683}, 2017.

\bibitem{laion_gpt4v_dataset}
LAION.
\newblock Gpt-4v dataset.
\newblock \url{https://huggingface.co/datasets/laion/gpt4v-dataset}, 2023.

\bibitem{lau2018dataset}
Jason~J Lau, Soumya Gayen, Asma Ben~Abacha, and Dina Demner-Fushman.
\newblock A dataset of clinically generated visual questions and answers about radiology images.
\newblock {\em Scientific data}, 5:1--10, 2018.

\bibitem{2024docmatrix}
Hugo Lauren{\c{c}}on, Andr{\'e}s Marafioti, Victor Sanh, and L{\'e}o Tronchon.
\newblock Building and better understanding vision-language models: insights and future directions.
\newblock {\em arXiv preprint arXiv:2408.12637}, 2024.

\bibitem{laurenccon2024unlocking}
Hugo Lauren{\c{c}}on, L{\'e}o Tronchon, and Victor Sanh.
\newblock Unlocking the conversion of web screenshots into html code with the websight dataset.
\newblock {\em arXiv preprint arXiv:2403.09029}, 2024.

\bibitem{lerner2022viquae}
Paul Lerner, Olivier Ferret, Camille Guinaudeau, Herv{\'e} Le~Borgne, Romaric Besan{\c{c}}on, Jos{\'e}~G Moreno, and Jes{\'u}s Lov{\'o}n~Melgarejo.
\newblock Viquae, a dataset for knowledge-based visual question answering about named entities.
\newblock In {\em Proceedings of the 45th International ACM SIGIR Conference on Research and Development in Information Retrieval}, pages 3108--3120, 2022.

\bibitem{li2024llavaov}
Bo~Li, Yuanhan Zhang, Dong Guo, Renrui Zhang, Feng Li, Hao Zhang, Kaichen Zhang, Yanwei Li, Ziwei Liu, and Chunyuan Li.
\newblock Llava-onevision: Easy visual task transfer.
\newblock {\em arXiv preprint arXiv:2408.03326}, 2024.

\bibitem{li2024seedbench2plus}
Bohao Li, Yuying Ge, Yi~Chen, Yixiao Ge, Ruimao Zhang, and Ying Shan.
\newblock Seed-bench-2-plus: Benchmarking multimodal large language models with text-rich visual comprehension.
\newblock {\em arXiv preprint arXiv:2404.16790}, 2024.

\bibitem{li2024r}
Chunyi Li, Jianbo Zhang, Zicheng Zhang, Haoning Wu, Yuan Tian, Wei Sun, Guo Lu, Xiaohong Liu, Xiongkuo Min, Weisi Lin, et~al.
\newblock R-bench: Are your large multimodal model robust to real-world corruptions?
\newblock {\em arXiv preprint arXiv:2410.05474}, 2024.

\bibitem{li2023cmmlu}
Haonan Li, Yixuan Zhang, Fajri Koto, Yifei Yang, Hai Zhao, Yeyun Gong, Nan Duan, and Timothy Baldwin.
\newblock Cmmlu: Measuring massive multitask language understanding in chinese.
\newblock {\em arXiv preprint arXiv:2306.09212}, 2023.

\bibitem{li2024numinamath}
Jia Li, Edward Beeching, Lewis Tunstall, Ben Lipkin, Roman Soletskyi, Shengyi Huang, Kashif Rasul, Longhui Yu, Albert~Q Jiang, Ziju Shen, et~al.
\newblock Numinamath: The largest public dataset in ai4maths with 860k pairs of competition math problems and solutions.
\newblock {\em Hugging Face repository}, 2024.

\bibitem{li2024chemvlm}
Junxian Li, Di~Zhang, Xunzhi Wang, Zeying Hao, Jingdi Lei, Qian Tan, Cai Zhou, Wei Liu, Yaotian Yang, Xinrui Xiong, et~al.
\newblock Chemvlm: Exploring the power of multimodal large language models in chemistry area.
\newblock {\em arXiv preprint arXiv:2408.07246}, 2024.

\bibitem{li2023videochat}
KunChang Li, Yinan He, Yi~Wang, Yizhuo Li, Wenhai Wang, Ping Luo, Yali Wang, Limin Wang, and Yu~Qiao.
\newblock Videochat: Chat-centric video understanding.
\newblock {\em arXiv preprint arXiv:2305.06355}, 2023.

\bibitem{li2024mvbench}
Kunchang Li, Yali Wang, Yinan He, Yizhuo Li, Yi~Wang, Yi~Liu, Zun Wang, Jilan Xu, Guo Chen, Ping Luo, et~al.
\newblock Mvbench: A comprehensive multi-modal video understanding benchmark.
\newblock In {\em Proceedings of the IEEE/CVF Conference on Computer Vision and Pattern Recognition}, pages 22195--22206, 2024.

\bibitem{li2024multimodal}
Lei Li, Yuqi Wang, Runxin Xu, Peiyi Wang, Xiachong Feng, Lingpeng Kong, and Qi~Liu.
\newblock Multimodal arxiv: A dataset for improving scientific comprehension of large vision-language models.
\newblock {\em arXiv preprint arXiv:2403.00231}, 2024.

\bibitem{li2024omnicorpus}
Qingyun Li, Zhe Chen, Weiyun Wang, Wenhai Wang, Shenglong Ye, Zhenjiang Jin, Guanzhou Chen, Yinan He, Zhangwei Gao, Erfei Cui, et~al.
\newblock Omnicorpus: An unified multimodal corpus of 10 billion-level images interleaved with text.
\newblock {\em arXiv preprint arXiv:2406.08418}, 2024.

\bibitem{li2024gmai}
Tianbin Li, Yanzhou Su, Wei Li, Bin Fu, Zhe Chen, Ziyan Huang, Guoan Wang, Chenglong Ma, Ying Chen, Ming Hu, et~al.
\newblock Gmai-vl \& gmai-vl-5.5 m: A large vision-language model and a comprehensive multimodal dataset towards general medical ai.
\newblock {\em arXiv preprint arXiv:2411.14522}, 2024.

\bibitem{lieffects}
Wei Li, William~E Bishop, Alice Li, Christopher Rawles, Folawiyo Campbell-Ajala, Divya Tyamagundlu, and Oriana Riva.
\newblock On the effects of data scale on ui control agents.
\newblock In {\em The Thirty-eight Conference on Neural Information Processing Systems Datasets and Benchmarks Track}, 2024.

\bibitem{li2020widget}
Yang Li, Gang Li, Luheng He, Jingjie Zheng, Hong Li, and Zhiwei Guan.
\newblock Widget captioning: Generating natural language description for mobile user interface elements.
\newblock {\em arXiv preprint arXiv:2010.04295}, 2020.

\bibitem{li2021improved}
Yanghao Li, Chao-Yuan Wu, Haoqi Fan, Karttikeya Mangalam, Bo~Xiong, Jitendra Malik, and Christoph Feichtenhofer.
\newblock Mvitv2: Improved multiscale vision transformers for classification and detection.
\newblock In {\em Proceedings of the IEEE/CVF Conference on Computer Vision and Pattern Recognition}, pages 4804--4814, 2022.

\bibitem{li2024miniGemini}
Yanwei Li, Yuechen Zhang, Chengyao Wang, Zhisheng Zhong, Yixin Chen, Ruihang Chu, Shaoteng Liu, and Jiaya Jia.
\newblock Mini-gemini: Mining the potential of multi-modality vision language models.
\newblock {\em arXiv preprint arXiv:2403.18814}, 2024.

\bibitem{li2023pope}
Yifan Li, Yifan Du, Kun Zhou, Jinpeng Wang, Wayne~Xin Zhao, and Ji-Rong Wen.
\newblock Evaluating object hallucination in large vision-language models.
\newblock In {\em The Conference on Empirical Methods in Natural Language Processing}, pages 292--305, 2023.

\bibitem{li2023monkey}
Zhang Li, Biao Yang, Qiang Liu, Zhiyin Ma, Shuo Zhang, Jingxu Yang, Yabo Sun, Yuliang Liu, and Xiang Bai.
\newblock Monkey: Image resolution and text label are important things for large multi-modal models.
\newblock {\em arXiv preprint arXiv:2311.06607}, 2023.

\bibitem{li2023superclevr}
Zhuowan Li, Xingrui Wang, Elias Stengel-Eskin, Adam Kortylewski, Wufei Ma, Benjamin Van~Durme, and Alan~L Yuille.
\newblock Super-clevr: A virtual benchmark to diagnose domain robustness in visual reasoning.
\newblock In {\em Proceedings of the IEEE/CVF Conference on Computer Vision and Pattern Recognition}, pages 14963--14973, 2023.

\bibitem{SlimOrca}
Wing Lian, Guan Wang, Bleys Goodson, Eugene Pentland, Austin Cook, Chanvichet Vong, and "Teknium".
\newblock Slimorca: An open dataset of gpt-4 augmented flan reasoning traces, with verification.
\newblock \url{https://https://huggingface.co/Open-Orca/SlimOrca}, 2023.

\bibitem{lin2024vila}
Ji~Lin, Hongxu Yin, Wei Ping, Pavlo Molchanov, Mohammad Shoeybi, and Song Han.
\newblock Vila: On pre-training for visual language models.
\newblock In {\em Proceedings of the IEEE/CVF Conference on Computer Vision and Pattern Recognition}, pages 26689--26699, 2024.

\bibitem{lindstrom2022clevrmath}
Adam~Dahlgren Lindstr{\"o}m and Savitha~Sam Abraham.
\newblock Clevr-math: A dataset for compositional language, visual and mathematical reasoning.
\newblock {\em arXiv preprint arXiv:2208.05358}, 2022.

\bibitem{liu2021slake}
Bo~Liu, Li-Ming Zhan, Li~Xu, Lin Ma, Yan Yang, and Xiao-Ming Wu.
\newblock Slake: A semantically-labeled knowledge-enhanced dataset for medical visual question answering.
\newblock In {\em 2021 IEEE 18th International Symposium on Biomedical Imaging}, pages 1650--1654. IEEE, 2021.

\bibitem{liu2020casia}
Brian Liu, Xianchao Xu, and Yu~Zhang.
\newblock Offline handwritten chinese text recognition with convolutional neural networks.
\newblock {\em arXiv preprint arXiv:2006.15619}, 2020.

\bibitem{liu2023vsr}
Fangyu Liu, Guy Emerson, and Nigel Collier.
\newblock Visual spatial reasoning.
\newblock {\em Transactions of the Association for Computational Linguistics}, 11:635--651, 2023.

\bibitem{liu2023lrv-instruction}
Fuxiao Liu, Kevin Lin, Linjie Li, Jianfeng Wang, Yaser Yacoob, and Lijuan Wang.
\newblock Aligning large multi-modal model with robust instruction tuning.
\newblock {\em arXiv preprint arXiv:2306.14565}, 2023.

\bibitem{liu2023mmc}
Fuxiao Liu, Xiaoyang Wang, Wenlin Yao, Jianshu Chen, Kaiqiang Song, Sangwoo Cho, Yaser Yacoob, and Dong Yu.
\newblock Mmc: Advancing multimodal chart understanding with large-scale instruction tuning.
\newblock {\em arXiv preprint arXiv:2311.10774}, 2023.

\bibitem{liu2023improved}
Haotian Liu, Chunyuan Li, Yuheng Li, and Yong~Jae Lee.
\newblock Improved baselines with visual instruction tuning.
\newblock {\em arXiv preprint arXiv:2310.03744}, 2023.

\bibitem{liu2024llavanext}
Haotian Liu, Chunyuan Li, Yuheng Li, Bo~Li, Yuanhan Zhang, Sheng Shen, and Yong~Jae Lee.
\newblock Llava-next: Improved reasoning, ocr, and world knowledge.
\newblock \url{https://llava-vl.github.io/blog/2024-01-30-llava-next/}, January 2024.

\bibitem{liu2020ntu}
Jun Liu, Amir Shahroudy, Mauricio Perez, Gang Wang, Ling-Yu Duan, and Alex~C Kot.
\newblock Ntu rgb+d 120: A large-scale benchmark for 3d human activity understanding.
\newblock {\em IEEE Transactions on Pattern Analysis and Machine Intelligence}, 42(10):2684--2701, 2020.

\bibitem{grounding_dino}
Shilong Liu, Zhaoyang Zeng, Tianhe Ren, Feng Li, Hao Zhang, Jie Yang, Qing Jiang, Chunyuan Li, Jianwei Yang, Hang Su, et~al.
\newblock Grounding dino: Marrying dino with grounded pre-training for open-set object detection.
\newblock In {\em European Conference on Computer Vision}, pages 38--55. Springer, 2025.

\bibitem{liu2024cmmmath}
Wentao Liu, Qianjun Pan, Yi~Zhang, Zhuo Liu, Ji~Wu, Jie Zhou, Aimin Zhou, Qin Chen, Bo~Jiang, and Liang He.
\newblock Cmm-math: A chinese multimodal math dataset to evaluate and enhance the mathematics reasoning of large multimodal models.
\newblock {\em arXiv preprint arXiv:2409.02834}, 2024.

\bibitem{liu2024mminstruct}
Yangzhou Liu, Yue Cao, Zhangwei Gao, Weiyun Wang, Zhe Chen, Wenhai Wang, Hao Tian, Lewei Lu, Xizhou Zhu, Tong Lu, et~al.
\newblock Mminstruct: A high-quality multi-modal instruction tuning dataset with extensive diversity.
\newblock {\em arXiv preprint arXiv:2407.15838}, 2024.

\bibitem{liu2023mmbench}
Yuan Liu, Haodong Duan, Yuanhan Zhang, Bo~Li, Songyang Zhang, Wangbo Zhao, Yike Yuan, Jiaqi Wang, Conghui He, Ziwei Liu, et~al.
\newblock Mmbench: Is your multi-modal model an all-around player?
\newblock {\em arXiv preprint arXiv:2307.06281}, 2023.

\bibitem{liu2024points}
Yuan Liu, Zhongyin Zhao, Ziyuan Zhuang, Le~Tian, Xiao Zhou, and Jie Zhou.
\newblock Points: Improving your vision-language model with affordable strategies.
\newblock {\em arXiv preprint arXiv:2409.04828}, 2024.

\bibitem{liu2023ocrbench}
Yuliang Liu, Zhang Li, Hongliang Li, Wenwen Yu, Mingxin Huang, Dezhi Peng, Mingyu Liu, Mingrui Chen, Chunyuan Li, Lianwen Jin, et~al.
\newblock On the hidden mystery of ocr in large multimodal models.
\newblock {\em arXiv preprint arXiv:2305.07895}, 2023.

\bibitem{liu2024mmdu}
Ziyu Liu, Tao Chu, Yuhang Zang, Xilin Wei, Xiaoyi Dong, Pan Zhang, Zijian Liang, Yuanjun Xiong, Yu~Qiao, Dahua Lin, et~al.
\newblock Mmdu: A multi-turn multi-image dialog understanding benchmark and instruction-tuning dataset for lvlms.
\newblock {\em arXiv preprint arXiv:2406.11833}, 2024.

\bibitem{liu2024oryx}
Zuyan Liu, Yuhao Dong, Ziwei Liu, Winston Hu, Jiwen Lu, and Yongming Rao.
\newblock Oryx mllm: On-demand spatial-temporal understanding at arbitrary resolution.
\newblock {\em arXiv preprint arXiv:2409.12961}, 2024.

\bibitem{loshchilov2017adamw}
Ilya Loshchilov and Frank Hutter.
\newblock Decoupled weight decay regularization.
\newblock {\em arXiv preprint arXiv:1711.05101}, 2017.

\bibitem{lu2024deepseekvl}
Haoyu Lu, Wen Liu, Bo~Zhang, Bingxuan Wang, Kai Dong, Bo~Liu, Jingxiang Sun, Tongzheng Ren, Zhuoshu Li, Yaofeng Sun, et~al.
\newblock Deepseek-vl: Towards real-world vision-language understanding.
\newblock {\em arXiv preprint arXiv:2403.05525}, 2024.

\bibitem{lu2023mathvista}
Pan Lu, Hritik Bansal, Tony Xia, Jiacheng Liu, Chunyuan Li, Hannaneh Hajishirzi, Hao Cheng, Kai-Wei Chang, Michel Galley, and Jianfeng Gao.
\newblock Mathvista: Evaluating mathematical reasoning of foundation models in visual contexts.
\newblock {\em arXiv preprint arXiv:2310.02255}, 2023.

\bibitem{lu2021geometry3k}
Pan Lu, Ran Gong, Shibiao Jiang, Liang Qiu, Siyuan Huang, Xiaodan Liang, and Song-Chun Zhu.
\newblock Inter-gps: Interpretable geometry problem solving with formal language and symbolic reasoning.
\newblock {\em arXiv preprint arXiv:2105.04165}, 2021.

\bibitem{lu2022scienceqa}
Pan Lu, Swaroop Mishra, Tanglin Xia, Liang Qiu, Kai-Wei Chang, Song-Chun Zhu, Oyvind Tafjord, Peter Clark, and Ashwin Kalyan.
\newblock Learn to explain: Multimodal reasoning via thought chains for science question answering.
\newblock {\em Advances in Neural Information Processing Systems}, 35:2507--2521, 2022.

\bibitem{lu2022tablemwp}
Pan Lu, Liang Qiu, Kai-Wei Chang, Ying~Nian Wu, Song-Chun Zhu, Tanmay Rajpurohit, Peter Clark, and Ashwin Kalyan.
\newblock Dynamic prompt learning via policy gradient for semi-structured mathematical reasoning.
\newblock {\em arXiv preprint arXiv:2209.14610}, 2022.

\bibitem{lu2021iconqa}
Pan Lu, Liang Qiu, Jiaqi Chen, Tony Xia, Yizhou Zhao, Wei Zhang, Zhou Yu, Xiaodan Liang, and Song-Chun Zhu.
\newblock Iconqa: A new benchmark for abstract diagram understanding and visual language reasoning.
\newblock {\em arXiv preprint arXiv:2110.13214}, 2021.

\bibitem{lu2024gui}
Quanfeng Lu, Wenqi Shao, Zitao Liu, Fanqing Meng, Boxuan Li, Botong Chen, Siyuan Huang, Kaipeng Zhang, Yu~Qiao, and Ping Luo.
\newblock Gui odyssey: A comprehensive dataset for cross-app gui navigation on mobile devices.
\newblock {\em arXiv preprint arXiv:2406.08451}, 2024.

\bibitem{lu2024ovis}
Shiyin Lu, Yang Li, Qing-Guo Chen, Zhao Xu, Weihua Luo, Kaifu Zhang, and Han-Jia Ye.
\newblock Ovis: Structural embedding alignment for multimodal large language model.
\newblock {\em arXiv preprint arXiv:2405.20797}, 2024.

\bibitem{lu2024bluelm}
Xudong Lu, Yinghao Chen, Cheng Chen, Hui Tan, Boheng Chen, Yina Xie, Rui Hu, Guanxin Tan, Renshou Wu, Yan Hu, et~al.
\newblock Bluelm-v-3b: Algorithm and system co-design for multimodal large language models on mobile devices.
\newblock {\em arXiv preprint arXiv:2411.10640}, 2024.

\bibitem{lu2024wildvision}
Yujie Lu, Dongfu Jiang, Wenhu Chen, William~Yang Wang, Yejin Choi, and Bill~Yuchen Lin.
\newblock Wildvision: Evaluating vision-language models in the wild with human preferences.
\newblock {\em arXiv preprint arXiv:2406.11069}, 2024.

\bibitem{luo2024mono_internvl}
Gen Luo, Xue Yang, Wenhan Dou, Zhaokai Wang, Jifeng Dai, Yu~Qiao, and Xizhou Zhu.
\newblock Mono-internvl: Pushing the boundaries of monolithic multimodal large language models with endogenous visual pre-training.
\newblock {\em arXiv preprint arXiv:2410.08202}, 2024.

\bibitem{luo2023wizardcoder}
Ziyang Luo, Can Xu, Pu~Zhao, Qingfeng Sun, Xiubo Geng, Wenxiang Hu, Chongyang Tao, Jing Ma, Qingwei Lin, and Daxin Jiang.
\newblock Wizardcoder: Empowering code large language models with evol-instruct.
\newblock {\em arXiv preprint arXiv:2306.08568}, 2023.

\bibitem{Maaz2024VideoGPT+}
Muhammad Maaz, Hanoona Rasheed, Salman Khan, and Fahad Khan.
\newblock Videogpt+: Integrating image and video encoders for enhanced video understanding.
\newblock {\em arXiv preprint arXiv:2406.09418}, 2024.

\bibitem{mangalam2023egoschema}
Karttikeya Mangalam, Raiymbek Akshulakov, and Jitendra Malik.
\newblock Egoschema: A diagnostic benchmark for very long-form video language understanding.
\newblock {\em Advances in Neural Information Processing Systems}, 36:46212--46244, 2023.

\bibitem{mao2017deepart}
Hui Mao, Ming Cheung, and James She.
\newblock Deepart: Learning joint representations of visual arts.
\newblock In {\em Proceedings of the ACM International Conference on Multimedia}, pages 1183--1191, 2017.

\bibitem{mao2016generation}
Junhua Mao, Jonathan Huang, Alexander Toshev, Oana Camburu, Alan~L Yuille, and Kevin Murphy.
\newblock Generation and comprehension of unambiguous object descriptions.
\newblock In {\em Proceedings of the IEEE/CVF Conference on Computer Vision and Pattern Recognition}, pages 11--20, 2016.

\bibitem{marino2019okvqa}
Kenneth Marino, Mohammad Rastegari, Ali Farhadi, and Roozbeh Mottaghi.
\newblock Ok-vqa: A visual question answering benchmark requiring external knowledge.
\newblock In {\em Proceedings of the IEEE/CVF Conference on Computer Vision and Pattern Recognition}, pages 3195--3204, 2019.

\bibitem{MarkrAI_KOpen_HQ_Hermes_2.5_60K}
MarkrAI.
\newblock Kopen-hq-hermes-2.5-60k dataset.
\newblock \url{https://huggingface.co/datasets/MarkrAI/KOpen-HQ-Hermes-2.5-60K}, 2023.

\bibitem{marti2002iam}
U-V Marti and Horst Bunke.
\newblock The iam-database: an english sentence database for offline handwriting recognition.
\newblock {\em International Journal on Document Analysis and Recognition}, 5:39--46, 2002.

\bibitem{masry2022chartqa}
Ahmed Masry, Xuan~Long Do, Jia~Qing Tan, Shafiq Joty, and Enamul Hoque.
\newblock Chartqa: A benchmark for question answering about charts with visual and logical reasoning.
\newblock In {\em Proceedings of the Annual Meeting of the Association for Computational Linguistics}, pages 2263--2279, 2022.

\bibitem{masry2023unichart}
Ahmed Masry, Parsa Kavehzadeh, Xuan~Long Do, Enamul Hoque, and Shafiq Joty.
\newblock Unichart: A universal vision-language pretrained model for chart comprehension and reasoning.
\newblock {\em arXiv preprint arXiv:2305.14761}, 2023.

\bibitem{mathew2022infographicvqa}
Minesh Mathew, Viraj Bagal, Rub{\`e}n Tito, Dimosthenis Karatzas, Ernest Valveny, and CV~Jawahar.
\newblock Infographicvqa.
\newblock In {\em Proceedings of the IEEE/CVF Winter Conference on Applications of Computer Vision}, pages 1697--1706, 2022.

\bibitem{mathew2021docvqa}
Minesh Mathew, Dimosthenis Karatzas, and CV~Jawahar.
\newblock Docvqa: A dataset for vqa on document images.
\newblock In {\em Proceedings of the IEEE/CVF Winter Conference on Applications of Computer Vision}, pages 2200--2209, 2021.

\bibitem{mckinzie2024mm1}
Brandon McKinzie, Zhe Gan, Jean-Philippe Fauconnier, Sam Dodge, Bowen Zhang, Philipp Dufter, Dhruti Shah, Xianzhi Du, Futang Peng, Floris Weers, et~al.
\newblock Mm1: Methods, analysis \& insights from multimodal llm pre-training.
\newblock {\em arXiv preprint arXiv:2403.09611}, 2024.

\bibitem{meng2024mmiu}
Fanqing Meng, Jin Wang, Chuanhao Li, Quanfeng Lu, Hao Tian, Jiaqi Liao, Xizhou Zhu, Jifeng Dai, Yu~Qiao, Ping Luo, et~al.
\newblock Mmiu: Multimodal multi-image understanding for evaluating large vision-language models.
\newblock {\em arXiv preprint arXiv:2408.02718}, 2024.

\bibitem{methani2020plotqa}
Nitesh Methani, Pritha Ganguly, Mitesh~M Khapra, and Pratyush Kumar.
\newblock Plotqa: Reasoning over scientific plots.
\newblock In {\em Proceedings of the IEEE/CVF Winter Conference on Applications of Computer Vision}, pages 1527--1536, 2020.

\bibitem{mishra2019ocrvqa}
Anand Mishra, Shashank Shekhar, Ajeet~Kumar Singh, and Anirban Chakraborty.
\newblock Ocr-vqa: Visual question answering by reading text in images.
\newblock In {\em International Conference on Document Analysis and Recognition}, pages 947--952, 2019.

\bibitem{mitra2024orcamath}
Arindam Mitra, Hamed Khanpour, Corby Rosset, and Ahmed Awadallah.
\newblock Orca-math: Unlocking the potential of slms in grade school math.
\newblock {\em arXiv preprint arXiv:2402.14830}, 2024.

\bibitem{nan2024openvid}
Kepan Nan, Rui Xie, Penghao Zhou, Tiehan Fan, Zhenheng Yang, Zhijie Chen, Xiang Li, Jian Yang, and Ying Tai.
\newblock Openvid-1m: A large-scale high-quality dataset for text-to-video generation.
\newblock {\em arXiv preprint arXiv:2407.02371}, 2024.

\bibitem{obeid2020chart}
Jason Obeid and Enamul Hoque.
\newblock Chart-to-text: Generating natural language descriptions for charts by adapting the transformer model.
\newblock {\em arXiv preprint arXiv:2010.09142}, 2020.

\bibitem{gpt4v}
OpenAI.
\newblock Gpt-4v(ision) system card.
\newblock \url{https://cdn.openai.com/papers/GPTV\_System\_Card.pdf}, 2023.

\bibitem{patraucean2024perception}
Viorica Patraucean, Lucas Smaira, Ankush Gupta, Adria Recasens, Larisa Markeeva, Dylan Banarse, Skanda Koppula, Mateusz Malinowski, Yi~Yang, Carl Doersch, et~al.
\newblock Perception test: A diagnostic benchmark for multimodal video models.
\newblock {\em Advances in Neural Information Processing Systems}, 36, 2024.

\bibitem{peng2023kosmos2}
Zhiliang Peng, Wenhui Wang, Li~Dong, Yaru Hao, Shaohan Huang, Shuming Ma, and Furu Wei.
\newblock Kosmos-2: Grounding multimodal large language models to the world.
\newblock {\em arXiv preprint arXiv:2306.14824}, 2023.

\bibitem{radford2021clip}
Alec Radford, Jong~Wook Kim, Chris Hallacy, Aditya Ramesh, Gabriel Goh, Sandhini Agarwal, Girish Sastry, Amanda Askell, Pamela Mishkin, Jack Clark, et~al.
\newblock Learning transferable visual models from natural language supervision.
\newblock In {\em International Conference on Machine Learning}, pages 8748--8763, 2021.

\bibitem{no_robots}
Nazneen Rajani, Lewis Tunstall, Edward Beeching, Nathan Lambert, Alexander~M. Rush, and Thomas Wolf.
\newblock No robots.
\newblock Hugging Face repository, \url{https://huggingface.co/datasets/HuggingFaceH4/no\_robots}, 2023.

\bibitem{ranjan2021learning}
Viresh Ranjan, Udbhav Sharma, Thu Nguyen, and Minh Hoai.
\newblock Learning to count everything.
\newblock In {\em Proceedings of the IEEE/CVF Conference on Computer Vision and Pattern Recognition}, pages 3394--3403, 2021.

\bibitem{rawles2024androidinthewild}
Christopher Rawles, Alice Li, Daniel Rodriguez, Oriana Riva, and Timothy Lillicrap.
\newblock Androidinthewild: A large-scale dataset for android device control.
\newblock {\em Advances in Neural Information Processing Systems}, 36, 2024.

\bibitem{recht2019imagenetv2}
Benjamin Recht, Rebecca Roelofs, Ludwig Schmidt, and Vaishaal Shankar.
\newblock Do imagenet classifiers generalize to imagenet?
\newblock In {\em International Conference on Machine Learning}, pages 5389--5400, 2019.

\bibitem{reid2024gemini1_5}
Machel Reid, Nikolay Savinov, Denis Teplyashin, Dmitry Lepikhin, Timothy Lillicrap, Jean-baptiste Alayrac, Radu Soricut, Angeliki Lazaridou, Orhan Firat, Julian Schrittwieser, et~al.
\newblock Gemini 1.5: Unlocking multimodal understanding across millions of tokens of context.
\newblock {\em arXiv preprint arXiv:2403.05530}, 2024.

\bibitem{rohrbach2015dataset}
Anna Rohrbach, Marcus Rohrbach, Niket Tandon, and Bernt Schiele.
\newblock A dataset for movie description.
\newblock In {\em Proceedings of the IEEE/CVF Conference on Computer Vision and Pattern Recognition}, pages 3202--3212, 2015.

\bibitem{sakaguchi2020winogrande}
Keisuke Sakaguchi, Ronan Le~Bras, Chandra Bhagavatula, and Yejin Choi.
\newblock Winogrande: An adversarial winograd schema challenge at scale.
\newblock In {\em Proceedings of the AAAI Conference on Artificial Intelligence}, volume~34, pages 8732--8740, 2020.

\bibitem{schuhmann2022laion5b}
Christoph Schuhmann, Romain Beaumont, Richard Vencu, Cade Gordon, Ross Wightman, Mehdi Cherti, Theo Coombes, Aarush Katta, Clayton Mullis, Mitchell Wortsman, et~al.
\newblock Laion-5b: An open large-scale dataset for training next generation image-text models.
\newblock {\em Advances in Neural Information Processing Systems}, 35:25278--25294, 2022.

\bibitem{schuhmann2022laioncoco}
Christoph Schuhmann, Andreas Köpf, Richard Vencu, Theo Coombes, and Romain Beaumont.
\newblock Laion coco: 600m synthetic captions from laion2b-en.
\newblock \url{https://laion.ai/blog/laion-coco/}, 2022.

\bibitem{schwenk2022aokvqa}
Dustin Schwenk, Apoorv Khandelwal, Christopher Clark, Kenneth Marino, and Roozbeh Mottaghi.
\newblock A-okvqa: A benchmark for visual question answering using world knowledge.
\newblock In {\em European Conference on Computer Vision}, pages 146--162, 2022.

\bibitem{seo2015solving}
Minjoon Seo, Hannaneh Hajishirzi, Ali Farhadi, Oren Etzioni, and Clint Malcolm.
\newblock Solving geometry problems: Combining text and diagram interpretation.
\newblock In {\em Proceedings of the 2015 Conference on Empirical Methods in Natural Language Processing}, pages 1466--1476, 2015.

\bibitem{shah2019kvqa}
Sanket Shah, Anand Mishra, Naganand Yadati, and Partha~Pratim Talukdar.
\newblock Kvqa: Knowledge-aware visual question answering.
\newblock In {\em Proceedings of the AAAI Conference on Artificial Intelligence}, volume~33, pages 8876--8884, 2019.

\bibitem{shao2019objects365}
Shuai Shao, Zeming Li, Tianyuan Zhang, Chao Peng, Gang Yu, Xiangyu Zhang, Jing Li, and Jian Sun.
\newblock Objects365: A large-scale, high-quality dataset for object detection.
\newblock In {\em Proceedings of the IEEE/CVF International Conference on Computer Vision}, pages 8430--8439, 2019.

\bibitem{shi2017rctw17}
Baoguang Shi, Cong Yao, Minghui Liao, Mingkun Yang, Pei Xu, Linyan Cui, Serge Belongie, Shijian Lu, and Xiang Bai.
\newblock Icdar2017 competition on reading chinese text in the wild (rctw-17).
\newblock In {\em International Conference on Document Analysis and Recognition}, volume~1, pages 1429--1434, 2017.

\bibitem{shi2024eagle}
Min Shi, Fuxiao Liu, Shihao Wang, Shijia Liao, Subhashree Radhakrishnan, De-An Huang, Hongxu Yin, Karan Sapra, Yaser Yacoob, Humphrey Shi, et~al.
\newblock Eagle: Exploring the design space for multimodal llms with mixture of encoders.
\newblock {\em arXiv preprint arXiv:2408.15998}, 2024.

\bibitem{sidorov2020textcaps}
Oleksii Sidorov, Ronghang Hu, Marcus Rohrbach, and Amanpreet Singh.
\newblock Textcaps: a dataset for image captioning with reading comprehension.
\newblock In {\em European Conference on Computer Vision}, pages 742--758, 2020.

\bibitem{singh2019textvqa}
Amanpreet Singh, Vivek Natarajan, Meet Shah, Yu~Jiang, Xinlei Chen, Dhruv Batra, Devi Parikh, and Marcus Rohrbach.
\newblock Towards vqa models that can read.
\newblock In {\em Proceedings of the IEEE/CVF Conference on Computer Vision and Pattern Recognition}, pages 8317--8326, 2019.

\bibitem{singh2021textocr}
Amanpreet Singh, Guan Pang, Mandy Toh, Jing Huang, Wojciech Galuba, and Tal Hassner.
\newblock Textocr: Towards large-scale end-to-end reasoning for arbitrary-shaped scene text.
\newblock In {\em Proceedings of the IEEE/CVF Conference on Computer Vision and Pattern Recognition}, pages 8802--8812, 2021.

\bibitem{singh2025benchmarking}
Shweta Singh, Aayan Yadav, Jitesh Jain, Humphrey Shi, Justin Johnson, and Karan Desai.
\newblock Benchmarking object detectors with coco: A new path forward.
\newblock In {\em European Conference on Computer Vision}, pages 279--295. Springer, 2025.

\bibitem{snell2024scaling}
Charlie Snell, Jaehoon Lee, Kelvin Xu, and Aviral Kumar.
\newblock Scaling llm test-time compute optimally can be more effective than scaling model parameters.
\newblock {\em arXiv preprint arXiv:2408.03314}, 2024.

\bibitem{suhr2018corpus}
Alane Suhr, Stephanie Zhou, Ally Zhang, Iris Zhang, Huajun Bai, and Yoav Artzi.
\newblock A corpus for reasoning about natural language grounded in photographs.
\newblock {\em arXiv preprint arXiv:1811.00491}, 2018.

\bibitem{sujet-finance}
Hamed~Rahimi Sujet~AI, Allaa~Boutaleb.
\newblock Sujet-finance-qa-vision-100k: A large-scale dataset for financial document vqa.
\newblock \url{https://huggingface.co/datasets/sujet-ai/Sujet-Finance-QA-Vision-100k}, 2024.

\bibitem{sun2024parrot}
Hai-Long Sun, Da-Wei Zhou, Yang Li, Shiyin Lu, Chao Yi, Qing-Guo Chen, Zhao Xu, Weihua Luo, Kaifu Zhang, De-Chuan Zhan, et~al.
\newblock Parrot: Multilingual visual instruction tuning.
\newblock {\em arXiv preprint arXiv:2406.02539}, 2024.

\bibitem{sun2019investigating}
Kai Sun, Dian Yu, Dong Yu, and Claire Cardie.
\newblock Investigating prior knowledge for challenging chinese machine reading comprehension.
\newblock {\em Transactions of the Association for Computational Linguistics}, 8:141--155, 2020.

\bibitem{sun2023emu}
Quan Sun, Qiying Yu, Yufeng Cui, Fan Zhang, Xiaosong Zhang, Yueze Wang, Hongcheng Gao, Jingjing Liu, Tiejun Huang, and Xinlong Wang.
\newblock Generative pretraining in multimodality.
\newblock In {\em The International Conference on Learning Representations}, 2024.

\bibitem{sun2023moss}
Tianxiang Sun, Xiaotian Zhang, Zhengfu He, Peng Li, Qinyuan Cheng, Hang Yan, Xiangyang Liu, Yunfan Shao, Qiong Tang, Xingjian Zhao, et~al.
\newblock Moss: Training conversational language models from synthetic data.
\newblock {\em arXiv preprint arXiv:2307.15020}, 7, 2023.

\bibitem{sun2019lsvt}
Yipeng Sun, Zihan Ni, Chee-Kheng Chng, Yuliang Liu, Canjie Luo, Chun~Chet Ng, Junyu Han, Errui Ding, Jingtuo Liu, Dimosthenis Karatzas, et~al.
\newblock Icdar 2019 competition on large-scale street view text with partial labeling-rrc-lsvt.
\newblock In {\em International Conference on Document Analysis and Recognition}, pages 1557--1562, 2019.

\bibitem{sun2023aligning}
Zhiqing Sun, Sheng Shen, Shengcao Cao, Haotian Liu, Chunyuan Li, Yikang Shen, Chuang Gan, Liang-Yan Gui, Yu-Xiong Wang, Yiming Yang, et~al.
\newblock Aligning large multimodal models with factually augmented rlhf.
\newblock {\em arXiv preprint arXiv:2309.14525}, 2023.

\bibitem{suzgun2023bigbench}
Mirac Suzgun, Nathan Scales, Nathanael Sch{\"a}rli, Sebastian Gehrmann, Yi~Tay, Hyung~Won Chung, Aakanksha Chowdhery, Quoc~V Le, Ed~H Chi, Denny Zhou, et~al.
\newblock Challenging big-bench tasks and whether chain-of-thought can solve them.
\newblock {\em arXiv preprint arXiv:2210.09261}, 2022.

\bibitem{tanaka2021visualmrc}
Ryota Tanaka, Kyosuke Nishida, and Sen Yoshida.
\newblock Visualmrc: Machine reading comprehension on document images.
\newblock In {\em Proceedings of the AAAI Conference on Artificial Intelligence}, volume~35, pages 13878--13888, 2021.

\bibitem{tang2023vistext}
Benny~J Tang, Angie Boggust, and Arvind Satyanarayan.
\newblock Vistext: A benchmark for semantically rich chart captioning.
\newblock {\em arXiv preprint arXiv:2307.05356}, 2023.

\bibitem{tang2024mtvqa}
Jingqun Tang, Qi~Liu, Yongjie Ye, Jinghui Lu, Shu Wei, Chunhui Lin, Wanqing Li, Mohamad Fitri Faiz~Bin Mahmood, Hao Feng, Zhen Zhao, et~al.
\newblock Mtvqa: Benchmarking multilingual text-centric visual question answering.
\newblock {\em arXiv preprint arXiv:2405.11985}, 2024.

\bibitem{team2023gemini}
Gemini Team, Rohan Anil, Sebastian Borgeaud, Yonghui Wu, Jean-Baptiste Alayrac, Jiahui Yu, Radu Soricut, Johan Schalkwyk, Andrew~M Dai, Anja Hauth, et~al.
\newblock Gemini: a family of highly capable multimodal models.
\newblock {\em arXiv preprint arXiv:2312.11805}, 2023.

\bibitem{qwen2.5}
Qwen Team.
\newblock Qwen2.5: A party of foundation models.
\newblock \url{https://qwenlm.github.io/blog/qwen2.5/}, September 2024.

\bibitem{qwq-32b-preview}
Qwen Team.
\newblock Qwq: Reflect deeply on the boundaries of the unknown.
\newblock \url{https://qwenlm.github.io/blog/qwq-32b-preview/}, November 2024.

\bibitem{teknium2024hermes3}
Ryan Teknium, Jeffrey Quesnelle, and Chen Guang.
\newblock Hermes 3 technical report.
\newblock {\em arXiv preprint arXiv:2408.11857}, 2024.

\bibitem{tian2024mminterleaved}
Changyao Tian, Xizhou Zhu, Yuwen Xiong, Weiyun Wang, Zhe Chen, Wenhai Wang, Yuntao Chen, Lewei Lu, Tong Lu, Jie Zhou, et~al.
\newblock Mm-interleaved: Interleaved image-text generative modeling via multi-modal feature synchronizer.
\newblock {\em arXiv preprint arXiv:2401.10208}, 2024.

\bibitem{tito2023hier}
Rub{\`e}n Tito, Dimosthenis Karatzas, and Ernest Valveny.
\newblock Hierarchical multimodal transformers for multipage docvqa.
\newblock {\em Pattern Recognition}, 144:109834, 2023.

\bibitem{tong2024cambrian}
Shengbang Tong, Ellis Brown, Penghao Wu, Sanghyun Woo, Manoj Middepogu, Sai~Charitha Akula, Jihan Yang, Shusheng Yang, Adithya Iyer, Xichen Pan, et~al.
\newblock Cambrian-1: A fully open, vision-centric exploration of multimodal llms.
\newblock {\em arXiv preprint arXiv:2406.16860}, 2024.

\bibitem{touvron2023llama2}
Hugo Touvron, Louis Martin, Kevin Stone, Peter Albert, Amjad Almahairi, Yasmine Babaei, Nikolay Bashlykov, Soumya Batra, Prajjwal Bhargava, Shruti Bhosale, et~al.
\newblock Llama 2: Open foundation and fine-tuned chat models.
\newblock {\em arXiv preprint arXiv:2307.09288}, 2023.

\bibitem{ustalov2023toloka}
Dmitry Ustalov, Nikita Pavlichenko, Sergey Koshelev, Daniil Likhobaba, and Alisa Smirnova.
\newblock Toloka visual question answering benchmark.
\newblock {\em arXiv preprint arXiv:2309.16511}, 2023.

\bibitem{van2018inaturalist}
Grant Van~Horn, Oisin Mac~Aodha, Yang Song, Yin Cui, Chen Sun, Alex Shepard, Hartwig Adam, Pietro Perona, and Serge Belongie.
\newblock The inaturalist species classification and detection dataset.
\newblock In {\em Proceedings of the IEEE/CVF Conference on Computer Vision and Pattern Recognition}, pages 8769--8778, 2018.

\bibitem{veit2016coco}
Andreas Veit, Tomas Matera, Lukas Neumann, Jiri Matas, and Serge Belongie.
\newblock Coco-text: Dataset and benchmark for text detection and recognition in natural images.
\newblock {\em arXiv preprint arXiv:1601.07140}, 2016.

\bibitem{cyrillic}
Konstantin Verner.
\newblock Cyrillic handwriting dataset.
\newblock \url{https://www.kaggle.com/datasets/constantinwerner/cyrillic-handwriting-dataset}, 2020.

\bibitem{wang2021screen2words}
Bryan Wang, Gang Li, Xin Zhou, Zhourong Chen, Tovi Grossman, and Yang Li.
\newblock Screen2words: Automatic mobile ui summarization with multimodal learning.
\newblock In {\em The 34th Annual ACM Symposium on User Interface Software and Technology}, pages 498--510, 2021.

\bibitem{wang2024muirbench}
Fei Wang, Xingyu Fu, James~Y Huang, Zekun Li, Qin Liu, Xiaogeng Liu, Mingyu~Derek Ma, Nan Xu, Wenxuan Zhou, Kai Zhang, et~al.
\newblock Muirbench: A comprehensive benchmark for robust multi-image understanding.
\newblock {\em arXiv preprint arXiv:2406.09411}, 2024.

\bibitem{wang2019imagenet_sketch}
Haohan Wang, Songwei Ge, Zachary Lipton, and Eric~P Xing.
\newblock Learning robust global representations by penalizing local predictive power.
\newblock {\em Advances in Neural Information Processing Systems}, 32, 2019.

\bibitem{wang2023v3det}
Jiaqi Wang, Pan Zhang, Tao Chu, Yuhang Cao, Yujie Zhou, Tong Wu, Bin Wang, Conghui He, and Dahua Lin.
\newblock V3det: Vast vocabulary visual detection dataset.
\newblock In {\em Proceedings of the IEEE/CVF International Conference on Computer Vision}, pages 19844--19854, 2023.

\bibitem{wang2023lvisinstruct4v}
Junke Wang, Lingchen Meng, Zejia Weng, Bo~He, Zuxuan Wu, and Yu-Gang Jiang.
\newblock To see is to believe: Prompting gpt-4v for better visual instruction tuning.
\newblock {\em arXiv preprint arXiv:2311.07574}, 2023.

\bibitem{wang2024measuring}
Ke~Wang, Junting Pan, Weikang Shi, Zimu Lu, Mingjie Zhan, and Hongsheng Li.
\newblock Measuring multimodal mathematical reasoning with math-vision dataset.
\newblock {\em arXiv preprint arXiv:2402.14804}, 2024.

\bibitem{wang2024qwen2vl}
Peng Wang, Shuai Bai, Sinan Tan, Shijie Wang, Zhihao Fan, Jinze Bai, Keqin Chen, Xuejing Liu, Jialin Wang, Wenbin Ge, et~al.
\newblock Qwen2-vl: Enhancing vision-language model's perception of the world at any resolution.
\newblock {\em arXiv preprint arXiv:2409.12191}, 2024.

\bibitem{one-peace}
Peng Wang, Shijie Wang, Junyang Lin, Shuai Bai, Xiaohuan Zhou, Jingren Zhou, Xinggang Wang, and Chang Zhou.
\newblock One-peace: Exploring one general representation model toward unlimited modalities.
\newblock {\em arXiv:2305.11172}, 2023.

\bibitem{wang2023cogvlm}
Weihan Wang, Qingsong Lv, Wenmeng Yu, Wenyi Hong, Ji~Qi, Yan Wang, Junhui Ji, Zhuoyi Yang, Lei Zhao, Xixuan Song, et~al.
\newblock Cogvlm: Visual expert for pretrained language models.
\newblock {\em arXiv preprint arXiv:2311.03079}, 2023.

\bibitem{wang2024mpo}
Weiyun Wang, Zhe Chen, Wenhai Wang, Yue Cao, Yangzhou Liu, Zhangwei Gao, Jinguo Zhu, Xizhou Zhu, Lewei Lu, Yu~Qiao, and Jifeng Dai.
\newblock Enhancing the reasoning ability of multimodal large language models via mixed preference optimization.
\newblock {\em arXiv preprint arXiv:2411.10442}, 2024.

\bibitem{wang2024allseeingv2}
Weiyun Wang, Yiming Ren, Haowen Luo, Tiantong Li, Chenxiang Yan, Zhe Chen, Wenhai Wang, Qingyun Li, Lewei Lu, Xizhou Zhu, et~al.
\newblock The all-seeing project v2: Towards general relation comprehension of the open world.
\newblock {\em arXiv preprint arXiv:2402.19474}, 2024.

\bibitem{wang2023allseeing}
Weiyun Wang, Min Shi, Qingyun Li, Wenhai Wang, Zhenhang Huang, Linjie Xing, Zhe Chen, Hao Li, Xizhou Zhu, Zhiguo Cao, et~al.
\newblock The all-seeing project: Towards panoptic visual recognition and understanding of the open world.
\newblock In {\em The International Conference on Learning Representations}, 2024.

\bibitem{wang2023internimage}
Wenhai Wang, Jifeng Dai, Zhe Chen, Zhenhang Huang, Zhiqi Li, Xizhou Zhu, Xiaowei Hu, Tong Lu, Lewei Lu, Hongsheng Li, et~al.
\newblock Internimage: Exploring large-scale vision foundation models with deformable convolutions.
\newblock In {\em Proceedings of the IEEE/CVF Conference on Computer Vision and Pattern Recognition}, pages 14408--14419, 2023.

\bibitem{wang2020general}
Xinyu Wang, Yuliang Liu, Chunhua Shen, Chun~Chet Ng, Canjie Luo, Lianwen Jin, Chee~Seng Chan, Anton van~den Hengel, and Liangwei Wang.
\newblock On the general value of evidence, and bilingual scene-text visual question answering.
\newblock In {\em Proceedings of the IEEE/CVF Conference on Computer Vision and Pattern Recognition}, pages 10126--10135, 2020.

\bibitem{wang2024mementos}
Xiyao Wang, Yuhang Zhou, Xiaoyu Liu, Hongjin Lu, Yuancheng Xu, Feihong He, Jaehong Yoon, Taixi Lu, Gedas Bertasius, Mohit Bansal, et~al.
\newblock Mementos: A comprehensive benchmark for multimodal large language model reasoning over image sequences.
\newblock {\em arXiv preprint arXiv:2401.10529}, 2024.

\bibitem{wang2024xcoder80k}
Yejie Wang, Keqing He, Dayuan Fu, Zhuoma Gongque, Heyang Xu, Yanxu Chen, Zhexu Wang, Yujia Fu, Guanting Dong, Muxi Diao, et~al.
\newblock How do your code llms perform? empowering code instruction tuning with high-quality data.
\newblock {\em arXiv preprint arXiv:2409.03810}, 2024.

\bibitem{wang2024internvideo2}
Yi~Wang, Kunchang Li, Xinhao Li, Jiashuo Yu, Yinan He, Guo Chen, Baoqi Pei, Rongkun Zheng, Jilan Xu, Zun Wang, et~al.
\newblock Internvideo2: Scaling video foundation models for multimodal video understanding.
\newblock {\em arXiv preprint arXiv:2403.15377}, 2024.

\bibitem{wang2024charxiv}
Zirui Wang, Mengzhou Xia, Luxi He, Howard Chen, Yitao Liu, Richard Zhu, Kaiqu Liang, Xindi Wu, Haotian Liu, Sadhika Malladi, et~al.
\newblock Charxiv: Charting gaps in realistic chart understanding in multimodal llms.
\newblock {\em arXiv preprint arXiv:2406.18521}, 2024.

\bibitem{wei2021finetuned}
Jason Wei, Maarten Bosma, Vincent~Y Zhao, Kelvin Guu, Adams~Wei Yu, Brian Lester, Nan Du, Andrew~M Dai, and Quoc~V Le.
\newblock Finetuned language models are zero-shot learners.
\newblock {\em arXiv preprint arXiv:2109.01652}, 2021.

\bibitem{wu2024star}
Bo~Wu, Shoubin Yu, Zhenfang Chen, Joshua~B Tenenbaum, and Chuang Gan.
\newblock Star: A benchmark for situated reasoning in real-world videos.
\newblock {\em arXiv preprint arXiv:2405.09711}, 2024.

\bibitem{pmccase}
Chaoyi Wu.
\newblock Pmc-casereport.
\newblock \url{https://huggingface.co/datasets/chaoyi-wu/PMC-CaseReport}, 2023.

\bibitem{wu2023towards}
Chaoyi Wu, Xiaoman Zhang, Ya~Zhang, Yanfeng Wang, and Weidi Xie.
\newblock Towards generalist foundation model for radiology.
\newblock {\em arXiv preprint arXiv:2308.02463}, 2023.

\bibitem{wu2024longvideobench}
Haoning Wu, Dongxu Li, Bei Chen, and Junnan Li.
\newblock Longvideobench: A benchmark for long-context interleaved video-language understanding.
\newblock {\em arXiv preprint arXiv:2407.15754}, 2024.

\bibitem{xia2023structchart}
Renqiu Xia, Bo~Zhang, Haoyang Peng, Hancheng Ye, Xiangchao Yan, Peng Ye, Botian Shi, Junchi Yan, and Yu~Qiao.
\newblock Structchart: Perception, structuring, reasoning for visual chart understanding.
\newblock {\em arXiv preprint arXiv:2309.11268}, 2023.

\bibitem{xiao2018upernet}
Tete Xiao, Yingcheng Liu, Bolei Zhou, Yuning Jiang, and Jian Sun.
\newblock Unified perceptual parsing for scene understanding.
\newblock In {\em European Conference on Computer Vision}, pages 418--434, 2018.

\bibitem{xu2024wizardlm}
Can Xu, Qingfeng Sun, Kai Zheng, Xiubo Geng, Pu~Zhao, Jiazhan Feng, Chongyang Tao, Qingwei Lin, and Daxin Jiang.
\newblock Wizardlm: Empowering large pre-trained language models to follow complex instructions.
\newblock In {\em The International Conference on Learning Representations}, 2024.

\bibitem{xu2024magpie}
Zhangchen Xu, Fengqing Jiang, Luyao Niu, Yuntian Deng, Radha Poovendran, Yejin Choi, and Bill~Yuchen Lin.
\newblock Magpie: Alignment data synthesis from scratch by prompting aligned llms with nothing.
\newblock {\em arXiv preprint arXiv:2406.08464}, 2024.

\bibitem{uninext}
B.~Yan, Yi~Jiang, Jiannan Wu, D.~Wang, Ping Luo, Zehuan Yuan, and Huchuan Lu.
\newblock Universal instance perception as object discovery and retrieval.
\newblock In {\em Proceedings of the IEEE/CVF Conference on Computer Vision and Pattern Recognition}, 2023.

\bibitem{yang2024qwen2}
An~Yang, Baosong Yang, Binyuan Hui, Bo~Zheng, Bowen Yu, Chang Zhou, Chengpeng Li, Chengyuan Li, Dayiheng Liu, Fei Huang, et~al.
\newblock Qwen2 technical report.
\newblock {\em arXiv preprint arXiv:2407.10671}, 2024.

\bibitem{yang2024vript}
Dongjie Yang, Suyuan Huang, Chengqiang Lu, Xiaodong Han, Haoxin Zhang, Yan Gao, Yao Hu, and Hai Zhao.
\newblock Vript: A video is worth thousands of words.
\newblock {\em arXiv preprint arXiv:2406.06040}, 2024.

\bibitem{Firefly}
Jianxin Yang.
\newblock Firefly: A chinese conversational large language model.
\newblock \url{https://github.com/yangjianxin1/Firefly}, 2023.

\bibitem{yang2023longqlora}
Jianxin Yang.
\newblock Longqlora: Efficient and effective method to extend context length of large language models.
\newblock {\em arXiv preprint arXiv:2311.04879}, 2023.

\bibitem{yang2023open}
Wenjuan Yang, Xuhui Zhang, Bing Ma, Yanqun Wang, Yujia Wu, Jianxing Yan, Yongwei Liu, Chao Zhang, Jicheng Wan, Yue Wang, et~al.
\newblock An open dataset for intelligent recognition and classification of abnormal condition in longwall mining.
\newblock {\em Scientific Data}, 10(1):416, 2023.

\bibitem{yao2011human}
Bangpeng Yao, Xiaoye Jiang, Aditya Khosla, Andy~Lai Lin, Leonidas Guibas, and Li~Fei-Fei.
\newblock Human action recognition by learning bases of action attributes and parts.
\newblock In {\em 2011 International Conference on Computer Vision}, pages 1331--1338, 2011.

\bibitem{yao2024minicpm}
Yuan Yao, Tianyu Yu, Ao~Zhang, Chongyi Wang, Junbo Cui, Hongji Zhu, Tianchi Cai, Haoyu Li, Weilin Zhao, Zhihui He, et~al.
\newblock Minicpm-v: A gpt-4v level mllm on your phone.
\newblock {\em arXiv preprint arXiv:2408.01800}, 2024.

\bibitem{ye2023mplug2}
Qinghao Ye, Haiyang Xu, Jiabo Ye, Ming Yan, Haowei Liu, Qi~Qian, Ji~Zhang, Fei Huang, and Jingren Zhou.
\newblock mplug-owl2: Revolutionizing multi-modal large language model with modality collaboration.
\newblock {\em arXiv preprint arXiv:2311.04257}, 2023.

\bibitem{yi2019clevrer}
Kexin Yi, Chuang Gan, Yunzhu Li, Pushmeet Kohli, Jiajun Wu, Antonio Torralba, and Joshua~B Tenenbaum.
\newblock Clevrer: Collision events for video representation and reasoning.
\newblock {\em arXiv preprint arXiv:1910.01442}, 2019.

\bibitem{mmtbench}
Kaining Ying, Fanqing Meng, Jin Wang, Zhiqian Li, Han Lin, Yue Yang, Hao Zhang, Wenbo Zhang, Yuqi Lin, Shuo Liu, Jiayi Lei, Quanfeng Lu, Runjian Chen, Peng Xu, Renrui Zhang, Haozhe Zhang, Peng Gao, Yali Wang, Yu~Qiao, Ping Luo, Kaipeng Zhang, and Wenqi Shao.
\newblock Mmt-bench: A comprehensive multimodal benchmark for evaluating large vision-language models towards multitask agi.
\newblock {\em arXiv preprint arXiv:2404.16006}, 2024.

\bibitem{you2023ferret}
Haoxuan You, Haotian Zhang, Zhe Gan, Xianzhi Du, Bowen Zhang, Zirui Wang, Liangliang Cao, Shih-Fu Chang, and Yinfei Yang.
\newblock Ferret: Refer and ground anything anywhere at any granularity.
\newblock {\em arXiv preprint arXiv:2310.07704}, 2023.

\bibitem{young2024yi}
Alex Young, Bei Chen, Chao Li, Chengen Huang, Ge~Zhang, Guanwei Zhang, Heng Li, Jiangcheng Zhu, Jianqun Chen, Jing Chang, et~al.
\newblock Yi: Open foundation models by 01. ai.
\newblock {\em arXiv preprint arXiv:2403.04652}, 2024.

\bibitem{yu2016refcoco}
Licheng Yu, Patrick Poirson, Shan Yang, Alexander~C Berg, and Tamara~L Berg.
\newblock Modeling context in referring expressions.
\newblock In {\em European Conference on Computer Vision}, pages 69--85, 2016.

\bibitem{yu2023metamath}
Longhui Yu, Weisen Jiang, Han Shi, Jincheng Yu, Zhengying Liu, Yu~Zhang, James~T Kwok, Zhenguo Li, Adrian Weller, and Weiyang Liu.
\newblock Metamath: Bootstrap your own mathematical questions for large language models.
\newblock {\em arXiv preprint arXiv:2309.12284}, 2023.

\bibitem{yu2024rlaifv}
Tianyu Yu, Haoye Zhang, Yuan Yao, Yunkai Dang, Da~Chen, Xiaoman Lu, Ganqu Cui, Taiwen He, Zhiyuan Liu, Tat-Seng Chua, and Maosong Sun.
\newblock Rlaif-v: Aligning mllms through open-source ai feedback for super gpt-4v trustworthiness.
\newblock {\em arXiv preprint arXiv:2405.17220}, 2024.

\bibitem{yu2023mmvet}
Weihao Yu, Zhengyuan Yang, Linjie Li, Jianfeng Wang, Kevin Lin, Zicheng Liu, Xinchao Wang, and Lijuan Wang.
\newblock Mm-vet: Evaluating large multimodal models for integrated capabilities.
\newblock {\em arXiv preprint arXiv:2308.02490}, 2023.

\bibitem{yu2024mmvetv2}
Weihao Yu, Zhengyuan Yang, Linfeng Ren, Linjie Li, Jianfeng Wang, Kevin Lin, Chung-Ching Lin, Zicheng Liu, Lijuan Wang, and Xinchao Wang.
\newblock Mm-vet v2: A challenging benchmark to evaluate large multimodal models for integrated capabilities.
\newblock {\em arXiv preprint arXiv:2408.00765}, 2024.

\bibitem{yu2024texthawk2}
Ya-Qi Yu, Minghui Liao, Jiwen Zhang, and Jihao Wu.
\newblock Texthawk2: A large vision-language model excels in bilingual ocr and grounding with 16x fewer tokens.
\newblock {\em arXiv preprint arXiv:2410.05261}, 2024.

\bibitem{yu2023paraphrasing}
Yijiong Yu.
\newblock "paraphrasing the original text" makes high accuracy long-context qa.
\newblock {\em arXiv preprint arXiv:2312.11193}, 2023.

\bibitem{yuan2019ctw}
Tai-Ling Yuan, Zhe Zhu, Kun Xu, Cheng-Jun Li, Tai-Jiang Mu, and Shi-Min Hu.
\newblock A large chinese text dataset in the wild.
\newblock {\em Journal of Computer Science and Technology}, 34:509--521, 2019.

\bibitem{yuan2022syntax}
Ye~Yuan, Xiao Liu, Wondimu Dikubab, Hui Liu, Zhilong Ji, Zhongqin Wu, and Xiang Bai.
\newblock Syntax-aware network for handwritten mathematical expression recognition.
\newblock In {\em Proceedings of the IEEE/CVF Conference on Computer Vision and Pattern Recognition}, pages 4553--4562, 2022.

\bibitem{yue2023mmmu}
Xiang Yue, Yuansheng Ni, Kai Zhang, Tianyu Zheng, Ruoqi Liu, Ge~Zhang, Samuel Stevens, Dongfu Jiang, Weiming Ren, Yuxuan Sun, et~al.
\newblock Mmmu: A massive multi-discipline multimodal understanding and reasoning benchmark for expert agi.
\newblock {\em arXiv preprint arXiv:2311.16502}, 2023.

\bibitem{yue2024mmmu}
Xiang Yue, Tianyu Zheng, Yuansheng Ni, Yubo Wang, Kai Zhang, Shengbang Tong, Yuxuan Sun, Ming Yin, Botao Yu, Ge~Zhang, et~al.
\newblock Mmmu-pro: A more robust multi-discipline multimodal understanding benchmark.
\newblock {\em arXiv preprint arXiv:2409.02813}, 2024.

\bibitem{zala2023hierarchical}
Abhay Zala, Jaemin Cho, Satwik Kottur, Xilun Chen, Barlas Oguz, Yashar Mehdad, and Mohit Bansal.
\newblock Hierarchical video-moment retrieval and step-captioning.
\newblock In {\em Proceedings of the IEEE/CVF Conference on Computer Vision and Pattern Recognition}, pages 23056--23065, 2023.

\bibitem{zellers2019hellaswag}
Rowan Zellers, Ari Holtzman, Yonatan Bisk, Ali Farhadi, and Yejin Choi.
\newblock Hellaswag: Can a machine really finish your sentence?
\newblock In {\em Proceedings of the Annual Meeting of the Association for Computational Linguistics}, pages 4791--4800, 2019.

\bibitem{zhai2023siglip}
Xiaohua Zhai, Basil Mustafa, Alexander Kolesnikov, and Lucas Beyer.
\newblock Sigmoid loss for language image pre-training.
\newblock In {\em Proceedings of the IEEE/CVF International Conference on Computer Vision}, pages 11975--11986, 2023.

\bibitem{zhang2019rmsnorm}
Biao Zhang and Rico Sennrich.
\newblock Root mean square layer normalization.
\newblock {\em Advances in Neural Information Processing Systems}, 32, 2019.

\bibitem{zhang2024inifinitymath}
Bo-Wen Zhang, Yan Yan, Lin Li, and Guang Liu.
\newblock Infinitymath: A scalable instruction tuning dataset in programmatic mathematical reasoning.
\newblock In {\em Proceedings of the 33rd ACM International Conference on Information and Knowledge Management}, pages 5405--5409, 2024.

\bibitem{zhang2024mm1_5}
Haotian Zhang, Mingfei Gao, Zhe Gan, Philipp Dufter, Nina Wenzel, Forrest Huang, Dhruti Shah, Xianzhi Du, Bowen Zhang, Yanghao Li, et~al.
\newblock Mm1.5: Methods, analysis \& insights from multimodal llm fine-tuning.
\newblock {\em arXiv preprint arXiv:2409.20566}, 2024.

\bibitem{ferretv2}
Haotian Zhang, Haoxuan You, Philipp Dufter, Bowen Zhang, Chen Chen, Hong-You Chen, Tsu-Jui Fu, William~Yang Wang, Shih-Fu Chang, Zhe Gan, et~al.
\newblock Ferret-v2: An improved baseline for referring and grounding with large language models.
\newblock {\em arXiv preprint arXiv:2404.07973}, 2024.

\bibitem{zhang2024longcite}
Jiajie Zhang, Yushi Bai, Xin Lv, Wanjun Gu, Danqing Liu, Minhao Zou, Shulin Cao, Lei Hou, Yuxiao Dong, Ling Feng, and Juanzi Li.
\newblock Longcite: Enabling llms to generate fine-grained citations in long-context qa.
\newblock {\em arXiv preprint arXiv:2409.02897}, 2024.

\bibitem{zhang2025mathverse}
Renrui Zhang, Dongzhi Jiang, Yichi Zhang, Haokun Lin, Ziyu Guo, Pengshuo Qiu, Aojun Zhou, Pan Lu, Kai-Wei Chang, Yu~Qiao, et~al.
\newblock Mathverse: Does your multi-modal llm truly see the diagrams in visual math problems?
\newblock In {\em European Conference on Computer Vision}, pages 169--186. Springer, 2025.

\bibitem{zhang2024mavis}
Renrui Zhang, Xinyu Wei, Dongzhi Jiang, Yichi Zhang, Ziyu Guo, Chengzhuo Tong, Jiaming Liu, Aojun Zhou, Bin Wei, Shanghang Zhang, et~al.
\newblock Mavis: Mathematical visual instruction tuning.
\newblock {\em arXiv preprint arXiv:2407.08739}, 2024.

\bibitem{zhang2019rects}
Rui Zhang, Yongsheng Zhou, Qianyi Jiang, Qi~Song, Nan Li, Kai Zhou, Lei Wang, Dong Wang, Minghui Liao, Mingkun Yang, et~al.
\newblock Icdar 2019 robust reading challenge on reading chinese text on signboard.
\newblock In {\em International Conference on Document Analysis and Recognition}, pages 1577--1581, 2019.

\bibitem{zhang2024vcr}
Tianyu Zhang, Suyuchen Wang, Lu~Li, Ge~Zhang, Perouz Taslakian, Sai Rajeswar, Jie Fu, Bang Liu, and Yoshua Bengio.
\newblock Vcr: Visual caption restoration.
\newblock {\em arXiv preprint arXiv:2406.06462}, 2024.

\bibitem{zhang2023pmc}
Xiaoman Zhang, Chaoyi Wu, Ziheng Zhao, Weixiong Lin, Ya~Zhang, Yanfeng Wang, and Weidi Xie.
\newblock Pmc-vqa: Visual instruction tuning for medical visual question answering.
\newblock {\em arXiv preprint arXiv:2305.10415}, 2023.

\bibitem{Zhang2023gaokao}
Xiaotian Zhang, Chunyang Li, Yi~Zong, Zhengyu Ying, Liang He, and Xipeng Qiu.
\newblock Evaluating the performance of large language models on gaokao benchmark.
\newblock {\em arXiv preprint arXiv:2305.12474}, 2023.

\bibitem{zhang2023llavar}
Yanzhe Zhang, Ruiyi Zhang, Jiuxiang Gu, Yufan Zhou, Nedim Lipka, Diyi Yang, and Tong Sun.
\newblock Llavar: Enhanced visual instruction tuning for text-rich image understanding.
\newblock {\em arXiv preprint arXiv:2306.17107}, 2023.

\bibitem{zhang2024mme}
Yi-Fan Zhang, Huanyu Zhang, Haochen Tian, Chaoyou Fu, Shuangqing Zhang, Junfei Wu, Feng Li, Kun Wang, Qingsong Wen, Zhang Zhang, et~al.
\newblock Mme-realworld: Could your multimodal llm challenge high-resolution real-world scenarios that are difficult for humans?
\newblock {\em arXiv preprint arXiv:2408.13257}, 2024.

\bibitem{zhang2024video}
Yuanhan Zhang, Jinming Wu, Wei Li, Bo~Li, Zejun Ma, Ziwei Liu, and Chunyuan Li.
\newblock Video instruction tuning with synthetic data.
\newblock {\em arXiv preprint arXiv:2410.02713}, 2024.

\bibitem{zhao2024mirb}
Bingchen Zhao, Yongshuo Zong, Letian Zhang, and Timothy Hospedales.
\newblock Benchmarking multi-image understanding in vision and language models: Perception, knowledge, reasoning, and multi-hop reasoning.
\newblock {\em arXiv preprint arXiv:2406.12742}, 2024.

\bibitem{zhao2023svit}
Bo~Zhao, Boya Wu, and Tiejun Huang.
\newblock Svit: Scaling up visual instruction tuning.
\newblock {\em arXiv preprint arXiv:2307.04087}, 2023.

\bibitem{zheng2024multimodal}
Mingyu Zheng, Xinwei Feng, Qingyi Si, Qiaoqiao She, Zheng Lin, Wenbin Jiang, and Weiping Wang.
\newblock Multimodal table understanding.
\newblock {\em arXiv preprint arXiv:2406.08100}, 2024.

\bibitem{opencodeinterpreter}
Tianyu Zheng, Ge~Zhang, Tianhao Shen, Xueling Liu, Bill~Yuchen Lin, Jie Fu, Wenhu Chen, and Xiang Yue.
\newblock Opencodeinterpreter: Integrating code generation with execution and refinement.
\newblock {\em arXiv preprint arXiv:2402.14658}, 2024.

\bibitem{zheng2021global}
Xinyi Zheng, Douglas Burdick, Lucian Popa, Xu~Zhong, and Nancy Xin~Ru Wang.
\newblock Global table extractor (gte): A framework for joint table identification and cell structure recognition using visual context.
\newblock In {\em Proceedings of the IEEE/CVF Winter Conference on Applications of Computer Vision}, pages 697--706, 2021.

\bibitem{zhou2017ade20k}
Bolei Zhou, Hang Zhao, Xavier Puig, Sanja Fidler, Adela Barriuso, and Antonio Torralba.
\newblock Scene parsing through ade20k dataset.
\newblock In {\em Proceedings of the IEEE/CVF Conference on Computer Vision and Pattern Recognition}, pages 633--641, 2017.

\bibitem{zhou2024lima}
Chunting Zhou, Pengfei Liu, Puxin Xu, Srinivasan Iyer, Jiao Sun, Yuning Mao, Xuezhe Ma, Avia Efrat, Ping Yu, Lili Yu, et~al.
\newblock Lima: Less is more for alignment.
\newblock {\em Advances in Neural Information Processing Systems}, 36, 2024.

\bibitem{MLVU}
Junjie Zhou, Yan Shu, Bo~Zhao, Boya Wu, Shitao Xiao, Xi~Yang, Yongping Xiong, Bo~Zhang, Tiejun Huang, and Zheng Liu.
\newblock Mlvu: A comprehensive benchmark for multi-task long video understanding.
\newblock {\em arXiv preprint arXiv:2406.04264}, 2024.

\bibitem{zhu2023minigpt4}
Deyao Zhu, Jun Chen, Xiaoqian Shen, Xiang Li, and Mohamed Elhoseiny.
\newblock Minigpt-4: Enhancing vision-language understanding with advanced large language models.
\newblock In {\em The International Conference on Learning Representations}, 2024.

\bibitem{zhu2016visual7w}
Yuke Zhu, Oliver Groth, Michael Bernstein, and Li~Fei-Fei.
\newblock Visual7w: Grounded question answering in images.
\newblock In {\em Proceedings of the IEEE/CVF Conference on Computer Vision and Pattern Recognition}, pages 4995--5004, 2016.

\end{thebibliography}
}

\end{document}